\documentclass{article}

\usepackage{microtype}
\usepackage{graphicx}
\usepackage{subcaption}
\usepackage{booktabs} 
\usepackage{multirow}
\usepackage{hyperref}
\usepackage{comment}

 \usepackage[preprint]{icml2026}

\usepackage{amsmath}
\usepackage{amssymb}
\usepackage{mathtools}
\usepackage{amsthm}
\usepackage{enumitem}

\usepackage{xspace}
\usepackage{overpic} 
\newcommand{\eg}{\textit{e.g.},\xspace}
\newcommand{\ie}{\textit{i.e.},\xspace}

\newcommand{\modelname}{\textcolor{black}{GranFEM}\xspace}

\usepackage[capitalize,noabbrev]{cleveref}

\theoremstyle{plain}

\theoremstyle{definition}

\theoremstyle{remark}

\usepackage[textsize=tiny]{todonotes}

\icmltitlerunning{Learning Sparse Finite Element Operators on Graphs}

\hypersetup{
    pdfsubject={Preprint}
}

\begin{document}

\twocolumn[
  \icmltitle{
  Learning Sparse Finite Element Operators on Graphs}



  \icmlsetsymbol{equal}{*}

  \begin{icmlauthorlist}
    \icmlauthor{Stefanos Bakirtzis}{yyy}
  \end{icmlauthorlist}

  \icmlaffiliation{yyy}{Department of Mathematics, ETH Zurich, Zurich, Switzerland}

  \icmlcorrespondingauthor{Stefanos Bakirtzis}{stefanos.bakirtzis@math.ethz.ch }

  \icmlkeywords{Finite Element Methods, Graph Neural Networks, Operator Learning, Scientific Machine Learning}

  \vskip 0.3in
]



\printAffiliationsAndNotice{}  
\begin{abstract}

Finite element methods (FEM) inherently induce graph structures that provide a natural interface to message passing. We show that the principal operations of FEM admit a common graph interpretation: local element interactions define messages, assembly becomes routing and aggregation, and sparse operator application and iterative solution can be realized through message passing on a mesh-induced graph. Building on this observation, we introduce \modelname, a graph-native framework that supports both classical numerical FEM simulations and physics-consistent neural operator learning within the same representation. For static PDEs, we demonstrate that \modelname is numerically equivalent to conventional FEM while reducing parts of the solution pipeline cost. More importantly, this graph-native representation enables a hybrid operator-learning formulation  for time-dependent problems, in which FEM-derived discrete operators define the graph connectivity and physics-consistent baseline dynamics, while learning is restricted to local
corrections of their edge interactions.   
In these settings, \modelname achieves accurate long-horizon rollouts and strong in- and out-of-distribution accuracy across three time-dependent PDE benchmarks, whilst, as shown for the wave equation, the learned operator transfers without fine-tuning across fine mesh resolutions and domain transformations, attaining substantial inference speedups over numerical simulation at large scale. 

\vspace{-4mm}
\end{abstract}

\section{Introduction}

\vspace{-2mm}

Partial differential equations (PDEs) model a broad range of physical systems \citep{Book_PDE, Applications_PDE}, but are generally intractable and thus solved via numerical approximations.
Among the existing numerical paradigms, the finite element method (FEM) is widely used due to its mathematical rigor and versatility \citep{PDE_FEM}. However, FEM can remain computationally expensive, particularly for high-dimensional problems or settings requiring repeated solver calls.
\  Consequently, learning-based solutions to PDEs have emerged as a promising route to overcome this core limitation \citep{brunton2024promising}. One paradigm is to deploy large-scale deep learning (DL) architectures that explicitly learn operators that map some input data to the target solution for different PDEs \citep{kovachki2023Operatorneural}. Hitherto, various such neural PDE solvers have been introduced,  employing a variety of architectures, including Fourier \citep{li2021fno}, convolutional \citep{long2018pde_net, Convolutional_Neural_Ops}, attention-based \citep{herde2024poseidon, wu2024transolver}, and message-passing operators \citep{Graph_Neural_Operators, pfaff2021meshgraphnets, brandstetter2022MPNN_PDE}. However, although  physical information can be incorporated through the model inputs or training objectives, typically, the structure of the neural solver itself is not directly informed by the discrete numerical operators arising from the underlying PDE problem. At the same time, such models require voluminous data to be effective, and it has been reported that their fidelity deteriorates in the low-data regime and out-of-distribution inference \citep{herde2024poseidon}.

Instead, physics-informed approaches integrate physical laws into the training process in a data-free manner, commonly through PDE residuals and boundary or initial conditions \citep{Dif_Eq_Karniadakis, karniadakis2021physics}. However, physics-informed neural networks (PINNs) can suffer from restrictive regularity requirements and ill-conditioned optimization \citep{wang2022and_PINNS_fail,
de2024error_PINNs}. More importantly, standard formulations typically learn a problem-specific solution rather than a transferable solution operator. An alternative is to intertwine numerical solvers, \eg FEM, with modern machine learning (ML) and automatic differentiation frameworks \citep{wen2026tensorgalerkin, xue2023jaxfem, bakirtzis2026solving}, such as PyTorch and JAX \citep{paszke2019pytorch, jax2018github}.  That enables endowing the solvers with differentiability and learning capacity, while at the same time benefiting from the hardware graphics processing unit (GPU) acceleration that ML frameworks proffer. However, even when harnessing GPU acceleration, these approaches remain computationally expensive, particularly for time-dependent PDEs and long-time rollouts that require the iterative solution of large sparse linear systems at every time step. In addition, although differentiability facilitates inverse problems, when it comes to designing learning-based solutions, these remain only loosely coupled with the differentiable FEM solver, \ie the FEM operators are used to construct a training loss or generate reference data, whereas the learned update itself is represented by a generic neural architecture \citep{wen2026tensorgalerkin}. Consequently, physical consistency is enforced only indirectly through training, rather than being guaranteed by the model architecture, and the resulting learned solution operator remains largely a black box.  Instead, hybrid approaches may embed learning directly within established numerical schemes, using data-driven components to accelerate or improve numerical PDE solutions \citep{mishra2018machine}.

A promising paradigm in this direction is to exploit the innate graph structures induced by finite element discretizations. This  connection has already motivated some  mesh-based neural simulators, FEM-derived graph models, and physics-informed graph architectures \citep{lienen2022fen,gao2022physics,
gladstone2024mesh_based_gnns,chenaud2024physics}.  However, these approaches exploit this insight only at selected stages, using the graph primarily as the support of a learned surrogate or to incorporate FEM-derived 
features and training objectives, rather than as a common computational representation of the numerical and neural pipeline itself. Hence, our work harnesses the intrinsic link between finite element computations and message passing on mesh-induced graphs to introduce a unified 
\ framework for classic numerical and hybrid neural PDE simulations, yielding the following contributions: 

\begin{itemize} [leftmargin=*, itemsep=2pt, topsep=2pt, parsep=0pt, partopsep=0pt]

\vspace{-2mm}

\item  \textbf{A complete finite element pipeline hinging on a native graph representation}. Specifically, we introduce \modelname, an end-to-end framework that expresses the principal stages of finite element computation within a common mesh-induced graph representation. Implemented in a GPU-compatible differentiable framework, \modelname  supports conventional numerical PDE simulations while
reducing the computational cost of selected stages of the FEM pipeline (see Appendix A).

    \item \textbf{We demonstrate that a graph-native FEM representation enables physics-consistent graph operator learning}, molding neural solvers whose architecture is directly informed by the governing  PDE and its finite element discretization. In particular, the graph inherits its connectivity from the FEM mesh and its baseline edge interactions from the discrete PDE operator. Consequently, the graph operator only needs to learn a transformation around physically meaningful edge features, and the PDE trajectory can be emulated via message passing.   We further employ a truncated Neumann-series approximation of the inverse mass matrix \cite{guermond2013correction_neumann}, which progressively recovers interactions between the graph nodes, providing a controllable accuracy–cost trade-off.

    \item   \textbf{Zero-shot mesh-resolution and domain extrapolation.} By reassembling the FEM operator on each target mesh while retaining the same learned local edge transformation, \modelname transfers without retraining, as demonstrated on the wave equation,  from a single training discretization to unseen meshes and domains of the same or finer resolution, while maintaining stable rollout accuracy and yielding substantial speed benefits.
    
 \end{itemize}

\vspace{-4mm}
\section{Related Work}

\vspace{-2mm}

\paragraph{Differentiable and accelerated FEM.} Modern FEM frameworks increasingly leverage high-level numerical and automatic-differentiation software suites to simplify assembly, exploit hardware acceleration, and interface numerical solvers with optimization and learning.
Scikit-fem \citep{gustafsson2020scikit_fem} provides a lightweight Python framework for finite element assembly, while JAX-FEM \citep{xue2023jaxfem} and torch-fem \citep{meyer_torchfem} implement differentiable FEM within JAX and PyTorch, respectively, allowing finite element simulations to benefit from GPU acceleration and automatic differentiation.  TensorGalerkin \citep{wen2026tensorgalerkin} further reformulates Galerkin assembly as tensorized local computations followed by deterministic sparse reductions, substantially reducing the overhead of conventional FEM implementations on GPUs.

\vspace{-4mm}

\paragraph{Neural and graph-based PDE simulation.}
Neural operators seek to learn mappings between function spaces, with representative architectures including DeepONet \citep{lu2021deeponet}, Fourier neural operators \citep{li2021fno, geo_fno}, convolutional operators \citep{Convolutional_Neural_Ops}, graph neural operators \citep{brandstetter2022MPNN_PDE}, or Radon operators \citep{lu2026solving_radon_operator}. Graph-based formulations are particularly natural for irregular geometries and unstructured discretizations: MeshGraphNets (MGN) \citep{pfaff2021meshgraphnets} and message-passing neural PDE
solvers \citep{brandstetter2022MPNN_PDE,  sanchez2020learning, bakirtzis2024towards} learn autoregressive dynamics through local graph interactions, while Finite Element Networks (FEN) \citep{lienen2022fen} derive continuous-time learned dynamics from finite element principles. More recent graph neural operators, such as RIGNO \citep{mousavi2026rigno}, introduce multiscale communication to support resolution- and geometry-aware operator learning. Closer to our setting, several works incorporate numerical or variational structure directly into graph models. Physics-informed graph neural Galerkin networks \citep{gao2022physics} combine graph convolutions with finite element discretizations and variational residuals, while physics-informed graph mesh
networks \citep{chenaud2024physics} couple graph architectures with finite element kernels.  Related approaches use FEM-derived mesh features \citep{gladstone2024mesh_based_gnns}, weak-form objectives within graph-based PDE solvers \citep{iparraguirre2026data}, or finite element weak forms for physics-informed learning \citep{yamazaki2025finite_pinn}.
Sparse FEM systems have also been represented as weighted graphs for learned linear solutions \citep{grementieri2022towards_neural_linear_sparse}, while \citep{li2025implementing_Gal_FEM_GNNT} implement an unstructured discontinuous Galerkin solver using graph and convolutional operations.  Graph-based reformulations of FEM have also been explored in computational mechanics, particularly for edge-based modeling of damage and fracture \cite{GraFEA}.

Building on the established connection between graphs and FEM, \modelname takes a complementary step by unifying FEM assembly, numerical solution, and operator learning within a common graph representation, enabling both conventional and hybrid neural FEM simulations.


 
\vspace{-4mm}

\section{Background}

\label{sec:background}

\subsection{The Finite Element Method}
\label{sec:background_fem}
\vspace{-2mm}

Let $\Omega \subset \mathbb{R}^{d}$ be a bounded spatial domain with boundary $\partial\Omega$. We consider a PDE, subject to appropriate boundary conditions and, for time-dependent problems, initial
conditions, $u_0$, of the abstract form:

\vspace{-6mm}

\begin{equation}
\mathcal{F}\bigl(\boldsymbol{x},t,u,\nabla u,\nabla^2u,\ldots;\boldsymbol{\mu}\bigr)
=
f\bigl(\boldsymbol{x},t;\boldsymbol{\mu}\bigr),
\qquad
\boldsymbol{x}\in\Omega,
\label{eq:PDE_equation}
\end{equation}

\vspace{-2.5mm}

where $u \in V$  denotes the unknown state variable ($V$ here is a suitable Banach space), $\mathcal{F}$   is the differential operator governing the physical system exhibiting dependence on the state $u$ and its spatial or temporal derivatives, whilst $f$ represents source or forcing terms and $\boldsymbol{\mu}$ contains physical or geometric parameters (inputs of the problem). For time-independent problems, a weak form of (\ref{eq:PDE_equation})  can be derived by multiplying the PDE by an admissible test function, $v \in W$, integrating over the domain, and applying integration by parts to reduce the order of the spatial derivatives. The resulting variational problem entails finding $u\in V$ such that \citep{Book_PDE}:

\vspace{-4mm}
\begin{equation}
a(u,v;\boldsymbol{\mu})
=
\ell(v;\boldsymbol{\mu}),
\qquad
\forall v\in W,
\label{eq:weak_form}
\end{equation}
\vspace{-6mm}

where   \(a(\cdot,\cdot;\boldsymbol{\mu})\) is a bilinear form portraying the governing physics, and \(\ell(\cdot;\boldsymbol{\mu})\) is a linear functional corresponding to the external forcing. Then,   let $\mathcal{T}_h$ be a finite element mesh that partitions $\Omega$ into a
collection of $E$ non-overlapping elements. We introduce a finite-dimensional subspace, $V_h  \subset V$, associated with $\mathcal{T}_h$,   used to approximate both the infinite-dimensional spaces $V$ and $W$,  and we define a basis $\{\varphi_i\}_{i=1}^{N}$ of $V_h$, where each basis function has local support over only a small number of neighboring elements. The finite element approximation $u_h \in V_h$ of the exact solution, $u$, is then expressed in terms of the ansatz:

\vspace{-6mm}

\begin{equation}
    u_h(\boldsymbol{x})
    =
    \sum_{j=1}^{N}
    U_j \varphi_j(\boldsymbol{x}),
    \label{eq:fem_ansatz}
\end{equation}

\vspace{-3mm}

where $ \boldsymbol{U}= [U_1,\ldots,U_N]^{\mathsf{T}} \in \mathbb{R}^{N}  $ is the vector of unknown finite element coefficients.  Applying the Galerkin method, we choose each basis function as a test function. Hence,  substituting (\ref{eq:fem_ansatz}) into the weak formulation (\ref{eq:weak_form}), yields: 
$ \sum_{j=1}^{N}
a\bigl(
\varphi_j,\varphi_i;
\boldsymbol{\mu}
\bigr)U_j
=
\ell\bigl(
\varphi_i;
\boldsymbol{\mu}
\bigr), \
i=1,\ldots,N$, producing one equation for each test basis function and, consequently, a system of $N$ algebraic equations for the $N$ unknown coefficients in $\boldsymbol{U}$,   written compactly as:

\vspace{-5mm}

\begin{equation}
\boldsymbol{K}  
\boldsymbol{U}
=
\boldsymbol{F}
\label{eq:fem_linear_system}
\end{equation}

\vspace{-3mm}

where the entries of the system matrix, $\boldsymbol{K}$, and the right-hand-side (RHS) vector, $ \boldsymbol{F}$, are defined as $ K_{ij}  = a\bigl( \varphi_j,\varphi_i; \boldsymbol{\mu} \bigr)$ and $  F_i  = \ell\bigl(
\varphi_i; \boldsymbol{\mu} \bigr) $, respectively, and correspond to the Galerkin stiffness matrix and load vector, accordingly (note that in (\ref{eq:fem_linear_system}) the dependence on $\boldsymbol{\mu}$ has been suppressed for notational simplicity). For  elements with $k$ local degrees of freedom (DoFs), the local element Galerkin matrix, $K_e \in  \mathbb{R}^{k \times k} $, is computed numerically via a  quadrature:

\vspace{-5mm}

\begin{equation}
(K_e)_{ab}
=
a_e
\left(
\phi_b^{(e)},
\phi_a^{(e)}
\right)
\approx
\sum_{q=1}^{N_Q}
\hat w_q
I_{ab}\!\left(\hat{\boldsymbol{\xi}}_q\right),
\label{eq:local_stiffness_matrix}
\end{equation}

\vspace{-3mm}

where $I_{ab}(\hat{\boldsymbol{\xi}}_q)$ denotes the value of the element integrand
evaluated at the quadrature point $\hat{\boldsymbol{\xi}}_q$ in the local element coordinate system, $\{\phi_a^{(e)}\}_{a=1}^{k}$ are the local basis functions for element $e$, $N_Q$ is the number of quadrature points, and
$\{\hat{w}_q\}_{q=1}^{Q}$ are the associated quadrature weights. Once the local matrices are computed, the global Galerkin matrix is assembled by fusing the local contributions. Formally, let \(g_e:\{1,\ldots,k\}\rightarrow\{1,\ldots,N\}\) denote the mapping from local to global DoFs. The global stiffness matrix is assembled by accumulating the element contributions according to the local-to-global mapping, \(K_{g_e(a),\,g_e(b)} \leftarrow K_{g_e(a),\,g_e(b)} + (K_e)_{ab}\),
for each pair of local degrees of freedom \(a,b=1,\ldots,k\) and every element \(e=1,\ldots,E\). Analogously, one can estimate $\boldsymbol{F}$. Typically, both processes involve the sequential computation of the local contributions over all elements and the subsequent accumulation to the global features.   With $\boldsymbol{K} $ and $\boldsymbol{F}$ assembled, the linear system of (\ref{eq:fem_linear_system}) can be solved by invoking classical linear algebra iterative solvers \citep{saad2003iterative}, \eg conjugate gradient (CG) or biconjugate gradient stabilized (BiCGSTAB).

\vspace{-2mm}

\subsection{General time-dependent finite element formulation}
\label{sec:background_time}

\vspace{-2mm}

To extend the preceding spatial finite element formulation to time-dependent problems, the Method of Lines is typically employed, whereby the spatial variables are discretized by the finite element method while time is left continuous \citep{hughes2003finite_timedependdant,brenner2008fem}. Applying the Galerkin procedure to the spatial operators yields the following semi-discrete variational formulation:

\vspace{-5mm}
\begin{equation}
\left\langle
\partial_t u,v
\right\rangle_M
+
a\!\left(
u,v;\boldsymbol{\mu}
\right)
+
\mathcal{N}l
\!\left(
u;v,\boldsymbol{\mu}
\right)
=
\ell\!\left(
v;\boldsymbol{\mu}
\right),
\qquad
\forall v\in V,
\label{eq:semidiscrete_variational}
\end{equation}
\vspace{-5mm}

where $\langle\cdot,\cdot\rangle_M$ denotes the mass inner product, and 
 $\mathcal Nl(\cdot;\cdot)$ is a semilinear form accounting for state-dependent nonlinear effects. Substituting the finite element approximation (\ref{eq:fem_ansatz}) into (\ref{eq:semidiscrete_variational}) yields the system of ordinary differential equations which provides a unified description for a broad class of transient first-order finite element models \citep{hughes2003finite_timedependdant}: 

\vspace{-4mm}
\begin{equation}
\mathbf{M}
\dot{\mathbf U}(t)
+
\mathbf{K}
\mathbf U(t)
+
\mathbf F_{\mathrm{nl}}
\!\left(
\mathbf U(t)
\right)
=
\mathbf F(t),
\label{eq:semidiscrete_ode}
\end{equation}
\vspace{-5mm}

where $\mathbf M$ is the finite element mass matrix  and $\mathbf F_{\mathrm{nl}}$ captures the nonlinear  contributions. Introducing a finite-difference approximation in time and applying a forward Euler scheme with an appropriate timestep size, $\Delta t$,  one can derive a recursive solution that expresses the DoF values at the next time step, $\mathbf U^{n+1}$, as:

\vspace{-6mm}

\begin{equation}
\mathbf M\mathbf U^{n+1}
=
(\mathbf M
-
\Delta t\,\mathbf K)\mathbf U^{n}
-
\Delta t\,\mathbf F_{\mathrm{nl}}(\mathbf U^{n})
+
\Delta t\,\mathbf F^{n},
\label{eq:generic_explicit_update}
\end{equation}

\vspace{-2mm}

which recursively advances the solution from time step $n$ to $n+1$. 
Note that if the PDE included a second-order derivative with respect to time, $ 
\ddot{\mathbf U}(t)$,  one would have to employ a second-order finite difference scheme, which would induce at the RHS an additional term depending on $\mathbf{U}^{n-1}$.


\vspace{-2mm}

 \section{Graph-Native FEM and Learnable Sparse Time Stepping}
\label{sec:gem}
\vspace{-2mm}

\begin{figure*}[!t]
\centering
    {\includegraphics[width = 0.850\linewidth]{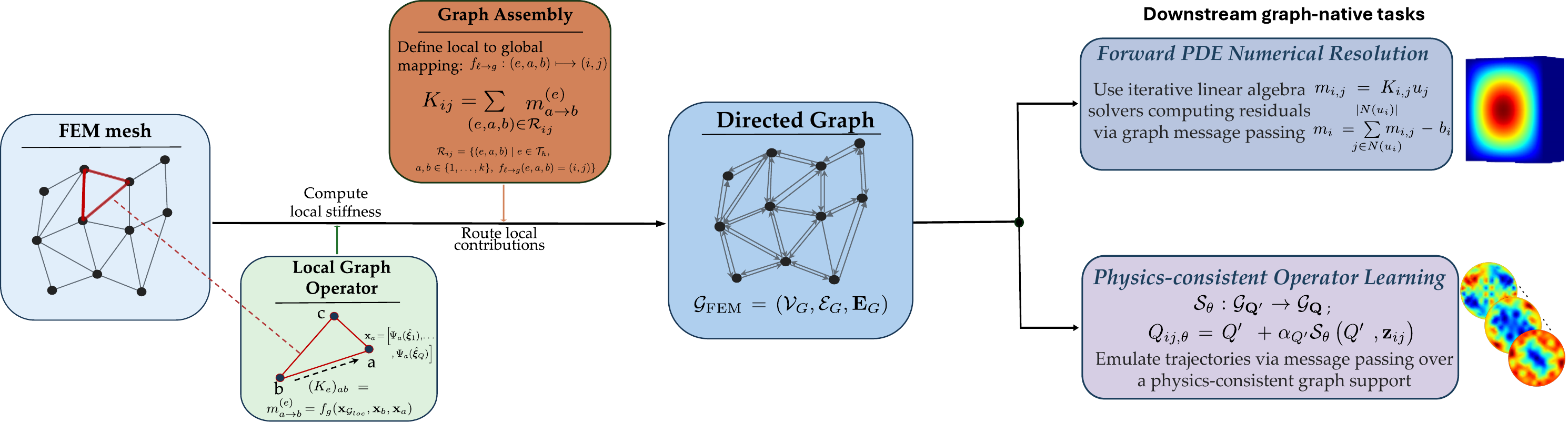}} 
    \vspace{-0.2cm}
\caption{ Overview of  \modelname: Finite element computations are represented as message passing on the mesh-induced graph, enabling graph-native numerical simulation and operator learning}
\vspace{-4mm}
\label{fig:General_Pipeline}
\end{figure*}

As discussed in \citep{wen2026tensorgalerkin}, the assembly 
stages (local element contribution computation and routing) can be reformulated as dense tensor contractions followed by deterministic sparse projections, enabling an efficient GPU-parallel implementation within modern automatic-differentiation frameworks. Our work builds upon this observation and software implementation to demonstrate that these same operations admit an exact graph interpretation, thereby enabling the expression of finite element assembly, sparse operator application, iterative solution, as well as operator learning, within a unified graph-native framework, depicted in Figure~\ref{fig:General_Pipeline}.  


\subsection{Graph-driven finite element assembly}
\label{sec:graph_assembly}

\vspace{-2mm}

The quadrature integrand in (\ref{eq:local_stiffness_matrix}) depends on  basis function-related quantities of the underlying PDE  and on the geometric mapping from the reference to the physical element, \ie $\mathcal{I}_{ab} (\hat{\boldsymbol{\xi}}_q)= \mathcal{I}\!\left( \Psi_a(\hat{\boldsymbol{\xi}}_q),
\Psi_b(\hat{\boldsymbol{\xi}}_q), J(\hat{\boldsymbol{\xi}}_q),|J(\hat{\boldsymbol{\xi}}_q)|, \boldsymbol{\mu} \right)$, where $J(\hat{\boldsymbol{\xi}}_q)$ represents the Jacobian matrix of the reference-to-physical mapping evaluated at quadrature point $\hat{\boldsymbol{\xi}}_q$ (used to transform derivatives), and $|J(\hat{\boldsymbol{\xi}}_q)|$ its determinant. Furthermore, $\Psi_a(\hat{\boldsymbol{\xi}}_q)$ denotes the collection of quantities related to the basis function, associated with the local DoF $a$, and evaluated at $\hat{\boldsymbol{\xi}}_q$. Depending on the problem, $\Psi_a$ may contain the basis function itself, first- or higher-order  spatial derivatives, and other quantities derived from the finite element basis. 

Let $\mathcal{G}_{loc}$ be a fully connected directed graph, including self-loops, with $k$ nodes, representing $k$ finite element DoFs, effectively resembling the topology of an element $e \in \mathcal{T}_h$. We endow each node $a$ with a feature vector $\mathbf{x}_a = \left[ \Psi_a(\hat{\boldsymbol{\xi}}_1), \Psi_a(\hat{\boldsymbol{\xi}}_2), \ldots, \Psi_a(\hat{\boldsymbol{\xi}}_Q) \right]$, which collects the basis-dependent quantities required to evaluate the element quadrature. We also assign a local feature vector with geometry-related quantities per graph element,  
$\mathbf{g}_{loc} = \left[ J(\hat{\boldsymbol{\xi}}_1), |J(\hat{\boldsymbol{\xi}}_1)|, \ldots, J(\hat{\boldsymbol{\xi}}_Q), |J(\hat{\boldsymbol{\xi}}_Q)|, \boldsymbol{\mu}\right]$. Then, for every PDE  admitting the variational formulation of (\ref{eq:weak_form}), the finite element quadrature kernel induces a deterministic message function, $f_g$, between the local graph-element feature vectors of $\mathcal{G}_{loc}$  such that the message received at node $a$ from node $b$ is the local stiffness entry, \ie $m_{b\to a} = f_g(\mathbf{g}_{loc}, \mathbf{x}_a, \mathbf{x}_b) = (K_e)_{ab}$.

\vspace{-2mm}

  For example, consider the Poisson equation, whose bilinear form is $a(u,v)=\int_{\Omega}\nabla_{\boldsymbol{x}}u\cdot \nabla_{\boldsymbol{x}}v\,\mathrm d\Omega$, where
$\nabla_{\boldsymbol{x}}$ denotes differentiation with respect to the physical coordinates. For each local DoF $a$, we define the node feature vector $\mathbf{x}_a= \left[\nabla_{\boldsymbol{x}}\phi_a^{(e)}(\hat{\boldsymbol{\xi}}_1),\ldots, \nabla_{\boldsymbol{x}}\phi_a^{(e)}(\hat{\boldsymbol{\xi}}_Q)
\right]$. The element-level geometric feature vector is $ \mathbf g_{loc} = \left[ |\det\mathbf J_e(\hat{\boldsymbol{\xi}}_1)|, \ldots, |\det\mathbf J_e(\hat{\boldsymbol{\xi}}_Q)| \right]$. The corresponding deterministic graph operator is defined as
$f_g(\mathbf g_{loc},\mathbf x_a,\mathbf x_b) = \hat{\mathbf w}^{\mathsf T} \left( \mathbf g_{loc}
\odot \left[ \mathbf x_a^{(q)\mathsf T} \mathbf x_b^{(q)} \right]_{q=1}^{Q} \right)$, where
$\hat{\mathbf w} = \left[ \hat w_1,\ldots,\hat w_Q \right]^{\mathsf T}$ and $\odot$ denotes the Hadamard (element-wise) product, and the message exchanged between nodes a and b reenacts the (a, b) entry of the local Poisson stiffness matrix.

Hence, the local stiffness matrix comprises the ensemble of all pairwise messages exchanged between the graph elements in $\mathcal{G}_{loc}$. Since all elements of $\mathcal{T}_h$ are topologically identical in the reference domain, the local matrices for all elements of the mesh can be computed simultaneously by applying the same deterministic graph operator to a batch of local graphs with different node
and element features. Subsequently, the local messages are routed to the corresponding global graph edges through the standard finite element local-to-global mapping,  $ f_{\ell\rightarrow g} $. This mapping depends solely on the mesh connectivity and is precomputed once as an inverse routing map that associates every local edge $(e,a,b)$  with a unique edge ($i, j$) in the global graph, \ie $ f_{\ell\rightarrow g} : (e,a,b) \longmapsto (i,j) $.  Global assembly is then performed through a scatter-add reduction over this routing map, 
 accumulating contributions from adjacent elements into the corresponding edge weights,
 \ie  $ K_{ij} = \sum_{ (e,a,b)\in\mathcal R_{ij} } m_{b\rightarrow a}^{(e)},$ where $\mathcal R_{ij} = \bigl\{ (e,a,b) \bigm| e\in\mathcal T_h,\allowbreak\; a,b\in\{1,\ldots,k\},\allowbreak\; f_{\ell\rightarrow g}(e,a,b)=(i,j) \bigr\}$  comprises all local edges mapped to $(i,j)$. Ultimately, the assembled operator is obtained directly as a weighted graph $ \mathcal G_{\mathrm{FEM}} = \left( \mathcal V_G, \mathcal E_G, \mathbf E_G \right)$,  where $\mathcal E_G$ conveys the sparsity pattern of the $\mathcal V_G$ nodes induced by the finite element discretization, and  $ \mathbf E_G = \left\{ K_{ij} \right\}_{(i,j)\in\mathcal E_G}$ comprises the edge weights/assembled stiffness.

\vspace{-2mm}

\subsection{Graph-driven finite element system solution}
\label{sec:graph_system_solution}
\vspace{-2mm}

The methodology of the previous section  enables expressing and structuring the stiffness matrix as a directed graph,  $\mathcal G_{\mathrm{FEM}}$, consisting of the finite element DoFs.  The linear system of (\ref{eq:fem_linear_system}) can itself assume a   \textit{graph-driven} formulation, where the unknown solution and forcing terms constitute the node features,  whilst the $K$ values are the edge features.  Let $N(v_i) = \{ v_{j} \in \mathcal{V}_G | (i, j) \in \mathcal{E}_G\}$ denote the neighborhood of node $v_i$, \ie all nodes $v_{j}$ with which $v_i$ interacts  via an active graph edge $(i, j)$. Under this representation, considering that every graph node $i$ receives messages, $m_{i,j} = K_{i,j}u_j $, from all its first-hop neighbors, the accumulated message $m_i = \sum_{j\in \mathcal{N}(v_i)} m_{i,j} - b_i$ represents the residual of the $i$-th linear equation of  (\ref{eq:fem_linear_system}). Therefore, under the graph representation of the assembled FEM operator, message passing between nodes is algebraically equivalent to the classical sparse matrix--vector product and introduces no additional approximation. Thence, iterative methods requiring consecutive residual evaluations, such as CG or BiCGSTAB, can be expressed as sequences of graph-operator applications, yielding the finite element system solution through a graph-driven representation while remaining numerically equivalent to the corresponding classical sparse solver. Note that representing sparse linear systems as weighted graphs has  been considered for learned approximate solvers \citep{grementieri2022towards_neural_linear_sparse}. However, here, the graph operator arises directly from the preceding finite element assembly and remains the native representation used for subsequent numerical operations.

\vspace{-2mm}

\subsection{Physics-Consistent Operator Learning for Time-Dependent PDEs}
\label{sec:learned_sparse_operators}

Considering the intrinsic link between graphs, message passing, and FEM, a \textit{graph-driven} approach provides a natural representation for operator learning for time-dependent PDEs.  Inspecting (\ref{eq:generic_explicit_update}), one can express a broad class of explicit time-stepping schemes through the following general polynomial operator approximation: 

\vspace{-6mm}

{\small
\begin{equation}
\mathbf U^{n+1}
=
\mathbf A\mathbf U^{n}
+
\mathbf B\mathbf U^{n-1}
+
\mathbf C
\left(\mathbf U^{n}\right)^{\odot 2}
+
\mathbf D
\left(\mathbf U^{n}\right)^{\odot 3}
+
\mathbf E
\mathbf F^n
\label{eq:general_polynomial_update}
\end{equation} }
\vspace{-5mm}

where $\mathbf A, \ \mathbf B, \ \mathbf C, \ \mathbf D, \ \mathbf E$ are coefficient matrices determined by the governing PDE and the selected temporal discretization, whilst $\left(\mathbf U\right)^{\odot p}$ denotes the element-wise power of the
solution vector, \ie $\left(\mathbf U\right)^{\odot p} = [U_1^p,\ldots, U_N^p]^{\mathsf T}$.  

Note that this formulation constitutes only a polynomial approximation of the underlying discrete update  (higher-order powers can be included analogously), since the nonlinear finite element contribution, $\mathbf{F}_{nl}(U(t))$, does not, in general, decompose into operators acting on component-wise powers. Further, this update scheme  natively decomposes into one equation per DoF, which assumes an explicit message passing  analogy:

\vspace{-6mm}

{\small
\begin{equation}
\begin{aligned}
U_i^{n+1}
={}&
\sum_{j\in\mathcal N_A(i)} A_{ij}U_j^n
+\sum_{j\in\mathcal N_B(i)} B_{ij}U_j^{n-1} +
\\[-1mm]
&\hspace{-4.2em}
\sum_{j\in\mathcal N_C(i)} C_{ij}(U_j^n)^2
+\sum_{j\in\mathcal N_D(i)} D_{ij}(U_j^n)^3
+\sum_{j\in\mathcal N_E(i)} E_{ij}F_j^n,
\end{aligned}
\label{eq:general_node_update_graphs}
\end{equation}
}

where $ \mathcal N_Q(i) = \left\{j \;\middle|\; Q_{ij}\neq0 \right\}, \mathbf Q\in\{\mathbf A,\mathbf B,\mathbf C,\mathbf D,\mathbf E\}$ denotes the neighborhood induced by the nonzero entries of the corresponding coefficient matrix. Note that by comparing (\ref{eq:general_node_update_graphs}) with (\ref{eq:generic_explicit_update}), these matrices generally absorb the action of the inverse mass matrix, $\mathbf{M}^{-1}$, and are therefore not guaranteed to preserve the sparsity of the original finite element matrices. More importantly, for large-scale problems, explicitly forming or repeatedly applying $\mathbf{M}^{-1}$ becomes computationally prohibitive. A common remedy is to replace the consistent mass matrix by its row-sum lumped approximation, $\mathbf{M}_{\mathrm{lump}}=\operatorname{diag}(\mathbf{M}\mathbf{1})$, with $\mathbf{1}$ the vector of all ones, \ie $M_{\mathrm{lump}_{ii}}=\sum_{j=1}^{N}M_{ij}$, whose inverse can be applied through inexpensive node-wise divisions rather than a sparse linear solve. Consequently, this yields approximate coefficient matrices, $\mathbf{A}',\mathbf{B}',\mathbf{C}',\mathbf{D}', \mathbf{E}'$, which are sparse and can be used to evaluate (\ref{eq:general_node_update_graphs}). However, this approximation alters the underlying finite element operator and introduces additional numerical error that may degrade the accuracy of long rollouts for transient simulations, as demonstrated in Section~\ref{sec:Results_Op_learning}.


This observation motivates the proposed \modelname, which learns sparse graph operator corrections around physics-consistent baseline edge weights. Formally, for each coefficient matrix $\mathbf Q\in\{\mathbf A,\mathbf B,\mathbf C,\mathbf D,\mathbf E\}$,   we construct a
physics-consistent baseline approximation $\mathbf Q'$ and its equivalent
weighted directed graph representation $\mathcal G_{\mathbf Q'} = \left( \mathcal V, \mathcal E_{\mathbf Q'} \right)$, where
 $\mathcal E_{\mathbf Q'} = \{(i,j)\mid Q'_{ij}\neq0\}$ denotes the induced graph support, \ie the graph connectivity structure stemming from the nonzero entries of the baseline finite element coefficient matrix.  Rather than learning a dense matrix-to-matrix mapping, we then seek a graph operator   $ \mathcal S_\theta:\mathcal G_{\mathbf Q'}\rightarrow\mathcal G_{\mathbf Q}$, which acts locally on every edge of the FEM graph. Specifically, for each edge $(i,j) \in \mathcal{E_{\mathbf Q'}}$ and the respective weight $Q_{i,j}'$, the corrected edge weight is computed as  $    Q_{ij,\theta}  =    Q'_{ij}   +    \alpha_{Q'} \mathcal S_{\theta} \left(   Q'_{ij},  \mathbf z_{ij} \right)$, where $\mathbf z_{ij}$ denotes static edge and nodal features independent of the evolving solution state. For instance, for the wave equation, the finite element-consistent coefficient  matrix is  $ \mathbf A' =2\mathbf I  -  c^2\Delta t^2 \mathbf M_{\mathrm{lump}}^{-1}\mathbf K,\label{eq:wave_baseline_operator}$ whose graph representation, $\mathcal{G}_{A'}$ stores the nonzero entries $A'_{ij}$ as edge weights. The operator  $ \mathcal S_\theta:\mathcal G_{\mathbf A'}\rightarrow\mathcal G_{\mathbf A}$ aims at transforming the edge weights $A_{i,j}'$ into the corrected weights $A_{i,j}$.  Consequently, with the graph edge weights transformed, the complete solution trajectory is generated auto-regressively by iteratively evaluating (\ref{eq:general_node_update_graphs}) via graph message passing.  Unlike neural PDE architectures whose graph topology is constructed from generic geometric neighborhoods or \(k\)-nearest neighbors, \modelname preserves the connectivity induced by the underlying finite element mesh. Consequently, operator learning is constrained to physically meaningful interactions, stemming from the numerical discretization, with corrections learned only over the existing FEM graph support (optionally augmented with self-loops or a limited $r$-hop extension) and the respective FEM matrix values.

\vspace{-2mm}

\subsection{Neumann-Series   Operators}
\label{sec:neumann}

\vspace{-2mm}

The baseline coefficient matrices of Section~\ref{sec:learned_sparse_operators} replace the consistent inverse mass matrix, $\mathbf{M}^{-1}$, with its lumped counterpart, $\mathbf{M}_{\mathrm{lump}}^{-1}$, thereby preserving the sparsity of the finite element graph at the cost of an approximation error that the learned correction, $\mathcal{S}_\theta$, has to compensate for. However, $\mathbf{M}^{-1}$ is in general dense, whereas $\mathcal{S}_\theta$ acts only on the (one-hop extended) support of the baseline operator. Instead of further expanding the graph support of the learned operator, following  \citep{guermond2013correction_neumann},  one can express the inverse mass matrix through a Neumann expansion
$\mathbf{M}^{-1} = (\mathbf{I}-\mathbf{S})^{-1}\mathbf{M}_{\mathrm{lump}}^{-1}
= \sum_{k=0}^{\infty} \mathbf{S}^{k}\,\mathbf{M}_{\mathrm{lump}}^{-1}$ ,  where $\mathbf{S} = \mathbf{I} - \mathbf{M}_{\mathrm{lump}}^{-1}\mathbf{M}$,  and each power, $k$, fills in another ring of neighbors. This expansion converges when the spectral radius of $\mathbf{S}$ is smaller than one, \ie when all eigenvalues of $\mathbf{M}_{\mathrm{lump}}^{-1}\mathbf{M}$ lie in $(0,2)$,  which holds whenever the mass matrix is symmetric positive definite with non-negative entries. 



Consequently, a truncated inverse can be used in place of the lumped mass matrix, $ \mathbf{M}_p^{-1} = \sum_{k=0}^{p} \mathbf{S}^{k}\mathbf{M}_{\mathrm{lump}}^{-1}$, where $p=0$ recovers the lumped approximation of Section~\ref{sec:learned_sparse_operators} and $p \to \infty$ the inverse of the consistent mass matrix, resulting in the approximate baseline coefficient matrices
$Q_p'\in\{\ \mathbf{A}'_p, \mathbf{B}'_p, \mathbf{C}'_p, \mathbf{D}'_p, \mathbf{E}'_p\}$. 
Importantly, neither $\mathbf{M}_p^{-1}$ nor the powers $\mathbf{S}^k$ need to be formed explicitly. In particular, $\mathbf{M}_p^{-1}$ can be applied \emph{matrix-free}, through $p$ consecutive rounds of message passing on the mass-matrix graph (see Appendix~\ref{sec:Appendix_Operator_Learning_Model_Details}), at the cost of $p$ additional sparse matrix--vector products per time step. Alternatively, since $\mathbf{S}$ and the coefficient matrices are fixed, the complete update can be \emph{precomputed} once, before the rollout, as a single sparse matrix on the $(p{+}1)$-hop support, so that each time step requires one sparse product irrespective of $p$. In both cases the receptive field grows by $p$ hops. 
The two options are compared in Appendix~\ref{sec:Appendix_Add_Res_Neuman}, and unless stated otherwise, we report results with the precomputed operator.

 \begin{table*}[t]
\centering
\caption{Operator-learning accuracy and computational cost. Relative $L_2$ trajectory errors (mean $\pm$ std.\ over the held-out test trajectories) and rollout time per trajectory, measured on one trajectory at a time on an A100 GPU.}
\label{table:operator_results}
\vspace{-3mm}

\scriptsize
\setlength{\tabcolsep}{2.5pt}
\renewcommand{\arraystretch}{0.92}

\resizebox{0.98\textwidth}{!}{%
\begin{tabular}{lllccccccccc}
\toprule
\multirow{2}{*}{PDE} & \multirow{2}{*}{Dataset} & \multirow{2}{*}{Metric} & \multicolumn{9}{c}{Method} \\
\cmidrule(lr){4-12}
& & & TensorPILS & \modelname & MGN & FEN & RIGNO & DeepONet & Transolver & Geo-FNO & Lumped \\
\midrule

\multirow{4}{*}{AC}
& \multirow{2}{*}{Explicit}
& ID $\downarrow$
& $0.031{\pm}0.003$ & \textbf{0.019{$\pm$}0.004} & $0.034{\pm}0.004$ & $0.041{\pm}0.007$ & $0.100{\pm}0.022$ & $0.597{\pm}0.051$ & $0.086{\pm}0.021$ & $0.063{\pm}0.006$ & $0.030{\pm}0.004$ \\
& & OOD $\downarrow$
& $0.049{\pm}0.005$ & \textbf{0.026{$\pm$}0.006} & $0.053{\pm}0.009$ & $0.047{\pm}0.007$ & $0.191{\pm}0.040$ & $0.557{\pm}0.066$ & $0.217{\pm}0.057$ & $0.118{\pm}0.015$ & $0.041{\pm}0.005$ \\
& \multirow{2}{*}{Implicit}
& ID $\downarrow$
& $0.054{\pm}0.010$ & \textbf{0.012{$\pm$}0.002} & $0.051{\pm}0.021$ & $0.060{\pm}0.022$ & $0.220{\pm}0.062$ & $0.462{\pm}0.060$ & $0.243{\pm}0.041$ & $0.115{\pm}0.015$ & $0.024{\pm}0.004$ \\
& & OOD $\downarrow$
& $0.083{\pm}0.022$ & \textbf{0.018{$\pm$}0.005} & $0.117{\pm}0.072$ & $0.063{\pm}0.027$ & $0.293{\pm}0.133$ & $0.799{\pm}0.143$ & $0.557{\pm}0.167$ & $0.215{\pm}0.048$ & $0.022{\pm}0.006$ \\
\midrule

\multirow{2}{*}{Wave}
& \multirow{2}{*}{Explicit}
& ID $\downarrow$
& $0.080{\pm}0.006$ & $0.047{\pm}0.005$ & \textbf{0.006{$\pm$}0.001} & $0.153{\pm}0.021$ & $0.271{\pm}0.035$ & $0.531{\pm}0.028$ & $0.407{\pm}0.026$ & $0.175{\pm}0.021$ & $0.191{\pm}0.012$ \\
& & OOD $\downarrow$
& $0.100{\pm}0.007$ & $0.053{\pm}0.005$ & \textbf{0.010{$\pm$}0.001} & $0.185{\pm}0.013$ & $0.448{\pm}0.034$ & $0.469{\pm}0.049$ & $0.477{\pm}0.030$ & $0.453{\pm}0.033$ & $0.234{\pm}0.017$ \\
\midrule

\multirow{2}{*}{AD}
& \multirow{2}{*}{Explicit}
& ID $\downarrow$
& $0.113{\pm}0.030$ & $0.031{\pm}0.004$ & \textbf{0.024{$\pm$}0.006} & $0.066{\pm}0.006$ & $0.339{\pm}0.033$ & $0.801{\pm}0.146$ & $0.591{\pm}0.086$ & $0.224{\pm}0.039$ & $0.080{\pm}0.012$ \\
& & OOD $\downarrow$
& $0.187{\pm}0.088$ & \textbf{0.036{$\pm$}0.011} & $0.063{\pm}0.028$ & $0.079{\pm}0.017$ & $0.458{\pm}0.172$ & $1.527{\pm}0.649$ & $1.118{\pm}0.404$ & $0.576{\pm}0.229$ & $0.040{\pm}0.016$ \\
\midrule

\multirow{3}{*}{\shortstack[l]{Rollout\\time (ms)}}
& & AC $\downarrow$
& $248.0$ & $21.8$ & $2731.8$ & $1209.9$ & $331.2$ & \textbf{3.3} & $1955.1$ & $1933.6$ & $22.0$ \\
& & Wave $\downarrow$
& $229.6$ & $9.5$ & $2809.2$ & $1357.8$ & $380.6$ & \textbf{4.7} & $1952.4$ & $1952.1$ & $9.4$ \\
& & AD $\downarrow$
& $328.4$ & $6.2$ & $2841.6$ & $2399.0$ & $333.7$ & \textbf{3.7} & $1974.8$ & $1942.9$ & $6.2$ \\
\bottomrule
\end{tabular}%
}
\vspace{-4mm}
\end{table*}

\vspace{-4mm}

\section{Results}

\vspace{-3mm}

In the following subsections, we exemplify the potential of \modelname to serve as a neural PDE solver, whilst  Appendix~\ref{sec:Appendix_Num_PDE_Solver} instantiates its capacity to also act as a conventional numerical solver.  For the implementation of \modelname, we build upon the existing \textsc{TensorMesh} (TM) software package \citep{wen2026tensorgalerkin}, 
replacing the local stiffness, mass, and load computations, the assembly process, as well as the linear system solving with the equivalent \textit{graph-driven} formulation described in Sections~\ref{sec:graph_assembly} and~\ref{sec:graph_system_solution}. All runtimes were measured on an NVIDIA A100 40 GB GPU.

\vspace{-2mm}

\subsection{Graph-driven Operator Learning} \label{sec:Results_Op_learning}
\vspace{-2mm}

\textbf{Benchmarks.} We evaluate \modelname against representative graph-based and neural operator learning approaches, including: ($i$) DeepONet, which combines branch and trunk networks to learn mappings between functions \citep{lu2021deeponet}, ($ii$) MGN, an autoregressive graph simulator  \citep{pfaff2021meshgraphnets}, ($iii$) FEN which derives a continuous-time graph model from finite element principles for spatio-temporal forecasting \citep{lienen2022fen}, ($iv$) TensorPILS, a physics-informed Galerkin approach based on finite element assembly \citep{wen2026tensorgalerkin}, ($v$) and RIGNO, a multiscale graph neural operator that communicates through a coarsened regional graph and targets arbitrary domains and resolutions \citep{mousavi2026rigno}, ($vi$) Transolver, an attention-based neural PDE solver that aggregates irregular-mesh features through learned Physics-Attention slices \citep{wu2024transolver}, ($vii$) and Geo-FNO, a geometry-aware Fourier neural operator applied autoregressively that maps irregular physical domains to a latent regular grid for spectral operator learning \citep{geo_fno}. Additional details on the model implementation can be found in  Appendix~\ref{sec:Appendix_Operator_Learning_Model_Details}.

\vspace{-2mm}

\textbf{Datasets.} We consider three time-dependent PDEs:   $(a)$ the parabolic Allen--Cahn (AC) equation, $(b)$ the hyperbolic wave equation, and $(c)$ the parabolic advection--diffusion (AD) equation. For each equation, we generate 32 independent trajectories, each comprising 400 time steps, from random initial conditions. The general form of (\ref{eq:generic_explicit_update}) is derived via an $\textit{Explicit}$ scheme for the three PDEs, whilst for AC we also use an $\textit{Implicit}$ scheme, as detailed in Appendix~\ref{sec:Appendix_Operator_Learning_Setup}. For each case, out of the total 32 trajectories, 20 are used for training, 4 are employed in validation to select the best-performing model, whilst 8 unseen trajectories are held out to probe the performance of each model.  Further, similarly to \citep{wen2026tensorgalerkin}, we consider  two test settings: an In-Distribution (ID) and an Out-of-
Distribution (OOD) setting. Specifically, we split each trajectory into two consecutive segments: the first 200 steps are treated as the ID regime and utilized as the training horizon, whilst the last 200 steps are treated as the OOD regime, as they evaluate longer-horizon rollout behavior beyond the training horizon. Hence,  for the 20 training trajectories, only the first 200 steps are seen by the models, and they are never exposed to steps beyond that.

\vspace*{-2mm}

 \begin{figure}[h]
    \centering
    \includegraphics[width=1\columnwidth ]{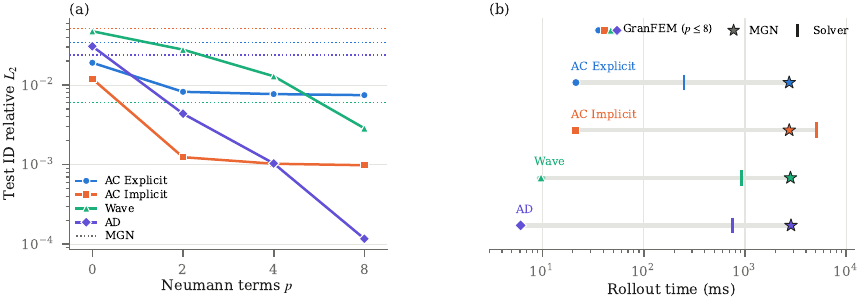}
    \vspace*{-6mm}
   \caption{\modelname with the Neumann-series base: (a) test ID error vs.\ $p$ (dotted: MGN); (b) rollout time per trajectory.}

    \label{fig:Neumman_Gem_vs_p}
   
\end{figure}

\textbf{Results.} The results for the various benchmarks and datasets are summarized in Table~\ref{table:operator_results}, reporting the mean and standard deviation for ID and OOD trajectory-level relative $L_2$ error, as per Appendix~\ref{sec:Metrics}. Snapshots for the emulated solutions for the various models can be seen in Appendix~\ref{sec:Appendix_Operator_Learning_Additional_Results}. Initially, we consider  \modelname without  the  Neumann expansion of Section~\ref{sec:neumann}, \ie $p =0$. The learned correction reduces the ID error of the lumped approximation by $1.5$--$4$ times, and \modelname attains the lowest errors for AC and the lowest OOD error for AD. Specifically, for AC, it attains a relative $L_2$ error equal to   $0.019/0.026$  and  $0.012/0.018$ (ID/OOD) on the \emph{Explicit} and \emph{Implicit} dataset, respectively, compared to  $0.031/0.047$ and $0.051/0.063$ for the best-performing baselines (TensorPILS/FEN and  MGN/FEN), \ie approximately 1.7 and 4 times lower. Notably, although the \emph{Implicit} scheme does not admit the explicit form of (\ref{eq:general_polynomial_update}) (see Appendix~\ref{sec:Appendix_Operator_Learning_Setup}), \modelname\ absorbs the discrepancy between the explicit baseline and the implicit reference dynamics, attaining a similar accuracy. For AD, it yields an ID error close to that of MGN, whilst in the OOD regime its error is about half. MGN scores a substantially smaller error for the wave equation; however, as shown in Figure~\ref{fig:Neumman_Gem_vs_p}, the Neumann expansion of Section~\ref{sec:neumann} improves \modelname's fidelity,  achieving a relative error below $1 \%$ for all PDEs, and for $p=8$ \modelname outperforms MGN in wave as well.

\begin{table*}[t]
\centering
\caption{Zero-shot resolution and domain extrapolation relative $L_2$ and computational scaling.}\vspace{-2mm}
\label{table:Extrapolation_dof_and_domain}

\begin{subtable}[t]{0.36\textwidth}
\centering
\caption{DoF extrapolation.}\vspace{-2mm}
\label{table:Extrapolation_dof}
\scriptsize
\setlength{\tabcolsep}{2.5pt}
\renewcommand{\arraystretch}{0.75}
\begin{tabular}{c c ccc}
\toprule
\multirow{2}{*}{DoF} & \multirow{2}{*}{$h$} & \multirow{2}{*}{TensorPILS} & \multirow{2}{*}{\modelname} & \multirow{2}{*}{MGN} \\
& & & & \\
\midrule
250 & 0.132 & 0.59/0.68 & \textbf{0.28/0.36} & 5.04/5.73 \\
500 & 0.090 & 0.23/0.30 & \textbf{0.07/0.10} & 0.46/0.81 \\
633$^\dagger$ & 0.08$^\dagger$ & 0.08/0.10 & 0.05/0.05 & \textbf{0.01/0.01} \\
1000 & 0.063 & 0.31/0.49 & \textbf{0.06/0.08} & 0.28/0.87 \\
2500 & 0.039 & 0.62/0.77 & \textbf{0.07/0.08} & 3.67/5.45 \\
5000 & 0.027 & 0.73/0.92 & \textbf{0.06/0.07} & 2.02/2.25 \\
10000 & 0.019 & 0.86/1.05 & \textbf{0.06/0.06} & 1.54/1.19 \\
25000 & 0.012 & 1.06/1.23 & \textbf{0.05/0.05} & 1.62/1.08 \\
50000 & 0.0086 & 1.28/1.47 & \textbf{0.05/0.04} & 1.71/1.13 \\
100000 & 0.0061 & 1.47/1.75 & \textbf{0.05/0.04} & 1.82/1.30 \\
\bottomrule
\end{tabular}
\end{subtable}%
\hfill
\begin{subtable}[t]{0.60\textwidth}
\centering
\caption{Domain change extrapolation.}\vspace{-2mm}
\label{table:Extrapolation_domain}
\scriptsize
\setlength{\tabcolsep}{2pt}
\renewcommand{\arraystretch}{1.05}
\resizebox{\linewidth}{!}{%
\begin{tabular}{c ccc ccc ccc}
\toprule
& \multicolumn{3}{c}{Rescale} & \multicolumn{3}{c}{Shift} & \multicolumn{3}{c}{Rescale+Shift} \\
\cmidrule(lr){2-4}\cmidrule(lr){5-7}\cmidrule(lr){8-10}
$h$ & TensorPILS & \modelname & MGN & TensorPILS & \modelname & MGN & TensorPILS & \modelname & MGN \\
\midrule
0.16
& 0.56/0.33 & \textbf{0.45/0.26} & 3.53/2.55
& 0.56/0.40 & \textbf{0.40/0.31} & 3.01/2.03
& 0.64/0.98 & \textbf{0.53/1.71} & 4.12/4.99 \\

0.12
& 0.48/0.36 & \textbf{0.27/0.27} & 4.16/3.56
& 0.46/0.35 & \textbf{0.21/0.22} & 3.77/3.00
& 0.46/0.72 & \textbf{0.28/0.82} & 3.54/5.00 \\

0.10
& 0.32/0.32 & \textbf{0.17/0.19} & 3.54/3.66
& 0.30/0.29 & \textbf{0.13/0.14} & 4.00/3.72
& 0.28/0.49 & \textbf{0.21/0.44} & 0.81/1.31 \\

0.08
& \textbf{0.07/0.07} & 0.10/0.11 & 0.75/0.87
& 0.09/0.09 & \textbf{0.06/0.07} & 0.36/0.52
& \textbf{0.15/0.33} & \textbf{0.15/0.32} & 0.32/0.97 \\

0.04
& 0.56/0.56 & \textbf{0.07/0.08} & 3.39/4.87
& 0.55/0.49 & \textbf{0.05/0.05} & 4.01/5.21
& 0.68/1.61 & \textbf{0.11/0.19} & 4.90/11.75 \\

0.02
& 0.73/0.76 & \textbf{0.06/0.06} & 1.84/1.72
& 0.70/0.61 & \textbf{0.05/0.04} & 1.86/1.41
& 0.93/2.16 & \textbf{0.09/0.13} & 1.62/1.91 \\

0.01
& 1.04/0.96 & \textbf{0.05/0.04} & 1.53/1.13
& 1.01/0.82 & \textbf{0.04/0.03} & 1.38/0.94
& 1.40/2.66 & \textbf{0.08/0.10} & 1.65/1.71 \\

0.008
& 1.21/1.12 & \textbf{0.05/0.04} & 1.52/1.08
& 1.10/0.92 & \textbf{0.04/0.03} & 1.36/0.93
& 1.52/2.89 & \textbf{0.08/0.10} & 1.73/1.79 \\
\bottomrule
\end{tabular}%
}
\end{subtable}

\par\vspace{1mm}
\begin{subtable}[t]{1\textwidth}
\centering
\caption{Runtime per trajectory (ms) and speedup.}\vspace{-2mm}
\label{table:extrapolation_runtime_scaling}
\scriptsize
\setlength{\tabcolsep}{3pt}
\renewcommand{\arraystretch}{0.85}
\begin{tabular*}{\textwidth}{@{\extracolsep{\fill}}lrrrrrrrrrrr}
\toprule
DoF & 100 & 250 & 500 & 633 & 1000 & 2500 & 5000 & 10000 & 25000 & 50000 & 100000 \\
\midrule
\modelname\ (ms) & 8.9 & 9.0 & 9.1 & 9.0 & 9.0 & 8.9 & 9.1 & 9.1 & 9.0 & 11.8 & 16.5 \\
Solver (ms) & 870 & 917 & 913 & 910 & 914 & 917 & 925 & 934 & 938 & 946 & 1182 \\
Speedup & 97$\times$ & 102$\times$ & 101$\times$ & 101$\times$ & 101$\times$ & 103$\times$ & 101$\times$ & 103$\times$ & 104$\times$ & 80$\times$ & 72$\times$ \\
\bottomrule
\end{tabular*}
\end{subtable}
\vspace{-3mm}
\end{table*}

At the same time, \modelname exhibits superior computational efficiency; it employs a simple three hidden-layer MLP, 
applying corrections once before the rollout, and subsequently the PDE trajectory is emulated via sparse message passing within 22, 9, and 6 ms for the AC, wave, and AD equations, respectively. DeepONet is the only faster baseline, but demonstrates substantially worse accuracy. In contrast, MGN, which is the main competitor in terms of accuracy, is characterized by prolonged rollout times, requiring more than  2000 ms per trajectory. 
Importantly, with the \emph{precomputed} operator on this resolution, the rollout time of \modelname\ is practically independent of $p$ ($21.7$--$22.0$, $9.5$, and $6.2$\,ms for AC, wave, and AD, respectively, for $p = 0$ to $8$). Further, considering that simulating a single AC,  wave, and AD trajectory on a GPU takes $\approx 250/5060$ (\emph{Explicit/Implicit}), 916, and 756 ms, respectively, \modelname substantially reduces the simulation time by one to two orders of magnitude while retaining the fidelity of the emulated PDE solution. Importantly, as also remarked in \citep{mcgreivy2024weak_baselines_time},  it can be observed that neural solvers can be slower than well-optimized GPU-accelerated conventional PDE solvers. Indeed, on these meshes, single-trajectory GPU inference is dominated by per-operation latency, and large deep networks, such as MGN with 15 message-passing layers, execute hundreds of GPU kernels per step and exceed the reference solver, whereas \modelname  executes one or two sparse products per step.   Overall, the results corroborate that the learned operator $S_\theta$ can effectively compensate for the approximate nature of (\ref{eq:general_polynomial_update}), accurately capture the interaction between nodes, and emulate the dynamics of different PDEs, yielding the best accuracy--efficiency trade-off.

\vspace{-2mm}

\subsection{DoF and Domain Extrapolation}   \label{sec:Results_Extrapolation}
\vspace{-2mm}
We next probe the potential of directly applying the learned operators to different mesh resolutions and domains. This is a zero-shot evaluation: a model trained once, on samples of a specific domain and mesh granularity, is applied to meshes with substantially more DoFs and to different domains, with no retraining or fine-tuning. For the wave equation, we first explore resolution extrapolation and then study the impact of a domain change. To this end, we gradually reduce the element size, $h$, increasing the number of DoFs up to $10^5$, and we generate additional trajectories by rescaling and shifting the original disk (radius $1$, centered at the origin) to radius $2$ (rescale), center $(1,1)$ (shift), and radius $0.5$ with center $(0.5,0.5)$ (rescale and shift). Further details on these datasets are provided in Appendix~\ref{sec:Appendix_Domain_Extr}. We consider \modelname ($p=0$; results for other values of $p$ are given in Appendix~\ref{sec:Appendix_Add_Res_Neuman}) and the two strongest baselines on the wave equation, \ie TensorPILS and MGN, using the trained models of Section~\ref{sec:Results_Op_learning} ($h{=}0.08$, 633-node mesh). In addition to the resolution and domain, the test trajectories themselves have not been seen by any model during training.
 
Tables~\ref{table:Extrapolation_dof} and~\ref{table:Extrapolation_domain} report the ID/OOD relative $L_2$ error for the DoF and domain extrapolation, respectively; the training resolution is marked with a dagger. All models degrade when the target mesh is very coarse relative to the underlying dynamics, both for DoF and domain extrapolation. However, on finer meshes ($h<0.08$), TensorPILS and MGN break down, whereas the error of \modelname remains consistent and below $10\%$ for all DoFs. Similarly,  for the domain extrapolation,  the error of \modelname\ decreases to $0.03$--$0.05$ on the finest rescaled and shifted meshes, and to $0.08$--$0.10$ when both transformations are combined. Additionally,  as shown in Appendix~\ref{sec:Appendix_Add_Res_Neuman}, the Neumann expansion further reduces the errors, at a modest additional cost. At the training resolution, TensorPILS outperforms \modelname\ on the rescaled domains, and on the coarsest rescaled and shifted mesh, \modelname's OOD error grows. However, once the mesh is refined  \modelname\ yields consistently the highest accuracy. At the same time, the per-trajectory rollout time of \modelname\ is about two orders of magnitude lower than that of the conventional solver (Table~\ref{table:extrapolation_runtime_scaling}).  Note that the speedups in Table~\ref{table:extrapolation_runtime_scaling} refer to $p=0$. With the Neumann expansion, which reduces the error to $0.3$--$0.7\%$ for $p=8$ on meshes from $10^3$ to $5\times10^4$ nodes (Appendix~\ref{sec:Appendix_Add_Res_Neuman}), the precomputed operator remains $14$--$100\times$ faster than the reference solver up to $5\times10^4$ nodes.

 \vspace{-2mm}

  \vspace{-2 mm}

\section{Discussion and Conclusion}
\vspace{-2 mm}

 We introduced \modelname, a hybrid graph-native framework in which the graph is not merely induced by the FEM mesh or supplied to a downstream neural model, but serves as the backbone of computational representation itself. Beyond recovering conventional FEM solutions, this common representation provides a direct interface to operator learning: the FEM discretization determines the graph connectivity and baseline interactions, while learning is restricted to local corrections around these quantities. A truncated Neumann-series approximation further refines the inverse-mass baseline beyond mass lumping, capturing longer-range interactions between graph nodes and improving rollout fidelity. Our results demonstrate accurate long-horizon predictions and, notably, as shown for the wave equation, zero-shot transfer across unseen mesh resolutions and affine  domain transformations, yielding substantial computational speedups at larger scales.    Several directions remain open. The present learning experiments consider individual PDE systems rather than coupled multiphysics problems, and the efficient local numerical kernels are currently PDE-specific. Extending \modelname to heterogeneous coupled fields and automatically deriving graph kernels from general weak forms would broaden its applicability.   More broadly, the common operator representation opens the possibility of training a single model across multiple PDEs, physical parameters, discretizations, and domains, providing a potential path toward finite element-conditioned foundation models for scientific computing.




\vspace{-3mm}
\section*{Acknowledgements}
\vspace{-2mm}
This work was supported by the European Commission through the  Marie Skłodowska-Curie Actions Postdoctoral Fellowship Program, Grant No. 101279286, PANDA. The author would like to thank Siddhartha Mishra for insightful scientific discussions, valuable feedback on the manuscript, and his support throughout this work.

\clearpage

\bibliography{example_paper}
\bibliographystyle{icml2026}

\newpage
\appendix
\onecolumn

\renewcommand{\thefigure}{A\arabic{figure}}
\setcounter{figure}{0}

\renewcommand{\thetable}{A\arabic{table}}
\setcounter{table}{0}

\section{Numerical PDE Solver} \label{sec:Appendix_Num_PDE_Solver}

\subsection{Problem Setup} 
\label{sec:Appendix_Num_PDE_Prob_Setupt}

\subsubsection{3D  problems}
Here, we demonstrate the potential of \modelname to serve as a classic numerical PDE solver, setting the foundations for the neural experiments presented earlier in this work. We study the following two PDE benchmarks. 

\paragraph{Benchmark I: 3D Poisson Equation.}

On the unit cube $\Omega = [0,1]^3$, we solve

\begin{equation}
\begin{cases}
-\Delta u(\mathbf{x}) = f(\mathbf{x}) & \text{in } \Omega, \\
u(\mathbf{x}) = 0 & \text{on } \partial\Omega,
\end{cases}
\label{eq:poisson}
\end{equation}

with source $f(\mathbf{x}) = \sin(\pi x_1)\sin(\pi x_2)\sin(\pi x_3)$ and homogeneous Dirichlet boundary conditions on all faces.

\paragraph{Benchmark II: 3D Linear Elasticity.}

We consider the static equilibrium of an isotropic linear elastic body on the unit cube $\Omega = [0,1]^3$:

\begin{equation}
\begin{cases}
-\nabla \cdot \boldsymbol{\sigma}(\mathbf{x}) = \mathbf{f}(\mathbf{x})
    & \text{in } \Omega, \\
\mathbf{u}(\mathbf{x}) = \mathbf{0}
    & \text{on } \Gamma_0, \\
\boldsymbol{\sigma}(\mathbf{x})\,\mathbf{n} = \mathbf{0}
    & \text{on } \partial\Omega \setminus \Gamma_0,
\end{cases}
\label{eq:elasticity}
\end{equation}

where $\mathbf{u} : \Omega \rightarrow \mathbb{R}^3$ is the displacement, $\Gamma_0 = \{\mathbf{x} \in \partial\Omega : x_3 = 0\}$ is the base face, and the Cauchy stress satisfies Hooke's law:

\begin{equation}
\boldsymbol{\sigma}
=
\lambda\,\operatorname{tr}(\boldsymbol{\varepsilon})\mathbf{I}
+ 2\mu\boldsymbol{\varepsilon}  =\mathbb{C}:\boldsymbol{\varepsilon},
\qquad
\boldsymbol{\varepsilon}
=
\frac{1}{2}
\left(
\nabla\mathbf{u} + (\nabla\mathbf{u})^\top
\right), 
\label{eq:hooke}
\end{equation}

where $\mathbb{C}$ is the fourth-order elasticity tensor of Hooke's law, and we set Young's modulus $E = 1$ and Poisson's ratio $\nu = 0.3$, evaluating the following $\lambda$ and $\mu$ values:

\begin{equation}
\lambda
=
\frac{E\nu}{(1+\nu)(1-2\nu)},
\qquad
\mu
=
\frac{E}{2(1+\nu)}.
\label{eq:lame}
\end{equation}

 A lateral body force $\mathbf{f}(\mathbf{x}) = \big(\sin(\pi x_1)\sin(\pi x_2)\sin(\pi x_3),\, 0,\, 0\big)^\top$ is applied, inducing bending of the cantilevered block. This is a base-clamped cantilever configuration, where only the face $\Gamma_0$ is fixed, while the remaining five faces are traction-free.

\subsubsection{Equivalent 2D  problems} To complement these three-dimensional benchmarks, we also evaluate both methods on their two-dimensional counterparts. For Poisson, we solve the same boundary-value problem as (\ref{eq:poisson}), restricted to the unit square $\Omega = [0,1]^2$, with source $f(\mathbf{x}) = \sin(\pi x_1)\sin(\pi x_2)$ and homogeneous Dirichlet boundary conditions on all four sides. The number of DoF is varied   by gradually increasing the mesh resolution  from 16 to 3560 and 8 to 118 subdivisions per side for the 2D and 3D case, respectively.  

For elasticity, we evaluate the same base-clamped cantilever configuration as in three dimensions, restricted to the unit square $\Omega=[0,1]^2$ under plane-strain conditions: only the base edge $x_2=0$ is fixed ($\mathbf{u}=\mathbf{0}$), the remaining three edges are traction-free, and a lateral body force $\mathbf{f}(\mathbf{x}) = \big(\sin(\pi x_1)\sin(\pi x_2),\, 0\big)^{\top}$ is applied, inducing bending of the cantilevered block. Material properties ($E=1$, $\nu=0.3$) and Hooke's law, ~(\ref{eq:hooke}), are unchanged from the three-dimensional case. As with Poisson, the number of DoFs is varied by gradually increasing the mesh resolution from 16 to 1536 and 8 to 52 subdivisions per side for the 2D and 3D case, respectively (note that different upper resolution limits apply to the Poisson and elasticity, due to the higher GPU memory consumption of the latter).

\subsection{Instantiating the Graph-Driven Framework}
\label{sec:Appendix_Example_Graph_Driven_Framework}

To tangibly exemplify the general framework of Section~\ref{sec:graph_assembly}, we derive $\mathbf{x}_a$, $\mathbf{g}_{loc}$, and $f_g$ explicitly for both PDEs, starting from the general quadrature formula in (\ref{eq:local_stiffness_matrix}). The key observation underpinning the \textit{graph-driven} approach is that, for any bilinear form obtained via the standard isoparametric mapping from a reference element
$\hat{\Omega}$ to the physical element $\Omega_e$, the quadrature integrand always factors as:

\begin{equation}
I_{ab}(\hat{\boldsymbol{\xi}}_q) = \underbrace{|\!\det \mathbf{J}_e(\hat{\boldsymbol{\xi}}_q)|}_{\text{depends only on the element, not on } a, b}
\;\times\;
\underbrace{\tilde{I}_{ab}(\hat{\boldsymbol{\xi}}_q)}_{\text{depends on the pair of basis functions } (a,b)},
\label{eq:factorization}
\end{equation}

 where the Jacobian determinant, $\mathrm{d}\Omega_e = |\!\det \mathbf{J}_e(\hat{\boldsymbol{\xi}})| \, \mathrm{d}\hat{\boldsymbol{\xi}}$, arises purely from the reference-to-physical change of coordinates and is common for every element local matrix, regardless of which basis function pair is being integrated.  This allows distinguishing a per-element geometric quantity, $\mathbf{g}_{loc}$, which is computed once and shared by every pair of element nodes, and a per-node quantity, $\mathbf{x}_a$, computed once per basis function and reused across all of its messages. The message function $f_g$ then entwines these two to compute $(K_e)_{ab}$.

\paragraph{Poisson equation.} The bilinear form is $a(u,v) = \int_\Omega
\nabla_{\mathbf{x}} u \cdot \nabla_{\mathbf{x}} v \, \mathrm{d}\Omega$, so on
element $e$,
\begin{equation}
a_e(\phi_b, \phi_a) = \int_{\Omega_e} \nabla_{\mathbf{x}}\phi_b \cdot
\nabla_{\mathbf{x}}\phi_a \, \mathrm{d}\Omega_e
\;\approx\;
\sum_{q=1}^{N_Q} \hat{w}_q \, |\!\det \mathbf{J}_e(\hat{\boldsymbol{\xi}}_q)|
\;
\nabla_{\mathbf{x}}\phi_a(\hat{\boldsymbol{\xi}}_q) \cdot
\nabla_{\mathbf{x}}\phi_b(\hat{\boldsymbol{\xi}}_q).
\end{equation}

Consequently, the pair-independent factor is $|\det \mathbf{J}_e(\hat{\boldsymbol{\xi}}_q)|$, so we set $\mathbf{g}_{\mathrm{loc}} = \mathopen{}\big[|\det \mathbf{J}_e(\hat{\boldsymbol{\xi}}_1)|,\ldots, |\det \mathbf{J}_e(\hat{\boldsymbol{\xi}}_Q)|\big]$, whilst the pair-dependent factor $\tilde{I}_{ab}(\hat{\boldsymbol{\xi}}_q)
= \nabla_{\mathbf{x}}\phi_a(\hat{\boldsymbol{\xi}}_q) \cdot \nabla_{\mathbf{x}}\phi_b(\hat{\boldsymbol{\xi}}_q)$
entails the inner product of node basis function gradients, so each node is assigned with $\mathbf{x}_a = \big[\nabla_{\mathbf{x}}\phi_a(\hat{\boldsymbol{\xi}}_1),\ldots, \nabla_{\mathbf{x}}\phi_a(\hat{\boldsymbol{\xi}}_Q)\big]$, and the local deterministic graph operator  is:

\vspace{-6mm}

\begin{equation}
f_g(\mathbf{g}_{loc}, \mathbf{x}_a, \mathbf{x}_b)
=
\hat{\mathbf{w}}^\top
\Big(\mathbf{g}_{loc} \odot
\big[\mathbf{x}_a^{(q)\top}\mathbf{x}_b^{(q)}\big]_{q=1}^Q\Big)
=
\sum_{q=1}^{N_Q} \hat{w}_q \,
|\!\det \mathbf{J}_e(\hat{\boldsymbol{\xi}}_q)| \,
\mathbf{x}_a^{(q)\top}\mathbf{x}_b^{(q)}
= (K_e)_{ab}.
\end{equation}

\vspace{-4mm}

\paragraph{Linear elasticity.} The bilinear form is $a(\mathbf{u},\mathbf{v}) = \int_\Omega \boldsymbol{\varepsilon}(\mathbf{v}) : \mathbb{C} : \boldsymbol{\varepsilon}(\mathbf{u}) \, \mathrm{d}\Omega$, with $\mathbb{C}$ the fourth-order elasticity tensor of Hooke's law as per (\ref{eq:hooke}).  To carry out the same factorization, we first rewrite the strain-energy contraction in the bilinear form using the  Voigt (vector) notation:  $a(\mathbf{u},\mathbf{v}) = \int_\Omega \boldsymbol{\varepsilon}_V(\mathbf{v})^\top  \mathbf{C}_V \boldsymbol{\varepsilon}_V(\mathbf{u}) \, \mathrm{d}\Omega$, where $ \boldsymbol{\varepsilon}_V = [\varepsilon_{11}, \varepsilon_{22}, \gamma_{12}]$ in 2D or $ \boldsymbol{\varepsilon}_V = [\varepsilon_{11}, \varepsilon_{22}, \varepsilon_{33}, \gamma_{23}, \gamma_{31}, \gamma_{12}]$ in 3D, with $\varepsilon_{ij} = 0.5 \cdot \left( \partial u_i/ \partial x_j +  \partial u_j/ \partial x_i\right)$ and  $\gamma_{ij} = 2 \varepsilon_{ij}  $, and $\mathbf{C}_V$ is the isotropic elasticity matrix in the corresponding Voigt representation \citep{bower2010applied_voigt}. Consequently, for the finite element approximation (\ref{eq:fem_ansatz}),  the corresponding strain components are  $ \varepsilon_{ij}(\mathbf{u}_h)
= \frac{1}{2} \sum_{a=1}^{N_b} \left( U_{ai}\frac{\partial\varphi_a}{\partial x_j} + U_{aj}\frac{\partial\varphi_a}{\partial x_i} \right)
$. Plugging that into  $\boldsymbol{\varepsilon}_V $, the latter can be written compactly in  Voigt notation as $ \boldsymbol{\varepsilon}_{V}(\mathbf{u}_h) = \sum_{a=1}^{N_b}
\mathbf{B}_a\,\mathbf{U}_a = \mathbf{B}_V \mathbf{U} $,  where $\mathbf{B}_
a\in \mathbb{R}^{n_\varepsilon\times d}$ 
is the strain--displacement matrix associated with basis function $a$ \citep{bower2010applied_voigt},   $ n_\varepsilon$ is the number of strain components in the Voigt representation (\(3\) in 2D, \(6\) in 3D), and $\mathbf{B}_V = \left[\mathbf{B}_1, \mathbf{B}_2, ..., \mathbf{B}_{N_b} \right]$. Then, applying a similar treatment for the test function, the bilinear form reads:

\begin{equation}
a_e(\phi_b, \phi_a)
= \int_{\Omega_e} \big(\mathbf{B}_a \big)^\top \mathbf{C}_V \, \mathbf{B}_b
\, \mathrm{d}\Omega_e
\approx
\sum_{q=1}^{N_Q} \hat{w}_q \, |\!\det \mathbf{J}_e(\hat{\boldsymbol{\xi}}_q)|
\;
\mathbf{B}_a(\hat{\boldsymbol{\xi}}_q)^\top \, \mathbf{C}_V \,
\mathbf{B}_b(\hat{\boldsymbol{\xi}}_q),
\end{equation}

which assumes the same structure as (\ref{eq:factorization}). Hence, the pair-independent factor is again $|\det \mathbf{J}_e(\hat{\boldsymbol{\xi}}_q)|$, and  $\mathbf{g}_{loc}$ remains the same as in the Poisson case. The pair-dependent term now includes the matrix $\mathbf{B}_a(\hat{\boldsymbol{\xi}}_q) \in \mathbb{R}^{n_\varepsilon \times d}$, giving $\mathbf{x}_a = \big[\mathbf{B}_a(\hat{\boldsymbol{\xi}}_1), \ldots, \mathbf{B}_a(\hat{\boldsymbol{\xi}}_Q)\big]$, and the message function is defined as:

\begin{equation}
f_g(\mathbf{g}_{loc}, \mathbf{x}_a, \mathbf{x}_b)
=
\hat{\mathbf{w}}^\top
\Big(\mathbf{g}_{loc} \odot
\big[\mathbf{x}_a^{(q)\top}\mathbf{C}_V\,\mathbf{x}_b^{(q)}\big]_{q=1}^Q\Big)
\;\in\; \mathbb{R}^{d\times d}.
\label{eq:elasticity_message}
\end{equation}

Contrary to the Poisson case,  $\mathbf{x}_a^{(q)}$ is now a matrix rather than a vector, so the message $m_{b\to a} = f_g(\mathbf{g}_{loc},\mathbf{x}_a,\mathbf{x}_b) = (K_e)_{ab}\in \mathbb{R}^{d\times d}$ is a $d\times d$ block rather than a scalar. The assembled operator $\mathcal{G}_{\mathrm{FEM}}$ is therefore a block-weighted directed graph, with each edge $(i,j)$ carrying $\mathbf{K}_{ij} \in \mathbb{R}^{d\times d}$. Consequently, Section~\ref{sec:graph_system_solution}'s message-passing residual and matrix-vector product carry over unchanged once every node/edge quantity e is reinterpreted block-wise, \ie the block graph is expanded into an ordinary scalar graph over $d\cdot N$ degrees of freedom before the linear solve.

\subsubsection{Assembly Runtime Breakdown} \label{sec:Appendix_Num_PDE_Time_Breakdown}

\begin{figure*}[h]
    \centering

    \begin{subfigure}[b]{0.32\columnwidth}
        \centering
        \includegraphics[width=\columnwidth]{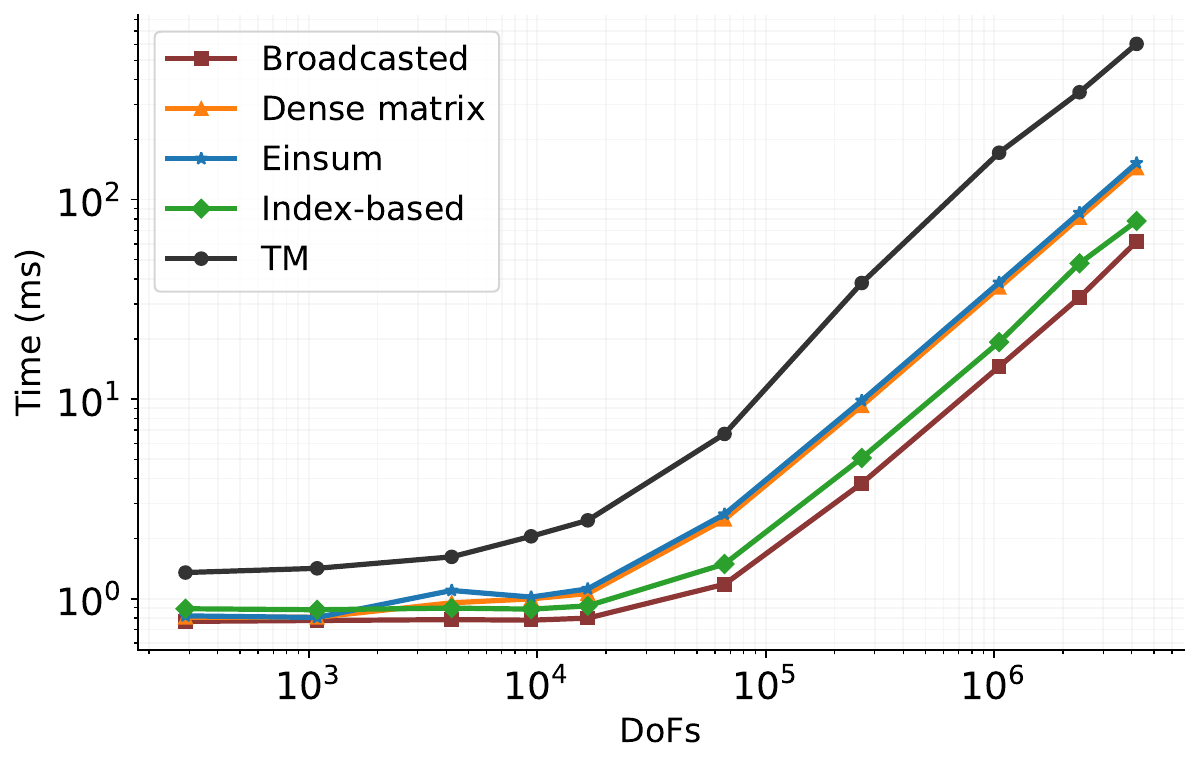}
        \caption{Poisson assembly 2D}
        \label{fig:Graph_Assembly_2D_Poisson_Time}
    \end{subfigure}
    \begin{subfigure}[b]{0.32\columnwidth}
        \centering
        \includegraphics[width=\columnwidth]{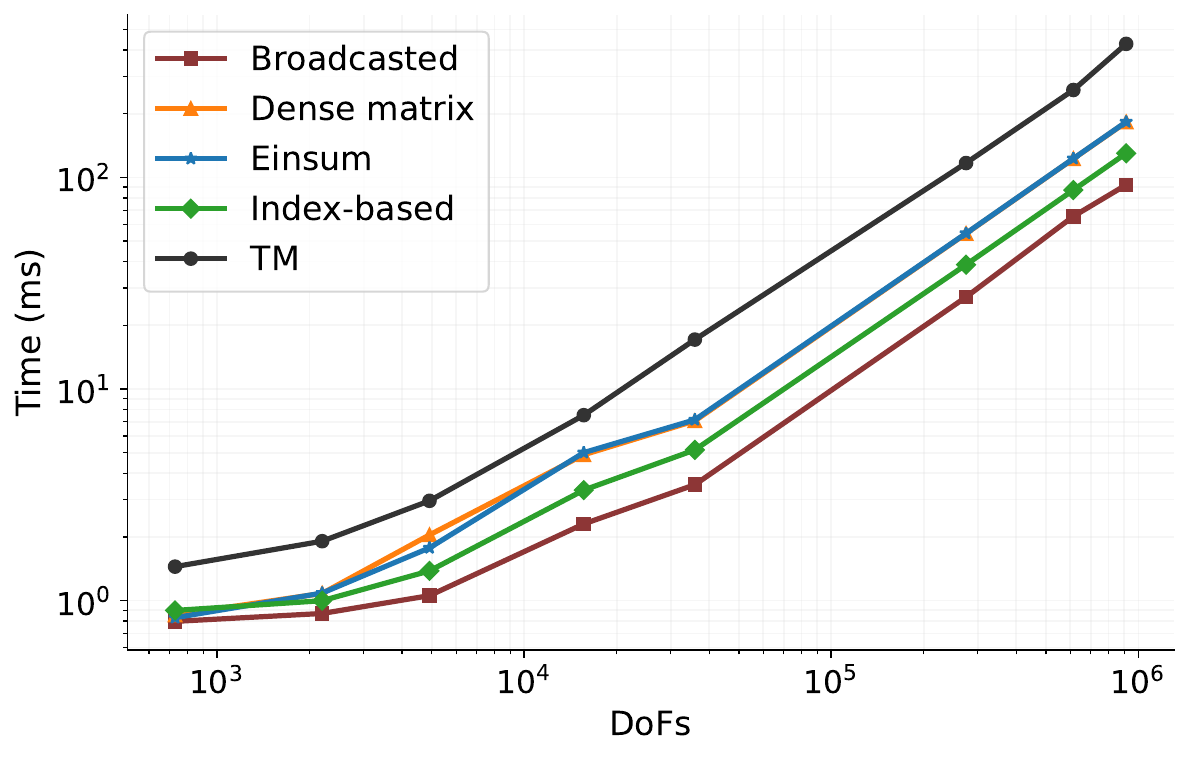}
        \caption{Poisson assembly 3D}
        \label{fig:Graph_Assembly_3D_Poisson_Time}
    \end{subfigure}
    \begin{subfigure}[b]{0.32\columnwidth}
        \centering
        \includegraphics[width=\columnwidth]{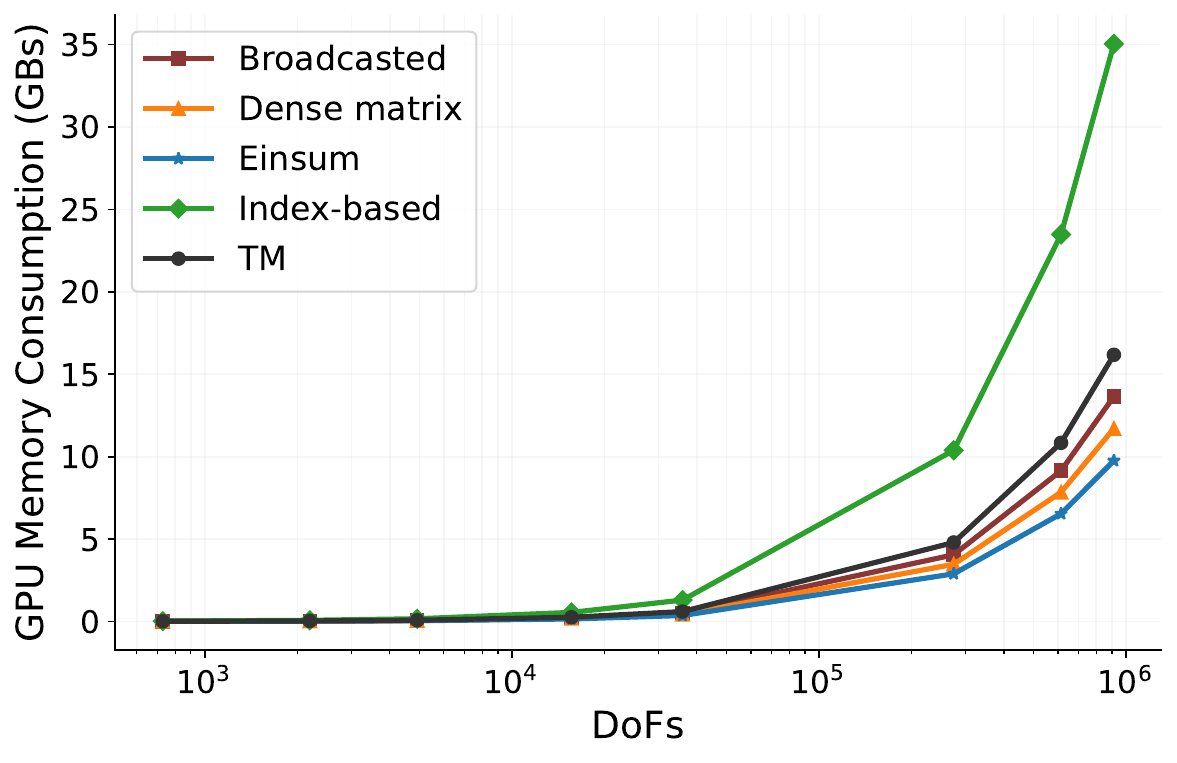}
        \caption{Memory Consumption  Poisson 3D}
        \label{fig:Graph_Assembly_Memory_Poisson}
    \end{subfigure}

    \begin{subfigure}[b]{0.32\columnwidth}
        \centering
        \includegraphics[width=\columnwidth]{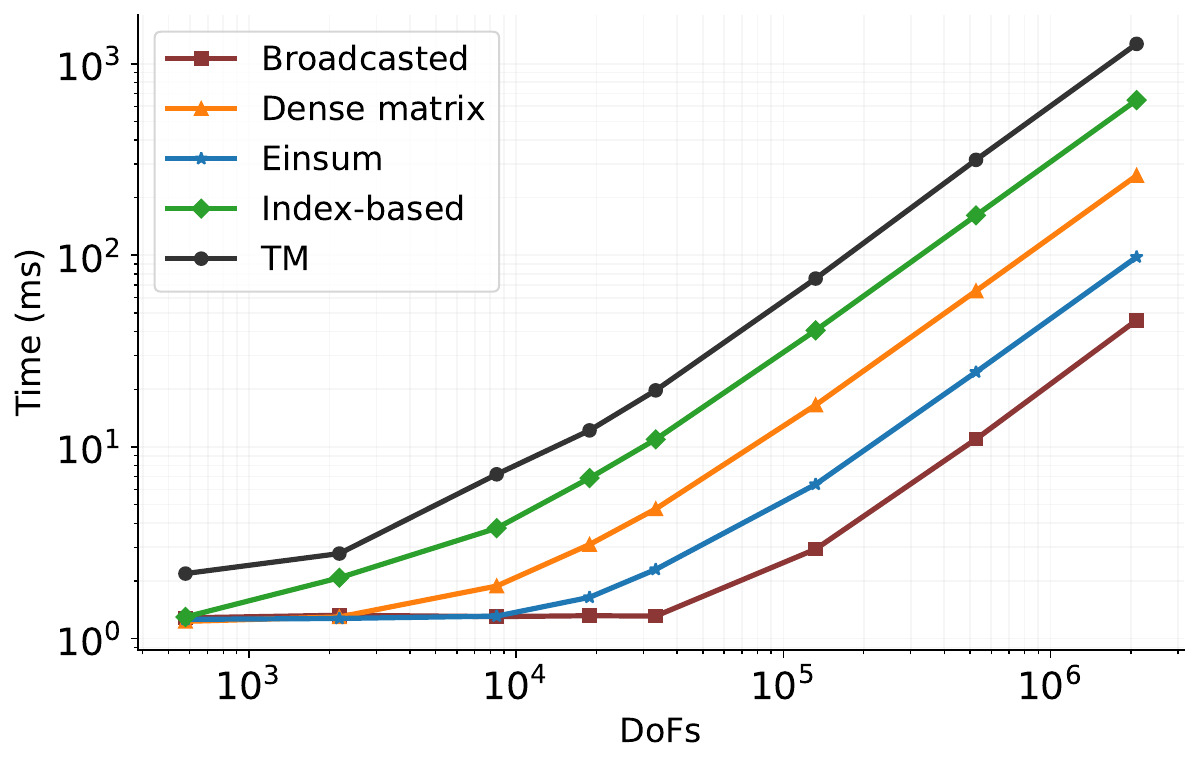}
        \caption{Elasticity 2D assembly}
        \label{fig:Elasticity_Graph_Assembly_2D_Time}
    \end{subfigure}
    \begin{subfigure}[b]{0.32\columnwidth}
        \centering
        \includegraphics[width=\columnwidth]{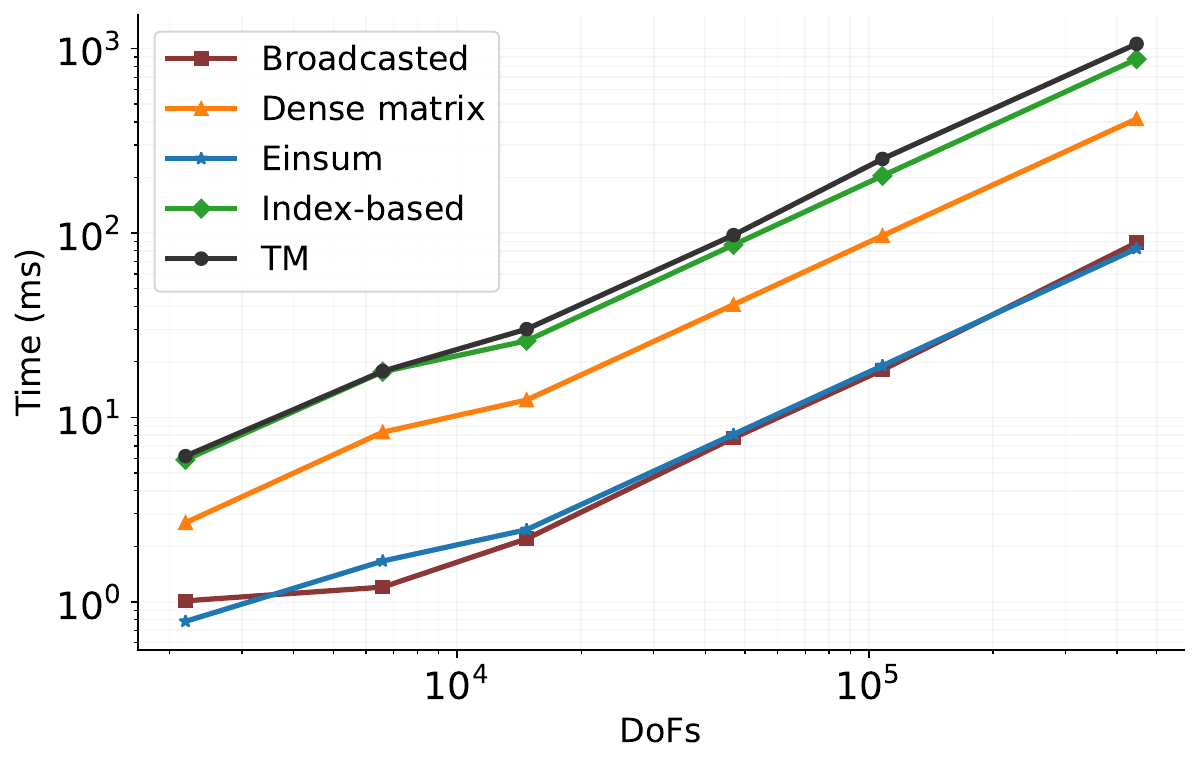}
        \caption{Elasticity 3D assembly}
        \label{fig:Elasticity_Graph_Assembly_3D_Time}
    \end{subfigure}
    \begin{subfigure}[b]{0.32\columnwidth}
        \centering
        \includegraphics[width=\columnwidth]{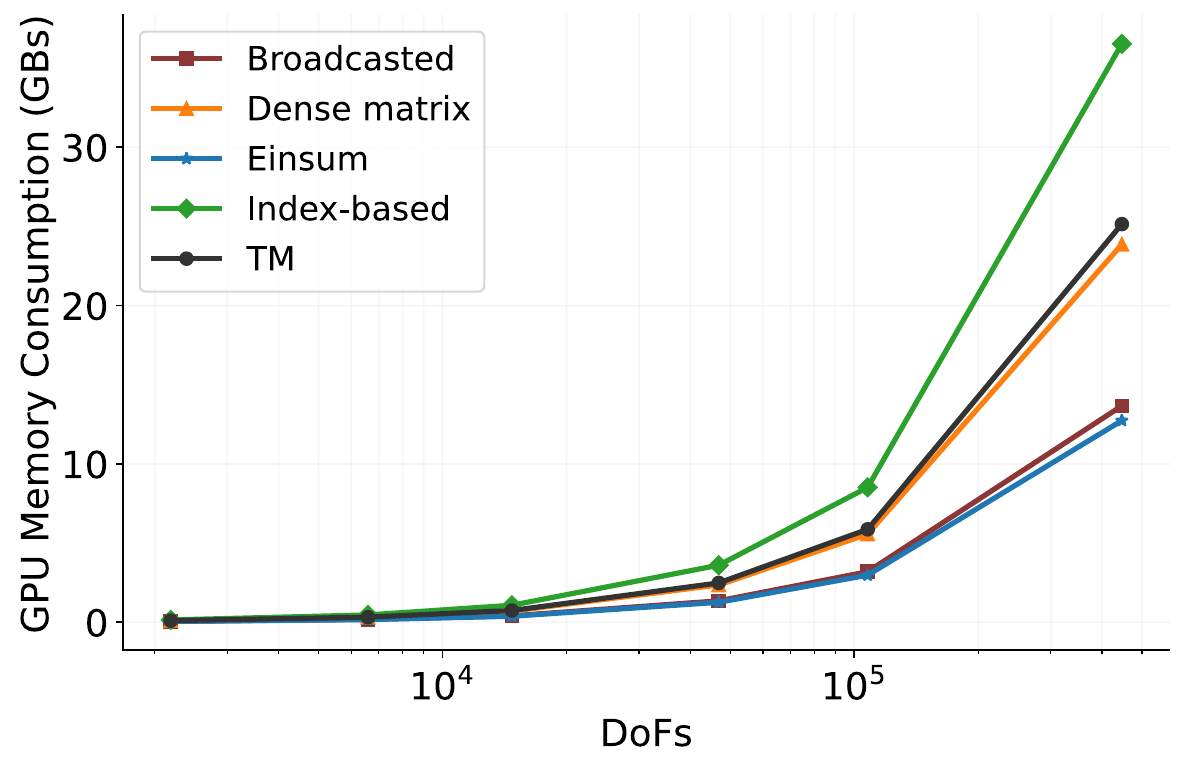}
        \caption{Memory Consumption Elasticity 3D}
        \label{fig:Elasticity_Graph_Assembly_Memory}
    \end{subfigure}
    \vspace{-0.75em}
    \caption{Graph-FEM assembly implementations. Runtime and GPU-memory scaling of alternative local-operator implementations for Poisson and linear elasticity.
    }
    \label{fig:Graph_Assembly_Implementations}
\end{figure*}

To construct the global stiffness matrix, it is necessary to calculate the local stiffness matrices for all finite elements and then fuse their contributions. For the \textit{graph-driven} computation of the local stiffness matrix, as described in Section~\ref{sec:graph_assembly}, we investigated different computational realizations of the same pairwise basis-function interactions. Specifically, assuming an [$E$, $Q$, $N_B$, $d$] tensor containing the feature vectors $\mathbf{x}_a$ (where $N_B$ is the number of local basis functions per element and $d$ is the spatial dimension), the local stiffness matrix can be computed using: (i) Einstein tensor contraction (\texttt{torch.einsum}), (ii) batched dense matrix multiplication, (iii) pairwise tensor broadcasting with multiply-reduction, and (iv) explicit edge-index-based pairwise multiplication. The first three exploit the fact that each element graph is small and fully connected and compute all edge quantities simultaneously using tensor algebra, whereas the last utilizes explicit indexes for each source--destination node pair before evaluating their pairwise interactions. Importantly, these alternatives differ only in the computational realization of the local interactions. In all cases, \modelname preserves its graph-native representation by bookkeeping the local-to-global node connectivity and eventually representing the assembled global stiffness operator as graph edge tuples $(i, j, K_{ij})$. The computation time along with the memory GPU consumption for the various graph-assembly implementations for the Poisson and linear elasticity PDE are depicted in  Figures~\ref{fig:Graph_Assembly_2D_Poisson_Time} to~\ref{fig:Graph_Assembly_Memory_Poisson}  and Figures~\ref{fig:Elasticity_Graph_Assembly_2D_Time} to~\ref{fig:Elasticity_Graph_Assembly_Memory}, respectively. From the four implementations considered, the pairwise tensor broadcasting yields the best performance trade-off in terms of computation time and memory consumption , \ie option (iii), and it is the one retained throughout this paper.

Figure~\ref{fig:Graph_Assembly_Implementations} also reveals a substantial difference in the time required by \modelname  and TM to compute the global stiffness matrix. To comprehend this difference, we delve into the submodules employed in the TM assembly pipeline. This entails performing calculations related to geometry/quadrature,  employing a vmap function in the computation of the local stiffness matrix, routing the local contributions to the global stiffness matrix, and ultimately building a sparse matrix. The last step is not required in \modelname , since it uses a native graph representation, whilst the routing process is identical. Similarly, the geometry/quadrature computation is also shared between TM and \modelname  outright — both call the same process, which evaluates the cell Jacobian and transforms shape-function gradients to physical coordinates via two batched tensor contractions and one batched matrix inversion applied across the whole mesh at once. Therefore, aside from the sparse matrix construction, the difference stems from the local matrix calculation. In particular, TM strategically uses a vmap function that maps a generic, PDE-agnostic leaf function, supplied by the user as the weak-form integrand   (\eg grad$u$ $\cdot$ grad$v$ for the Laplacian) over four nested axes: elements, quadrature points, and the two basis-function indices. Consequently, the same machinery can handle different PDEs, by only changing the leaf function, at the cost of per-call overhead relative to a form-specific fused kernel. On the other hand, \modelname  uses PDE-tailored kernels that are more computationally efficient at the cost of requiring distinct derivations per PDE.
 
\begin{figure*}[h!]
    \centering

    \begin{subfigure}[b]{0.45\textwidth}
        \centering
        \includegraphics[width=\textwidth]{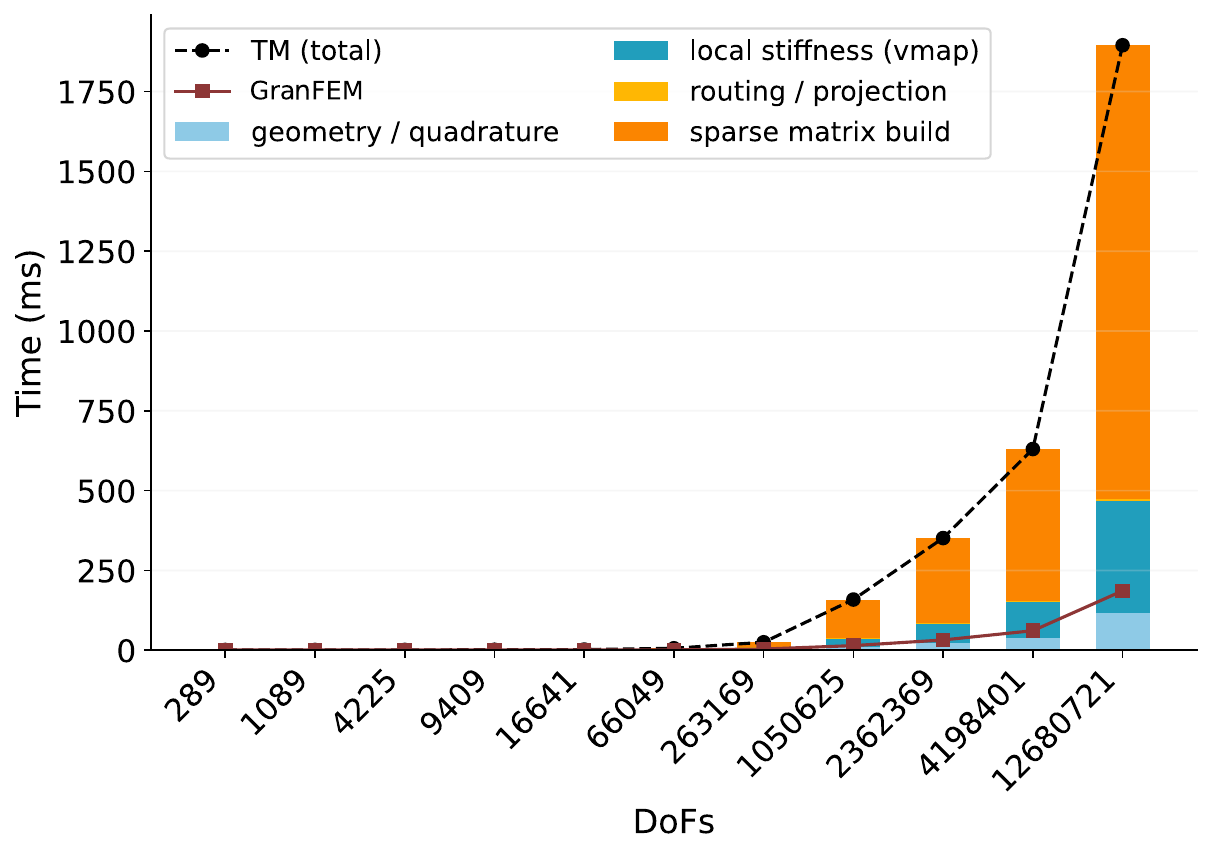}
        \caption{Poisson 2D runtime}
        \vspace{-2mm}
        \label{fig:Assembly_Breakdown_2D}
    \end{subfigure}
    \hfill
    \begin{subfigure}[b]{0.45\textwidth}
        \centering
        \includegraphics[width=\textwidth]{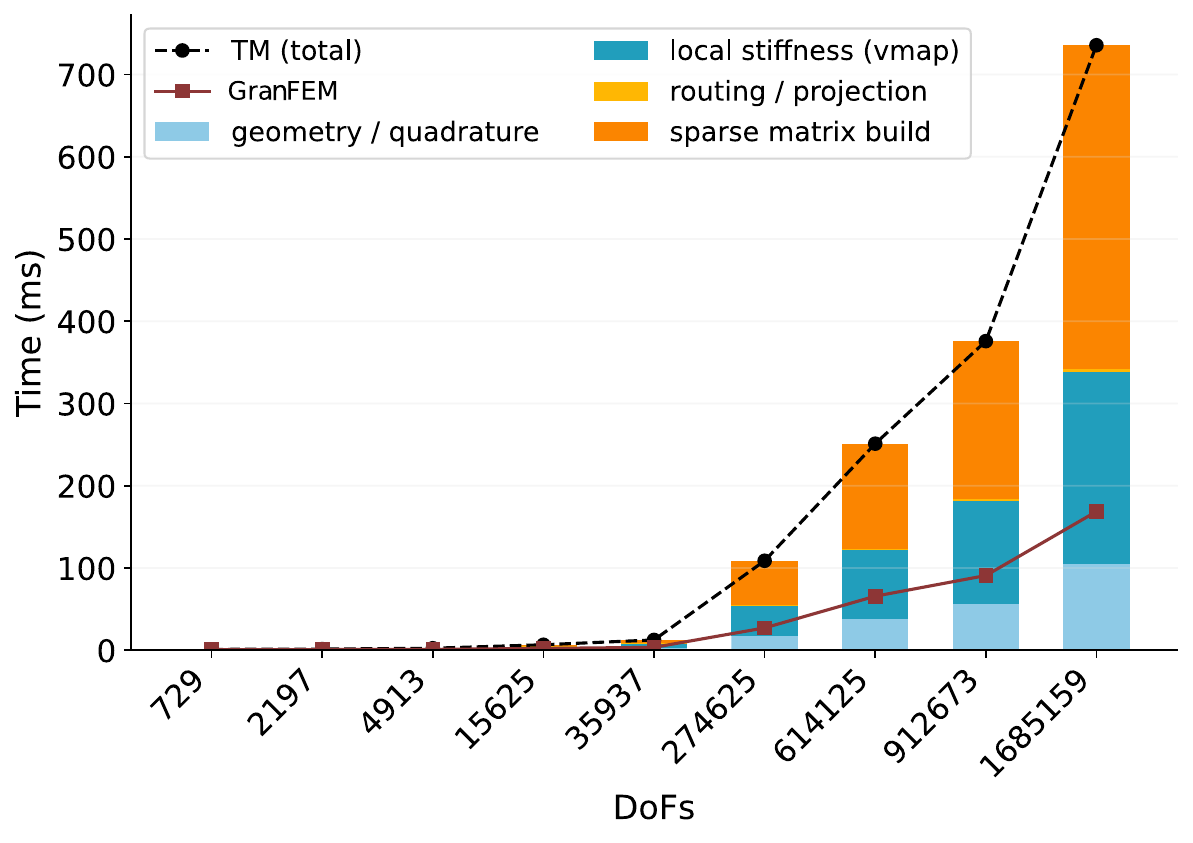}
        \vspace{-2mm}
        \caption{Poisson 3D runtime}
        \label{fig:Assembly_Breakdown_3D}
    \end{subfigure}
    \hfill 

    \caption{FEM assembly runtime breakdown. Runtime contributions of geometry/quadrature, local stiffness evaluation, routing/projection, and sparse-matrix construction compared with \modelname .}
    \label{fig:Assembly_Breakdowns}
\end{figure*}

 Figures~\ref{fig:Assembly_Breakdown_2D} and~\ref{fig:Assembly_Breakdown_3D} show the runtime of each computational TM submodule for the 2D and 3D Poisson assembly, respectively, and they also depict the \modelname  total assembly computational time  (using pairwise tensor broadcasting, \ie option (iii) ) as a distinct line. Given that geometry/quadrature-related calculations are the same for both \modelname  and TM, it can be seen that the local stiffness computation of \modelname  is only a small fraction of that of TM, being approximately 2-3 times faster, and that the TM assembly runtime is dominated by the sparse matrix construction. The sparse matrix construction time consists of building a coordinate (COO) matrix, $K$, for fast matrix multiplication, as well as a corresponding hash index array, which is used for a sparsity-pattern fingerprint and to confirm whether that structure has changed before applying boundary conditions. \modelname  avoids this step entirely: it does not require assembling the matrix into that structured form in the first place, and because it keeps working with the same underlying mesh connectivity throughout, one can straightforwardly probe changes without needing a hash array.

 To further probe the impact of the vmap function, we apply it separately to broadcast and einsum in the \modelname  pipeline. For Poisson, we find that there is no measurable impact, because the per quadrature point vmapped leaf function (a dot product of two gradient vectors) is too simple for the overhead associated with vmap to become noticeable. Conversely,  for elasticity, wrapping an einsum-based leaf in vmap incurs substantial overhead, increasing computational time by approximately 4 times compared to the vmap-free equivalent; however, wrapping the closed-form expression, posed purely as broadcasted elementwise operations, does not increase the runtime with respect to the vmap-free broadcast baseline.   This discrepancy stems from the operation-specific batching overhead of vmap. Broadcasted elementwise operations have simple batching rules, making vmap effectively cost-free, whereas the more general einsum requires additional batching transformations and dispatch. This overhead is amplified when vmap is nested across multiple axes, explaining the substantial slowdown of the vmapped einsum.


\subsection{Graph-driven Numerical PDE Solutions} \label{sec:Appendix_Num_PDE_Residuals}
 

 We then demonstrate the capacity of a fully graph-native framework to generate equivalent results to conventional FEM solvers and probe its computational efficiency both in two and three dimensions (2D/3D). With the stiffness matrix and the load vector assembled, the linear system (\ref{eq:fem_linear_system}) can be solved through classical linear algebra algorithms. Specifically, to solve the Poisson and the elastic equations, we use the CG and BiCGSTAB iterative algorithms, respectively. TM employs its native sparse linear equation solver library to run CG and BiCGSTAB \citep{chi2026torchsla}. For \modelname ,  we create an equivalent implementation,  mirroring the operation of   \citep{chi2026torchsla}, computing $\mathbf{K}\mathbf{U}$ on a graph representation as discussed in Section~\ref{sec:graph_system_solution}. We consider both a graph-native (scatter-based) and a CSR-based matvec, and use the latter for its speed on GPU. This is technically identical to the matvec used internally \citep{chi2026torchsla}  (both reduce to torch.mv on a cached CSR tensor), yet retains the graph representation, since the CSR structure is built directly from \modelname's own edges. Both CG and BiCGSTAB are run for up to a maximum of 10,000 iterations with Jacobi preconditioning, and they are terminated when the residual of the $k-$th iteration satisfies
$|r_k| < \max(\text{atol}, \ \text{rtol} \cdot |F|)$, where \text{atol} and \text{rtol} are the  absolute and relative tolerance set to $10^{-10}$, and $|F|$ is the Euclidean  norm of the load vector.

\begin{figure*}[t]
    \centering

    \begin{subfigure}[b]{0.32\columnwidth}
        \centering
        \includegraphics[width=\columnwidth]{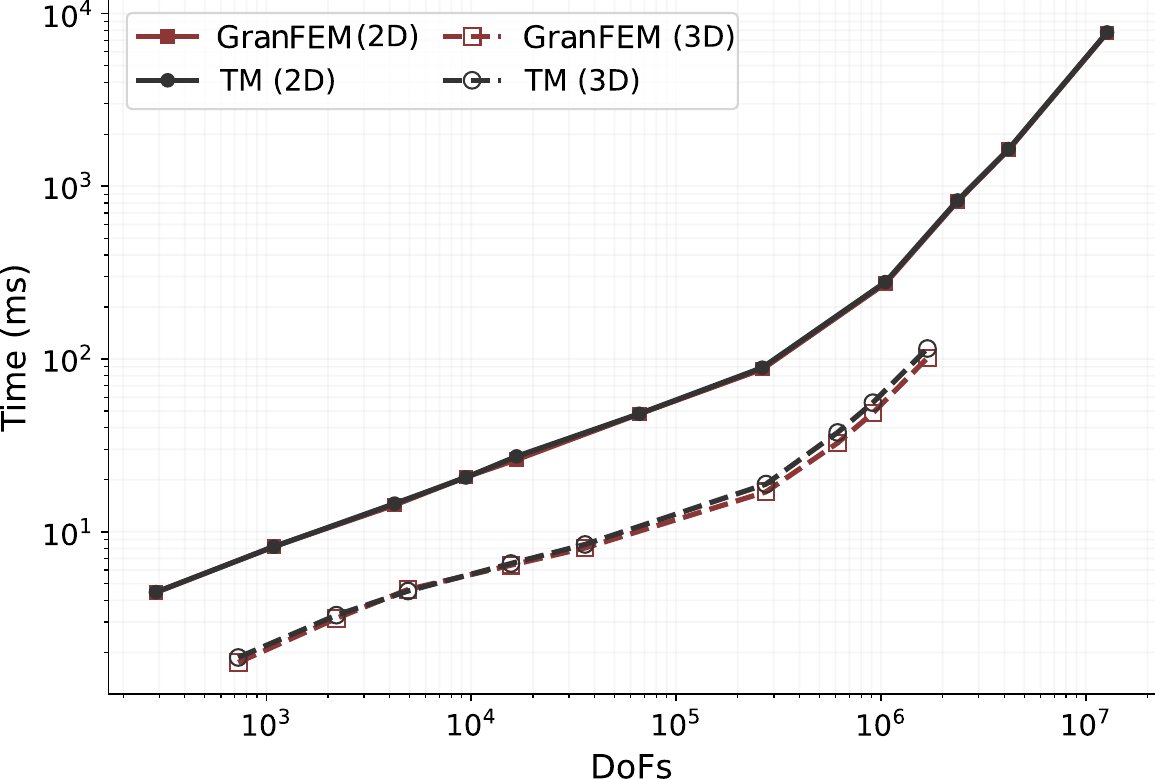}
        \caption{Poisson solution time}
        \label{fig:Linear_Solve_Time_Poisson}
    \end{subfigure}
    \begin{subfigure}[b]{0.32\columnwidth}
        \centering
        \includegraphics[width=\columnwidth]{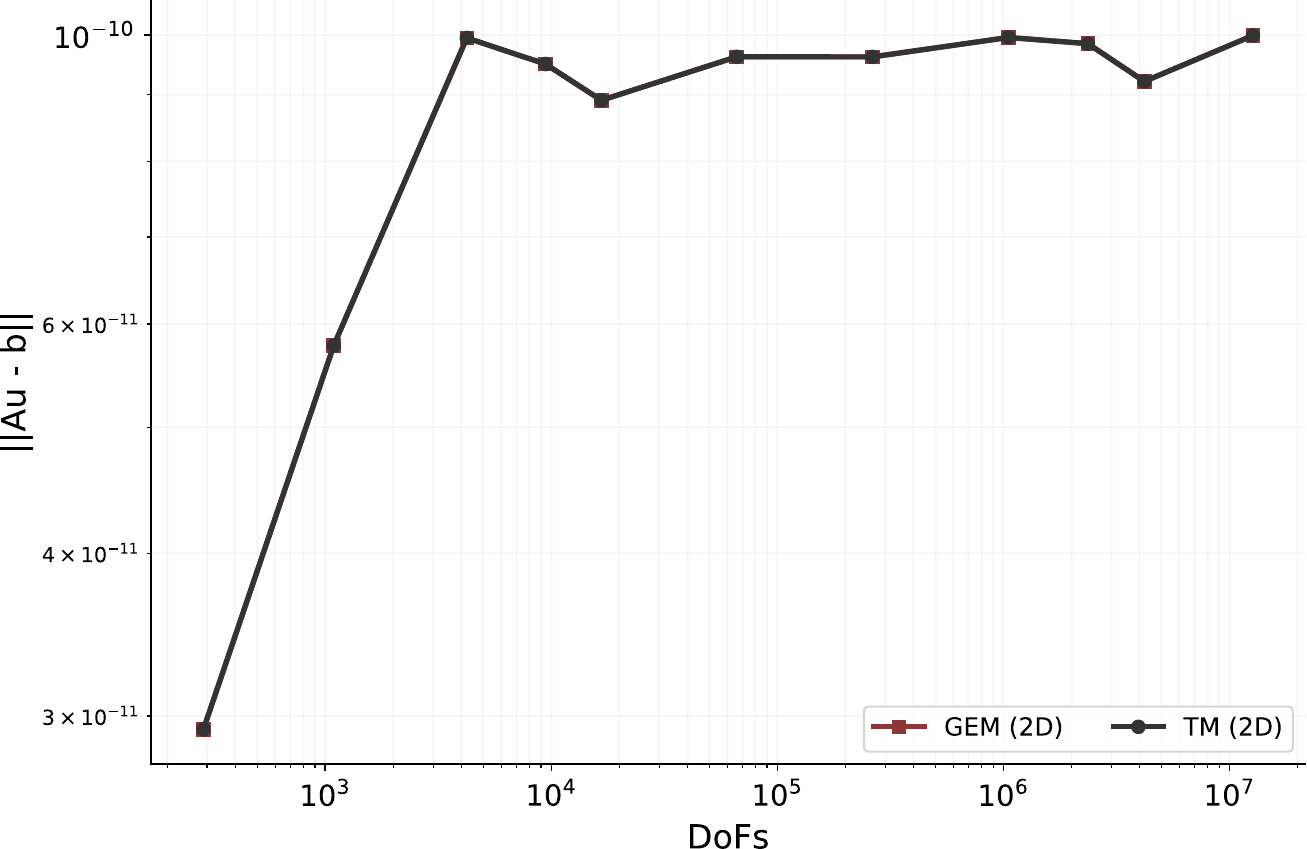}
        \caption{Poisson residual 2D}
        \label{fig:Residual_Poisson_2D}
    \end{subfigure}
    \begin{subfigure}[b]{0.32\columnwidth}
        \centering
        \includegraphics[width=\columnwidth]{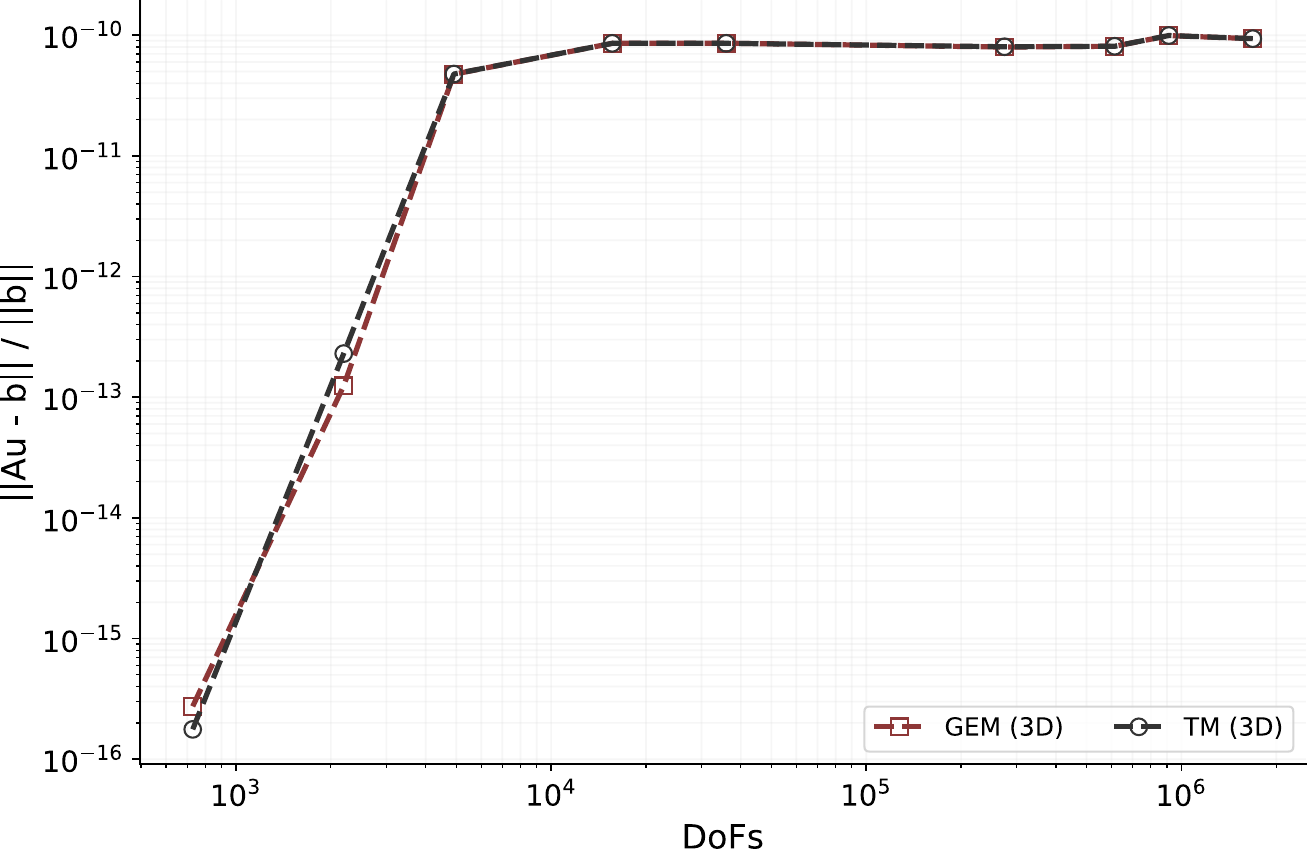}
        \caption{Poisson residual 3D}
        \label{fig:Residual_Poisson_3D}
    \end{subfigure}

    \begin{subfigure}[b]{0.32\columnwidth}
        \centering
        \includegraphics[width=\columnwidth]{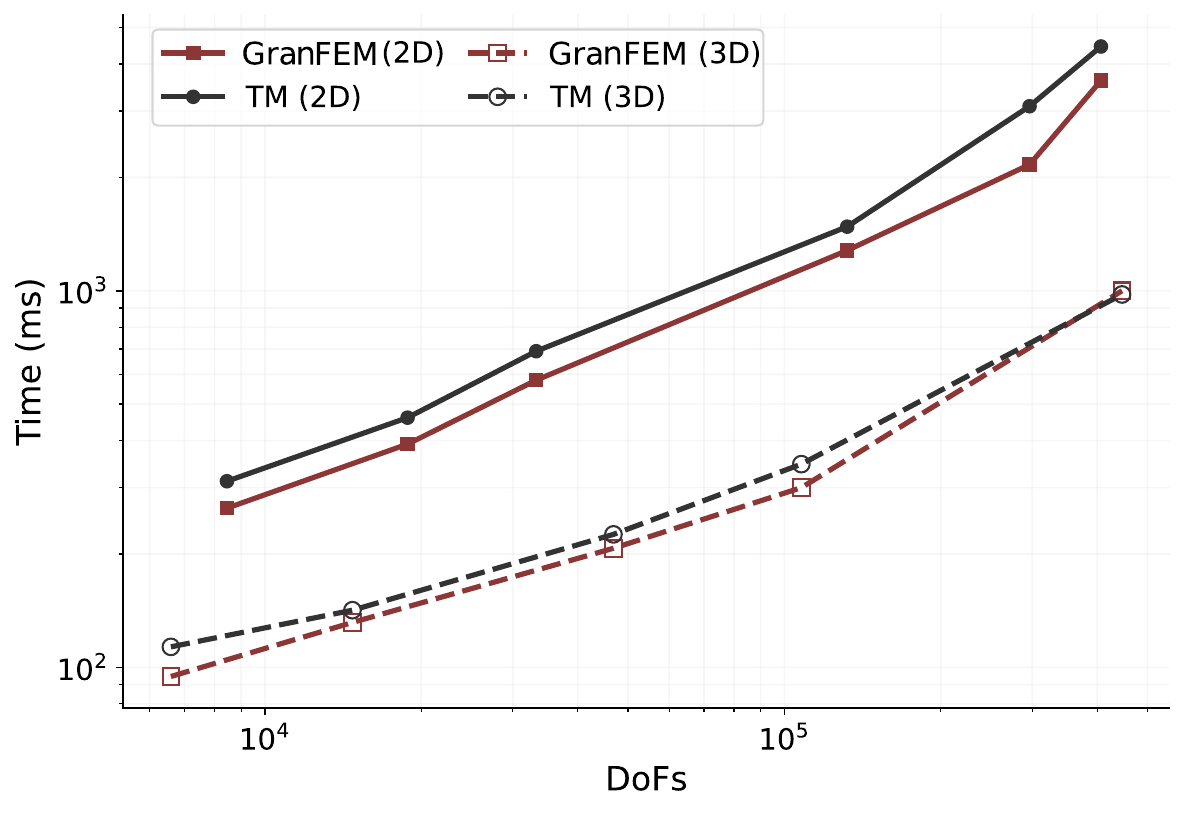}
        \caption{Elasticity solution time}
        \label{fig:Linear_Solve_Time_Elasticity}
    \end{subfigure}
    \begin{subfigure}[b]{0.32\columnwidth}
        \centering
        \includegraphics[width=\columnwidth]{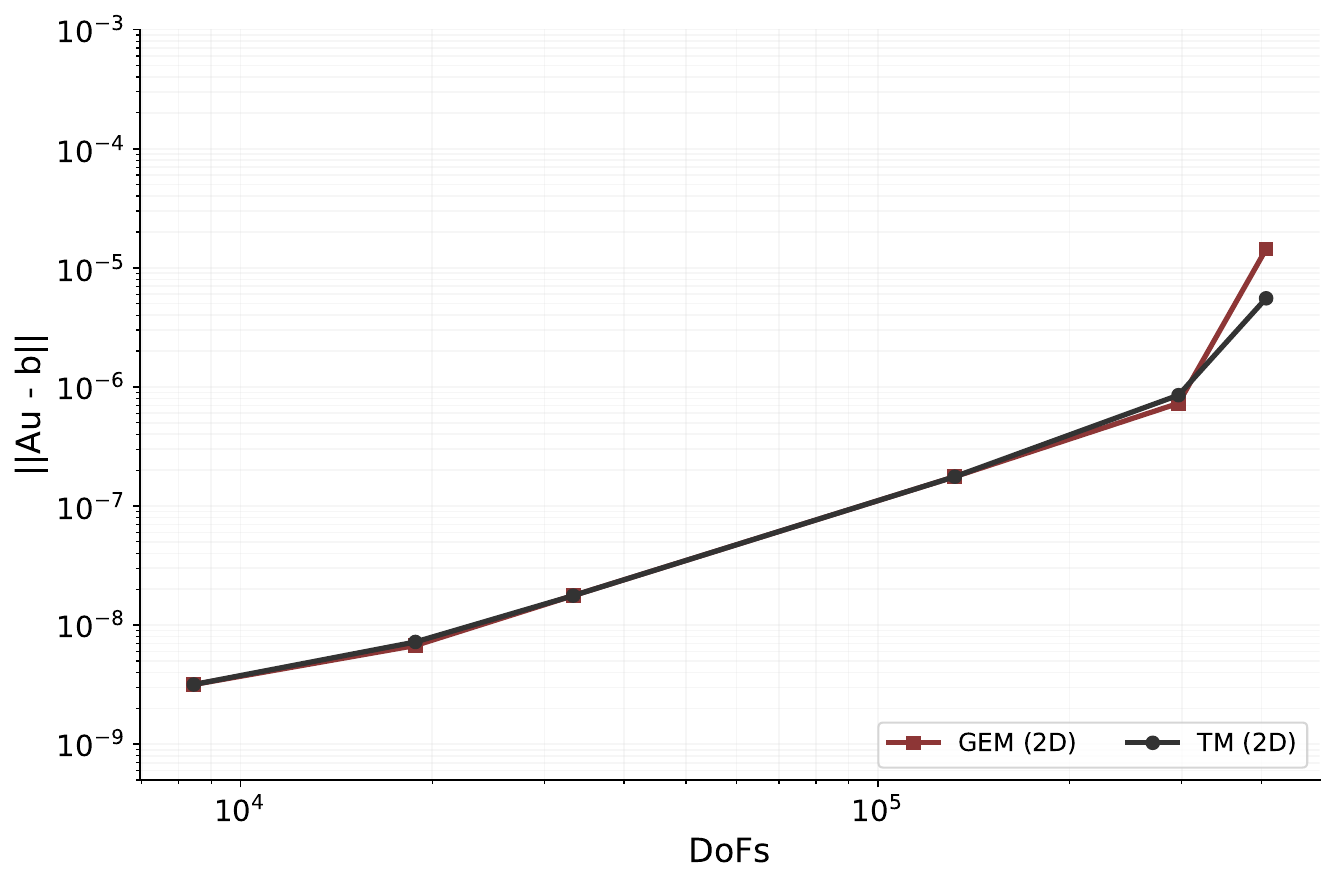}
        \caption{Elasticity residual 2D}
        \label{fig:Residual_Elasticity_2D}
    \end{subfigure}
    \begin{subfigure}[b]{0.32\columnwidth}
        \centering
        \includegraphics[width=\columnwidth]{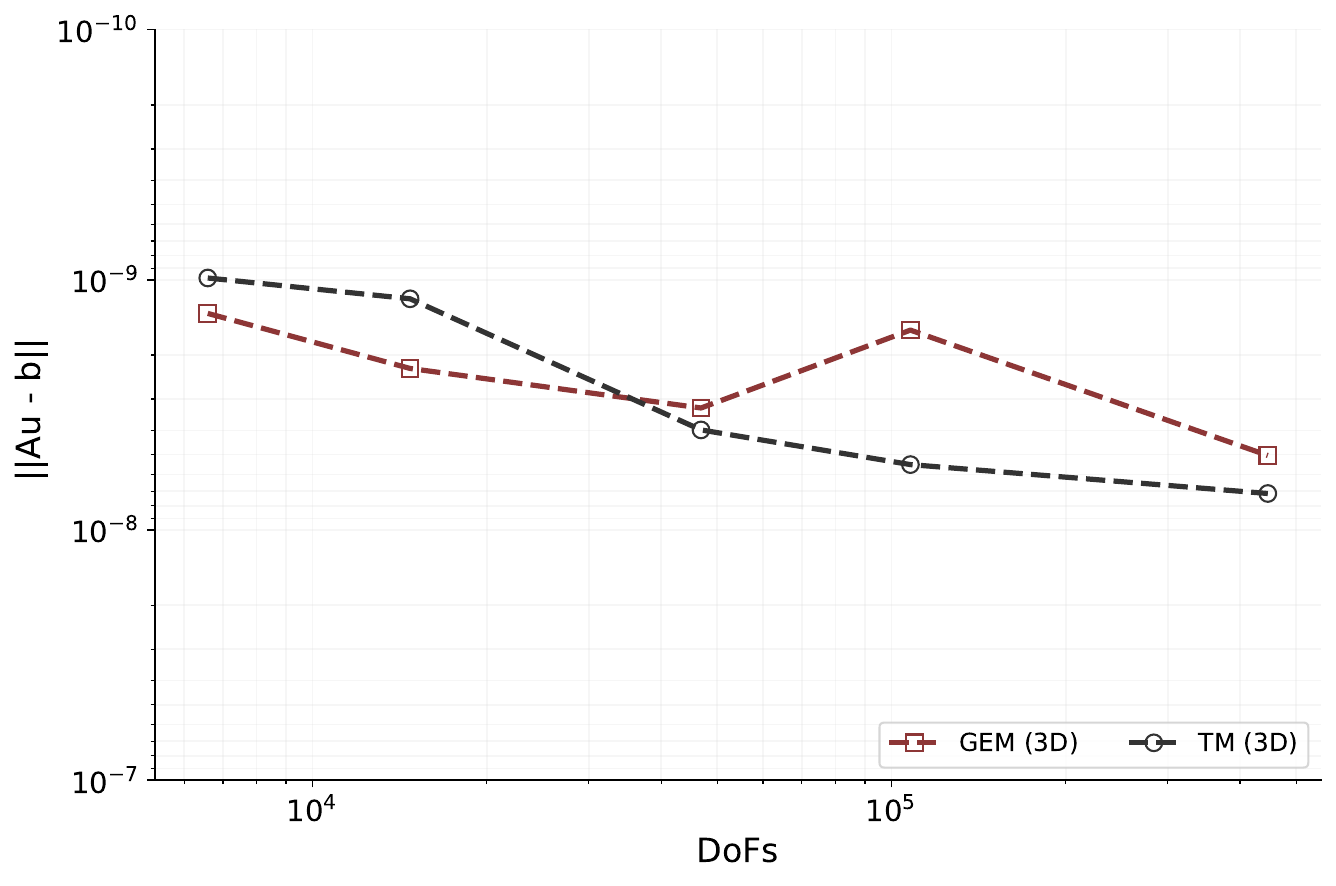}
        \caption{Elasticity residual 3D}
        \label{fig:Residual_Elasticity_3D}
    \end{subfigure}
    \caption{Graph-driven linear system resolution. Solution time and final residual $\|\mathbf{K}\mathbf{U}-\mathbf{F}\|$ of \modelname  and TM for Poisson (CG) and linear elasticity (BiCGSTAB). }
    \label{fig:Graph_Solve_Times_and_Residuals}
\end{figure*}

\begin{figure*}[t]
    \centering

    \begin{subfigure}[b]{0.35\textwidth}
        \centering
        \includegraphics[width=\textwidth]{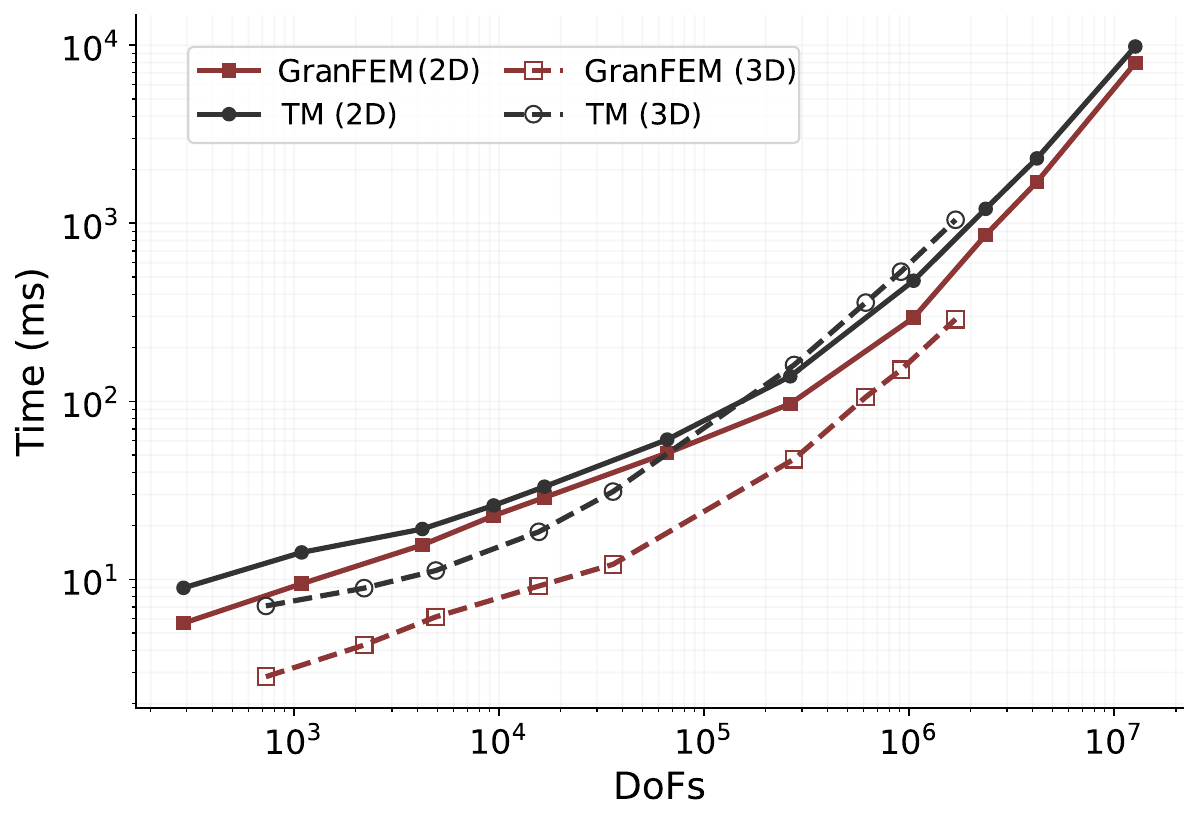}
        \caption{Poisson runtime}
        \label{fig:poisson-runtime}
    \end{subfigure}
    \hfill
    \begin{subfigure}[b]{0.35\textwidth}
        \centering
        \includegraphics[width=\textwidth]{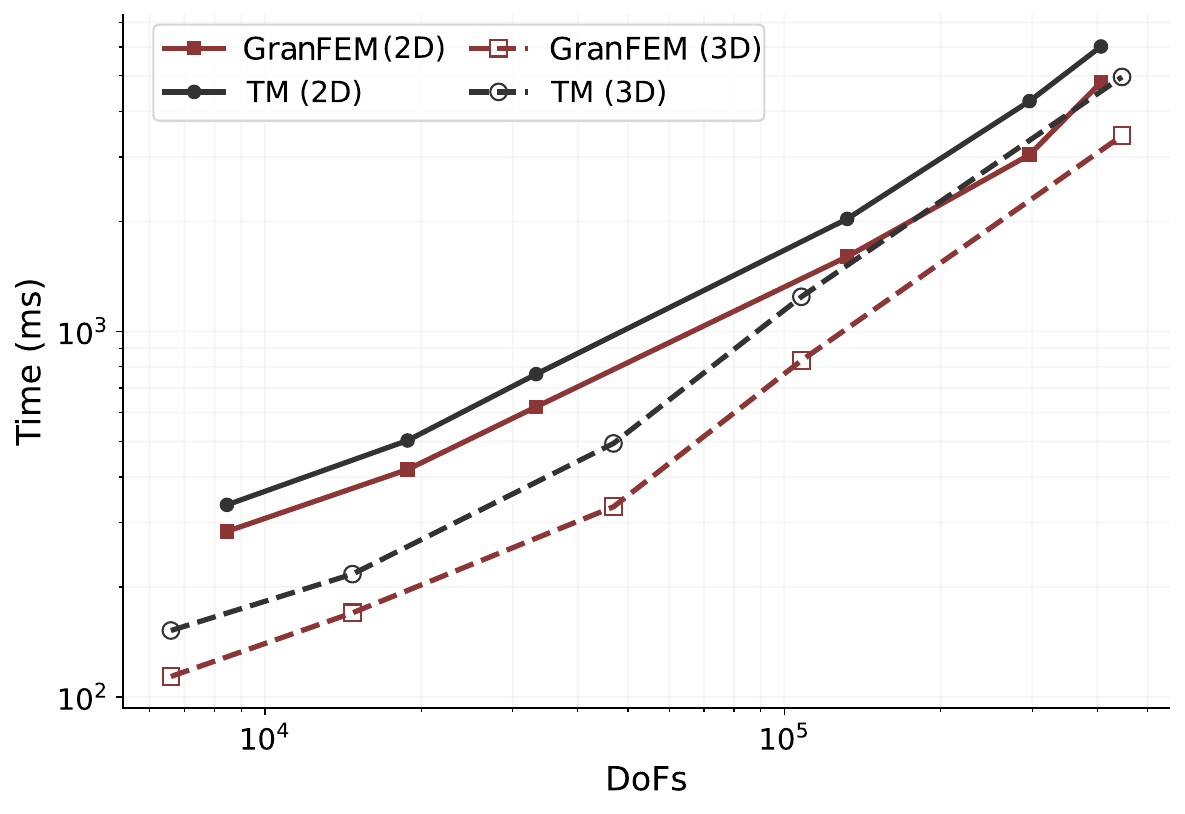}
        \caption{Elasticity runtime}
        \label{fig:elasticity-runtime}
    \end{subfigure}
    \hfill
    \begin{minipage}[b]{0.12\textwidth}
        \centering
        \captionsetup{font=footnotesize}
        \begin{subfigure}[b]{\textwidth}
            \centering
            \includegraphics[width=\textwidth]{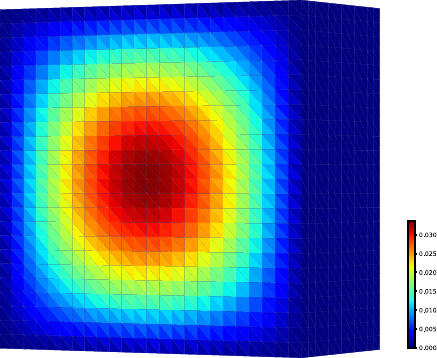}
            \caption{Poisson 3D }
            \label{fig:poisson-solution}
        \end{subfigure}

        \vspace{0.5em}

        \begin{subfigure}[b]{\textwidth}
            \centering
            \includegraphics[width=\textwidth]{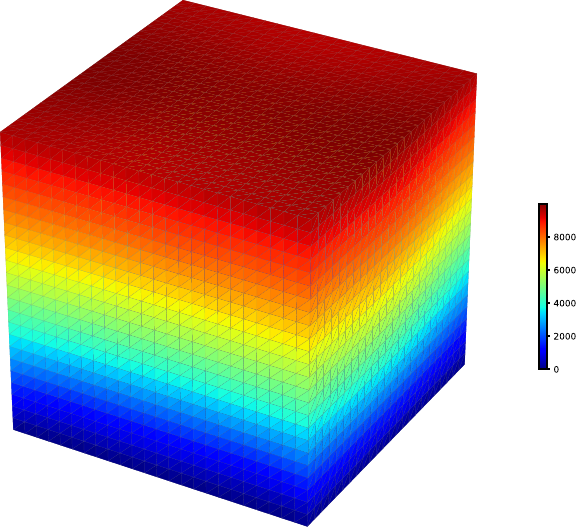}
            \caption{Elasticity 3D}
            \label{fig:elasticity-solution}
        \end{subfigure}

    \end{minipage}

    \vspace{-0.75em}
    \caption{ Runtime scaling of \modelname  and TensorMesh (TM) with the number of DoFs for (a) Poisson and (b) linear elasticity, with (c \& d) representative 3D solution fields. }
    \label{fig:pde-benchmarks}
\end{figure*}

 In Figures~\ref{fig:Linear_Solve_Time_Poisson}-\ref{fig:Residual_Poisson_3D}  and ~\ref{fig:Linear_Solve_Time_Elasticity}-\ref{fig:Residual_Elasticity_3D} we present the linear system solution time and the 2D/3D residuals attained at the end of the optimization per DoF for the Poisson and elasticity cases, respectively. The residuals are broadly identical (small fluctuations are observed for the elasticity due to the sensitivity of BiCGSTAB to tiny floating-point perturbations \citep{sleijpen1995BiCGstabconvergance}), evincing once again the equivalence of the two approaches, as well as their potential to yield accurate PDE solutions. Note that for the elasticity case, we provide the residuals only for up to $\approx 5 \cdot 10^5$ DoFs, since BiCGSTAB did not converge within the maximum iteration range (for more than $\approx 2 \cdot 10^5$ DoFs, we also observed that the two approaches converged at different iteration, which is again attributed to the sensitivity of BiCGSTAB).  The linear system solution time is modestly faster for \modelname across both PDEs and dimensions. For Poisson, there is negligible difference (only for the 3D case and larger DoFs, an approximately 8\% improvement is observed), whilst for elasticity, the gap is more noticeable with a speedup of roughly 10-20\%.  Finally, in Figures~\ref{fig:poisson-runtime} and~\ref{fig:elasticity-runtime} we show how the overall computation time of TM and \modelname scales with the number of DoFs for the Poisson and elasticity equations, respectively. The reported values correspond to the time required for the local element matrix estimation, the assembly, and the linear system resolution (setup times remain largely the same), as discussed earlier. Overall, the \textit{graph-driven} formulation expedites the computational pipeline, primarily through PDE-specific assembly kernels and by avoiding explicit sparse-matrix construction.

\subsection{Complementary Discussion} \label{sec:Appendix_Complementary_Discussion}


The significance of \modelname lies primarily in the representation rather than in accelerating an individual FEM operation. We show that the principal stages of the finite element pipeline can be posed within a common graph abstraction: local variational interactions define edge contributions, assembly becomes routing and aggregation (as in \citep{wen2026tensorgalerkin}), sparse operator application becomes message passing, and iterative solution repeatedly applies the resulting graph operator. Consequently, the graph is not merely constructed from a FEM mesh or used as the input
to a downstream neural model; it becomes the computational representation of the problem itself. Importantly, the same abstraction then provides a natural interface for learning-based problems, supporting hybrid solutions by exploiting physically meaningful node connections and operator coefficients, around which it only requires learning local corrections, rather than replacing the numerical discretization with an unrestricted surrogate.

At the same time, the graph-native pipeline reduces the runtime of the numerical FEM
experiments while reproducing the conventional solution. The majority of this
gain originates from operator construction and assembly rather than from the linear solver. As mentioned earlier, two implementation choices explain this behavior. First, \modelname evaluates the local interactions through PDE-specific kernels, whereas TM retains the flexibility of a generic variational-form implementation evaluated through nested \texttt{vmap} operations. Second, after local-to-global routing, \modelname directly retains the coefficients $K_{ij}$ as graph edge weights. Since this graph is the native representation used by subsequent computations, there is no need to additionally materialize the assembled operator as a sparse matrix or construct the corresponding sparsity hash representation. These differences reduce assembly overhead, whereas the \textit{graph-driven} linear solve provides only a smaller additional gain, as both approaches ultimately rely on efficient sparse matrix--vector operations.

We therefore regard these results as evidence that adopting a graph-native representation does not introduce a computational penalty and can, in fact, remove some overhead of the conventional pipeline, rather than as the central contribution of \modelname.  Indeed,  the resulting savings for the static problems considered here remain on the order of milliseconds, and their importance is consequently secondary to the broader outcome: the same representation that exactly realizes conventional FEM computation can be retained when transitioning from numerical simulation to physics-consistent operator learning, without introducing a separate computational abstraction.

\section{Operator Learning}

\subsection{Problem Setup and Dataset Generation} 
\label{sec:Appendix_Operator_Learning_Setup}

\subsubsection{PDE Benchmarks} \label{sec:Time_Dependent_PDEs}

We consider three time-dependent benchmark PDEs: the wave equation on a disk, the Allen--Cahn equation on an L-shaped domain, and the advection--diffusion equation on the unit square. All three are initialized with the multi-frequency sine expansion:

\begin{equation}
u_0(x,y) = \frac{\pi}{K^2}\sum_{i,j=1}^{K} a_{ij}\,(i^2+j^2)^{-r}\sin(\pi i x)\sin(\pi j y),
\label{eq:ic}
\end{equation}

with $a_{ij}$ sampled i.i.d.\ from $\mathcal{U}[-1,1]$; $K=6$ and $r=0.5$ for the wave and AC equations, whereas for AD we use $K=12$ and $r=0.25$. The expansion is evaluated at all nodes, and the Dirichlet nodes are then set to zero. The meshes and datasets are described in Appendix~\ref{sec:Dataset_Generation}.

\paragraph{Hyperbolic PDE: Wave Equation.} We consider the 2D wave equation on $\mathcal{D} \times [0, T]$:

\begin{equation}
\partial_{tt} u = c^2 \Delta u, \quad \text{in } \mathcal{D} \times [0, T],
\label{eq:wave}
\end{equation}

 Since the wave equation is second-order in time, it additionally requires an initial velocity, which we set to a constant, $\partial_t u(\cdot,0)=1$, as in the generator of \citet{wen2026tensorgalerkin}. The boundary values remain fixed by the Dirichlet condition, whilst the initial velocity enters only through the first time step, as in the reference generator. Multiplying  (\ref{eq:wave}) by the test function  $v = \phi_i$ and integrating the Laplacian term by parts for a valid FEM solution adhering to  (\ref{eq:fem_ansatz}), yields the semi-discrete system: 

\begin{equation}
    \mathbf{M}\ddot{\mathbf{U}}(t)+c^2\mathbf{K}\mathbf{U}(t)=\mathbf{0},
    \label{eq:wave_semi_disc_system}
\end{equation}

where $M_{ij}=\int_D\phi_i\phi_j\,\mathrm{d}\mathbf{x}$ and $K_{ij}=\int_D\nabla\phi_i\cdot\nabla\phi_j\,\mathrm{d}\mathbf{x}$ denote the FEM mass and stiffness matrices, respectively, which can be computed in a \textit{graph-driven} manner as per Section~\ref{sec:graph_assembly} and Appendix~\ref{sec:Appendix_Example_Graph_Driven_Framework}. Approximating the temporal derivative through a centered finite difference scheme:

\begin{equation}
\dot{\mathbf{U}}(t_n) \approx \frac{\mathbf{U}^{n+1}-\mathbf{U}^{n-1}}{2\Delta t}, \qquad \ddot{\mathbf{U}}(t_n) \approx \frac{\mathbf{U}^{n+1}-2\mathbf{U}^{n}+\mathbf{U}^{n-1}}{\Delta t^{2}}.
    \label{eq:finite_differences}
\end{equation}

results in the \emph{Explicit} update:

\begin{subequations}\label{eq:wave_explicit}
\begin{align}
    \mathbf{M}\mathbf{U}^{n+1}
    &= \left(2\mathbf{M}-c^2\Delta t^2\mathbf{K}\right)\mathbf{U}^n
    -\mathbf{M}\mathbf{U}^{n-1}   \quad \Longrightarrow
    \label{eq:wave_fem_step}\\[4pt]
    \mathbf{U}^{n+1}
    &= \left(2\mathbf{I}-c^2\Delta t^2\mathbf{M}^{-1}\mathbf{K}\right)\mathbf{U}^n
    -\mathbf{U}^{n-1}.
    \label{eq:wave_explicit_step}
\end{align}
\end{subequations}

from which one can deduce that $\mathbf{A}_{\mathrm{wave}} = 2\mathbf{I} - c^2\Delta t^2\mathbf{M}^{-1}\mathbf{K}$ and $\mathbf{B}_{\mathrm{wave}} = -\mathbf{I}$ in the notation of (9). However, since $\mathbf{M}^{-1}$ is dense, typically it is never formed, and instead of emulating the trajectory via consecutive matrix multiplications via~(\ref{eq:wave_explicit_step}),
 at every time step, the RHS of (\ref{eq:wave_fem_step}) is evaluated and the system $\mathbf{M}\mathbf{U}^{n+1}=\mathbf{b}^n_{\text{wave}}$  is solved with a Jacobi-preconditioned conjugate gradient method.


\paragraph{Parabolic PDE: Allen--Cahn Equation.} We further consider the Allen--Cahn equation:
\begin{equation}
\partial_t u = \nabla \cdot (a^2 \nabla u) - \varepsilon^2 u(u^2 - 1), \quad \text{in } \mathcal{D} \times (0, T),
\label{eq:ac}
\end{equation}

\vspace{-2mm}

with homogeneous Dirichlet boundary conditions and  the same initial condition we considered for the wave equation, \ie multi-frequency sine with the specified hyperparameters. For a FEM solution following the ansatz of  (\ref{eq:fem_ansatz}), one can define the assembled reaction vector:

\vspace{-2mm}

\begin{equation}
     [\mathbf{F}(\mathbf{U})]_j
    =
    \int_D
    \phi_j
    \left[-\varepsilon^2u_h(U)\left(u_h(U)^2-1\right)\right]
    \,\mathrm{d}\mathbf{x},
    \label{eq:ac_reaction_vector}
\end{equation}

and via a spatial Galerkin discretization derive, similarly to the wave equation, the semi-discrete nonlinear system:

\begin{equation}
    \mathbf{M}\dot{\mathbf{U}}(t)+a^2\mathbf{K}\mathbf{U}(t)-\mathbf{F}(\mathbf{U}(t))=\mathbf{0}.
    \label{eq:ac_semidiscrete}
\end{equation}

To solve (\ref{eq:ac_semidiscrete}) through an \emph{Explicit} scheme,  a forward-Euler discretization can be employed yielding:

\begin{subequations}\label{eq:ac_direct}
\begin{align}
    \mathbf{M}\mathbf{U}^{n+1}
    &= \left(\mathbf{M}-a^2\Delta t\mathbf{K}\right)\mathbf{U}^n
    +\Delta t\mathbf{F}(\mathbf{U}^n)   \quad \Longrightarrow
    \label{eq:ac_fem_step}\\[4pt]
    \mathbf{U}^{n+1}
    &= \left(\mathbf{I}-a^2\Delta t\mathbf{M}^{-1}\mathbf{K}\right)\mathbf{U}^n
    +\Delta t\mathbf{M}^{-1}\mathbf{F}(\mathbf{U}^n),
    \label{eq:ac_explicit_step}
\end{align}
\end{subequations}

which computes the corresponding explicit discrete trajectory, but its stable time step, $\Delta t$ is restricted by the spectral properties of the spatial discretization. Therefore, it is  subject to a more restrictive stability condition than the \emph{Implicit}  formulation which uses a backward-Euler discretization:

\begin{equation}
    \mathbf{M}\mathbf{U}^n=\left(\mathbf{M}+\Delta t\,a^2\mathbf{K}\right)\mathbf{U}^{n+1}
    -\Delta t\,\mathbf{F}(\mathbf{U}^{n+1}),
    \label{eq:ac_iterative}
\end{equation}

which, unlike  (\ref{eq:ac_fem_step}), does not admit an explicit linear evolution operator, since the unknown state $\mathbf{U}^{n+1}$ also appears inside the nonlinear reaction vector. Likewise, the nonlinearity introduced by the reaction term does not allow solving  (\ref{eq:ac_iterative})  as a single linear system. Instead, the implicit update can be cast as a root-finding problem for the residual $ R(\mathbf{U}^{n+1}) := (\mathbf{M}+\Delta t\,a^2\mathbf{K})\mathbf{U}^{n+1}-\Delta t\,\mathbf{F}(\mathbf{U}^{n+1})-\mathbf{M}\mathbf{U}^n $. This  residual can then be driven  to 0, for example, by using Newton's method. At each Newton iteration, the residual is linearized about the current iterate and a sparse linear system involving its Jacobian is solved using a linear algebra solver. Consequently, an implicit time step may require multiple sparse linear solves, one for each Newton iteration, rather than the single explicit operator evaluation required by (\ref{eq:ac_explicit_step}). This additional computational cost is accompanied by substantially improved stability compared with the fully explicit update, allowing time steps beyond the more restrictive stability regime of forward Euler.

\textbf{Parabolic PDE with transport: Advection--Diffusion Equation.} We finally consider the advection--diffusion equation:

\begin{equation}
\partial_t u + \mathbf{b}\cdot\nabla u = \kappa\,\Delta u, \quad \text{in } \Omega\times(0,T],
\label{eq:ad}
\end{equation}

on the unit square, $\Omega = (0,1)^2$, with homogeneous Dirichlet boundary conditions and diffusivity $\kappa = 0.03$. The transport is driven by the steady cellular flow $\mathbf{b}(x,y) = b_0\left(\sin(\pi x)\cos(\pi y),\, -\cos(\pi x)\sin(\pi y)\right)$, with $b_0 = 1$, \ie the
velocity field induced by the stream function $\psi = (b_0/\pi)\sin(\pi x)\sin(\pi y)$, which is divergence-free and tangential to the boundary, $\mathbf{b}\cdot\mathbf{n} = 0$ on $\partial\Omega$. The initial condition follows (\ref{eq:ic}) with $K=12$ and $r=0.25$, rescaled to $\max|u_0| = 1$. Multiplying (\ref{eq:ad}) by the test function $v=\phi_i$ and integrating the diffusion term by parts yields the semi-discrete system:

\begin{equation}
\mathbf{M}\dot{\mathbf{U}}(t) + \left(\kappa\mathbf{K} + \mathbf{T}\right)\mathbf{U}(t) = \mathbf{0},
\qquad
T_{ij} = \int_\Omega \phi_i\,\left(\mathbf{b}\cdot\nabla\phi_j\right)\mathrm{d}\mathbf{x},
\label{eq:ad_semidiscrete}
\end{equation}

where $\mathbf{M}$ and $\mathbf{K}$ are the mass and stiffness matrices of  (\ref{eq:wave_semi_disc_system}), and $\mathbf{T}$ is the advection matrix. Note that, since $\nabla\cdot\mathbf{b} = 0$ and $\mathbf{b}\cdot\mathbf{n} = 0$, $\mathbf{T}$ is skew-symmetric; hence, advection only redistributes the solution, whilst its decay stems solely from
diffusion. Contrary to the wave and AC equations, the operator is thus not symmetric, and the update of each node depends on the direction of the flow along every incident edge. Applying a forward-Euler discretization yields:

\begin{subequations}\label{eq:ad_euler}
\begin{align}
    \mathbf{M}\mathbf{U}^{n+1}
    &= \left(\mathbf{M} - \Delta t\,(\kappa\mathbf{K} + \mathbf{T})\right)\mathbf{U}^{n}
    \quad \Longrightarrow
    \label{eq:ad_fem_step}\\[4pt]
    \mathbf{U}^{n+1}
    &= \left(\mathbf{I} - \Delta t\,\mathbf{M}^{-1}(\kappa\mathbf{K} + \mathbf{T})\right)\mathbf{U}^{n}.
    \label{eq:ad_explicit_step}
\end{align}
\end{subequations}


\textbf{Remark.} The reference solvers expose the computational difficulty that motivates operator learning with \modelname. Although $\mathbf{M}$ and $\mathbf{K}$ are sparse, the inverse mass matrix appearing in the explicit updates is dense: forming it is infeasible beyond a few tens of thousands of DoFs, so explicit schemes solve a linear system with $\mathbf{M}$ at every step, and implicit schemes solve one at every Newton iteration of every step. Long trajectories therefore require thousands of sparse linear solves, whereas \modelname advances the solution with one or two sparse products per step.


\subsubsection{Dataset generation} \label{sec:Dataset_Generation}

For every PDE, we generate $32$ trajectories of $400$ time steps from independent initial conditions, split into $20$ training, $4$ validation, and $8$ held-out test trajectories, using the first $200$ steps as the ID regime and the remaining $200$ steps as OOD.

\paragraph{Wave.} The domain is a disk of radius $1$ centred at the origin, discretized with a triangular mesh of characteristic element size $h=0.08$ ($633$ nodes, $554$ interior), with homogeneous Dirichlet conditions on the whole boundary, and the wave speed parameter $c$ set to 4. The mass and stiffness matrices are assembled with a three-point quadrature rule, and (\ref{eq:wave_fem_step}) is advanced with $\Delta t = 5\cdot10^{-4}$, solving the mass-matrix system at every step with a Jacobi-preconditioned conjugate gradient method (relative tolerance $10^{-10}$, $\approx 16$ iterations per step).

\paragraph{Allen--Cahn.} The domain is an L-shaped region obtained by removing the upper-right quarter $[0.5,1]\times[0.5,1]$ from the unit square, discretized with $h=0.055$ ($390$ nodes), with homogeneous Dirichlet conditions on the entire boundary, including the re-entrant corner, and the AC parameters set to $a = 1$ and $\varepsilon = 5$. For the \emph{Explicit} dataset, $\mathbf{M}$ is factorized once (dense LU decomposition) and (\ref{eq:ac_explicit_step}) is advanced with $\Delta t = 10^{-5}$, as required by the stability of the explicit scheme. For the \emph{Implicit} dataset, (\ref{eq:ac_iterative}) is advanced with $\Delta t = 10^{-4}$ using the backward-Euler Newton solver of \citet{wen2026tensorgalerkin}, with a sparse direct solve at every Newton iteration.

\paragraph{Advection--diffusion.} The unit square is discretized with $h = 0.055$ ($470$ nodes, $\mathrm{Pe}_h = 0.92$), and (\ref{eq:ad_fem_step}) is advanced with $\Delta t = 2.5\cdot10^{-3}$, solving the mass-matrix system at every step with a Jacobi-preconditioned conjugate gradient method (relative tolerance $10^{-10}$). To avoid the spurious oscillations of the standard Galerkin discretization of advection-dominated problems without resorting to
stabilization, the mesh P\'eclet number, $\mathrm{Pe}_h = |\mathbf{b}|\,h/(2\kappa)$, is kept below one.

\subsection{Model Implementation Details} 
 
\label{sec:Appendix_Operator_Learning_Model_Details}

\paragraph{Lumped.} The PDE trajectory is computed by directly advancing (\ref{eq:wave_explicit_step}),  (\ref{eq:ac_explicit_step}), and (\ref{eq:ad_explicit_step})  autoregressively. However, instead of using the actual inverse mass matrix, $\mathbf{M}^{-1}$, a lumped version is exploited. In particular, initially,  $ \mathbf{M}_{\text{lump}} $ is constructed by row-summing the mass matrix into a diagonal matrix, \ie $ (M_{\mathrm{lump}})_{ii} = \sum_{j=1}^{N} M_{ij}, \  (M_{\mathrm{lump}})_{ij} = 0 \quad \text{if } i \neq j$. Then, one can straightforwardly compute the inverse of the diagonal $ \mathbf{M}^{-1}_{\text{lump}}$  through inexpensive node-wise divisions rather than a sparse linear solve required for the actual $\mathbf{M}^{-1}$, \eg via LU decomposition. Consequently,  the coefficient matrices appearing in (\ref{eq:wave_explicit_step}),  (\ref{eq:ac_explicit_step}), and (\ref{eq:ad_explicit_step}), \eg $\mathbf{A}_{\text{wave}}$, can be computed once in advance, and then the PDE trajectory is emulated via sparse matrix multiplications between the coefficients and solution matrices. That yields a simple  baseline that can be used to approximate the actual solution without requiring any training or having any severe computational bottlenecks.

\paragraph{\modelname.} Based on the previous Section analysis,  we derive the physics-consistent baseline coefficient matrices $\mathbf{A}', \mathbf{B}', \mathbf{C}', \mathbf{D}', \mathbf{E}'$ and their corresponding graph representations used in the implementations of Section~\ref{sec:Results_Op_learning}.

For the \emph{\textbf{wave equation}}  the baseline coefficient matrices  can  be straightforwardly deduced by  juxtaposing  (\ref{eq:wave_explicit_step}) to    the general polynomial   message-passing equation (\ref{eq:general_polynomial_update}).   Replacing the consistent inverse mass matrix by its lumped counterpart gives:

\[
\mathbf{A}'
=
2\mathbf{I}
-
c^2\Delta t^2\mathbf{M}_{\mathrm{lump}}^{-1}\mathbf{K},
\qquad
\mathbf{B}'=-\mathbf{I},
\qquad
\mathbf{C}'=\mathbf{D}'=\mathbf{E}'=\mathbf{0}.
\]

Consequently,  $\mathbf{A}'$ inherits the sparsity pattern of the stiffness matrix $\mathbf{K}$, complemented by self-loop connections arising from the identity contribution  $2\mathbf{I}$,  with edge weights $A'_{ij}$. The second term, $\mathbf{B}' = -\mathbf{I}$, is purely diagonal and known in advance; we therefore leave $\mathbf{B}'$ fixed rather than learning a correction over it. Hence, \modelname  learns a single correction operator $\mathcal{S}_\theta : \mathcal{G}_{\mathbf{A}'} \to \mathcal{G}_{\mathbf{A}}$,  which transforms the baseline edge weights $A'_{ij}$ into corrected weights $A_{ij}$ while preserving the prescribed sparse graph support.

For the \emph{\textbf{Allen--Cahn equation}} the explicit scheme of (\ref{eq:ac_explicit_step}), includes the assembled non-linear reaction term $\mathbf{F}(\mathbf{U}^n)$, and therefore does not immediately take the polynomial operator form of (\ref{eq:general_polynomial_update}).  To obtain a simple graph-compatible baseline, we approximate the assembled reaction contribution by the pointwise nodal reaction $\mathbf{F}_{\text{pt}}(\mathbf{U}^n) = \varepsilon^2\mathbf{U}^n - \varepsilon^2(\mathbf{U}^n)^{\odot 3}$, and re-write (\ref{eq:ac_direct}) as:

\begin{equation}
\mathbf{U}^{n+1}=\mathbf{U}^n-\Delta t\,a^2(\mathbf{M}^{-1}\mathbf{K})\mathbf{U}^n+\Delta t\,\varepsilon^2\mathbf{U}^n-\Delta t\,\varepsilon^2(\mathbf{U}^n)^{\odot 3}.
\label{eq:ac_gem_baseline}
\end{equation}

Collating the expression above with (\ref{eq:general_polynomial_update}) yields the following baseline approximation coefficient matrices:

\[
\mathbf{A}' = \mathbf{I} - \Delta t\,a^2 \mathbf{M}_{\mathrm{lump}}^{-1}\mathbf{K} + \Delta t\,\varepsilon^2\mathbf{I},
\qquad
\mathbf{D}' = -\Delta t\,\varepsilon^2\mathbf{I}, 
\qquad
 \mathbf{B}' = \mathbf{C}' = \mathbf{E}' = \mathbf{0}.
\]

  Again,   $\mathbf{A}'$  inherits $\mathbf{K}$'s sparsity pattern, similarly to the wave equation, along with self-loop connections, while $\mathbf{D}'$ consists solely of self-loops, since the cubic reaction term is evaluated pointwise and therefore couples each node only to itself. Hence,  \modelname learns corrections $\mathcal{S}_\theta^A : \mathcal{G}_{\mathbf{A}'} \to \mathcal{G}_{\mathbf{A}}$ and $\mathcal{S}_\theta^D : \mathcal{G}_{\mathbf{D}'} \to \mathcal{G}_{\mathbf{D}}$ over these two supports (note that $D'$ is also learnable to compensate for the reaction term approximation introduced), leaving $\mathbf{B}', \mathbf{C}', \mathbf{E}'$ at zero.  Importantly, (\ref{eq:ac_gem_baseline}) defines a \emph{physics-consistent initialization} for \modelname, without constraining the learned model to reproduce only the forward-Euler discretization from which it is derived. Indeed, as shown in Section~\ref{sec:Results_Op_learning},  when \modelname is trained with trajectories generated by the \emph{Implicit} solver (which does not admit the general form of (\ref{eq:general_polynomial_update})), the learned corrections can compensate not only for the approximations introduced by mass lumping and the pointwise treatment of the nonlinear reaction, but also for the discrepancy between the \emph{Explicit}  and \emph{Implicit} reference solution evolution. Hence, \modelname remains an explicit autoregressive operator at inference time, requiring no nonlinear root-finding or inner linear-system solves, while learning an effective state-to-state evolution map from trajectories produced by the more stable implicit discretization.

For the \emph{ \textbf{advection--diffusion equation}}, replacing $\mathbf{M}^{-1}$ in (\ref{eq:ad_explicit_step}) by its lumped counterpart gives \[\mathbf{A}' = \mathbf{I} - \Delta t\,\mathbf{M}_{\mathrm{lump}}^{-1}\left(\kappa\mathbf{K} + \mathbf{T}\right), \qquad \mathbf{B}' = \mathbf{C}' = \mathbf{D}' = \mathbf{E}' = \mathbf{0}.\] Since $\mathbf{T}$ is skew-symmetric, $\mathbf{A}'$ is not symmetric, and the weight of every edge depends on its orientation; the edge features are therefore complemented by the advection entries $T_{ij}$ and $T_{ji}$, which encode the flow direction along each edge, and \modelname learns a single correction $\mathcal{S}_\theta : \mathcal{G}_{\mathbf{A}'} \rightarrow \mathcal{G}_{\mathbf{A}}$.

The correction network $\mathcal{S}_\theta$ is a plain MLP with SiLU activations between hidden layers and no activation on the output. Its final linear layer is explicitly zero-initialized, so at initialization the model reduces exactly to the mass-lumped FEM baseline, and thus the training process learns a correction over  a numerically valid starting point. For all PDEs, $\mathcal{S}_\theta$ has four linear layers (three hidden layers of width $256$) and  for the baseline coefficient matrix $Q$ it predicts a single scalar correction per edge, applied as  $ Q_{\theta}  =    Q'   +    \alpha_{Q'} \mathcal S_{\theta} \left(   Q',  \mathbf z \right)$, where $\mathbf z_{ij}$ conveys static and node features independent of the time step.
Specifically, the edge feature vector $z_{ij}$ comprises: ($i$) the normalized baseline edge weight(s) (for AC, both $A'_{ij}$ and $D'_{ii}$ enter as separate normalized weights), ($ii$) the rescaled stiffness and ($iii$) mass entries $K_{ij}$, $M_{ij}$, ($iv$--$v$) the lumped mass at both node endpoints, 
($vi$) the normalized edge displacement and ($vii$) length, ($viii$) the timestep $\Delta t$, ($ix$)   the relevant PDE constants ($c$ for wave; $a, \varepsilon$ for AC; $\kappa$ and the advection entries $T_{ij}, T_{ji}$ for AD), and ($x$) a self-loop indicator flagging diagonal entries; all z-score normalized across edges.  The values of the scaling correction parameter $\alpha$ are set to $\alpha_{\mathbf{A}'} = 10^{-4}$ for the wave equation, whilst $\alpha_{\mathbf{A}'} = \alpha_{\mathbf{D}'} = 1$ for AC, and $\alpha_{\mathbf{A}'} = 1$ for AD. Finally, aside from the graph support induced by the lumped $Q'$, we extend the connectivity of each node to include the one-hop support to compensate for connections lost due to the lumped simplification in $Q'$. 

Moreover, considering that the correction predicted by the network is added directly to the baseline discrete operator's edge weights, its action should preserve a fundamental consistency property of the underlying Laplacian-like operator independently of the mesh on which it is evaluated. Specifically, each row of the correction must sum to zero or, equivalently, the correction must vanish on constant fields. Without enforcing this constraint, the learned correction can become implicitly calibrated to the operator scale of the training mesh. In particular, the mass-normalized discrete Laplacian $\mathbf{M}^{-1}_{\text{lump}}\mathbf{K}$ scales with the mesh resolution as $h^{-2}$, such that a correction learned at a fixed resolution can become disproportionately large or small relative to the baseline operator when transferred to a substantially different mesh. We therefore enforce a zero-row-sum constraint on every learned correction ($\mathbf{A}'$, and $\mathbf{D}'$ for AC), applied also during training to non-boundary nodes: 
the correction is predicted on the full one-hop support, and its diagonal entry is set to minus the sum of the off-diagonal entries of the same row. 
For nodes adjacent to the boundary, the interior block of $\mathbf{K}$ does not have zero row sums, so the constraint does not render the corrected operator consistent on constant fields there; it nonetheless prevents, though, the correction itself from introducing a spurious, resolution-dependent zeroth-order term, which is the property that matters for the transfer across meshes.

When using \textit{\textbf{Neumann-series base operators}}, \ie  $p>0$, the lumped inverse in the base coefficient matrices is replaced by the truncated expansion $\mathbf{M}_p^{-1}=\sum_{k=0}^{p}\mathbf{S}^k\mathbf{M}_{\mathrm{lump}}^{-1}$ of Section~\ref{sec:neumann}, which affects only the terms containing $\mathbf{M}^{-1}$, \ie $\mathbf{A}'_p = 2\mathbf{I}-c^2\Delta t^2\mathbf{M}_p^{-1}\mathbf{K}$ for the wave equation, $\mathbf{A}'_p=\mathbf{I}-\Delta t\,a^2\mathbf{M}_p^{-1}\mathbf{K}+\Delta t\,\varepsilon^2\mathbf{I}$ for AC (with $\mathbf{D}'$ unchanged, as the reaction is treated pointwise), and $\mathbf{A}'_p=\mathbf{I}-\Delta t\,\mathbf{M}_p^{-1}(\kappa\mathbf{K}+\mathbf{T})$ for AD. The learned correction, $\mathcal{S}_\theta$, its edge features, and its one-hop support remain as for $p=0$, so that the Neumann terms account for the long-range part of $\mathbf{M}^{-1}$, whilst the correction compensates for the remaining local error. We note, though, that for each $p$, the correction network is trained separately with the same architecture and objective. 

The expansion can be computed in two equivalent ways. First, in a  \emph{matrix-free} manner, where every time step computes $\mathbf{M}_p^{-1}\mathbf{r}$, with $\mathbf{r}$ the corresponding RHS (\eg $\mathbf{r}=\mathbf{K}\mathbf{U}^n$ for the wave equation), through the recursion $\mathbf{z}^{(0)} = \mathbf{M}_{\mathrm{lump}}^{-1}\mathbf{r}$;  $\mathbf{z}^{(k)} = \mathbf{z}^{(k-1)} - \mathbf{M}_{\mathrm{lump}}^{-1}\mathbf{M}\,\mathbf{z}^{(k-1)}$, and $\mathbf{M}_p^{-1}\mathbf{r} = \sum_{k=0}^{p}\mathbf{z}^{(k)}$. Hence, each term requires one round of message passing on the mass-matrix graph followed by node-wise division by the lumped mass, and reaches one hop further than the previous round. Second, in a \emph{precomputed} way, where, since all operators are static, the complete corrected update is computed once before the rollout, as a single sparse matrix on the $(p{+}1)$-hop support, so that every time step requires one sparse product (two for AC) irrespective of $p$. All results of Section~\ref{sec:Results_Op_learning} use the \emph{precomputed} operator, whose rollout time is practically independent of $p$ on these meshes. Note that on large meshes and for large $p$, its memory and per-step cost grow with the number of nonzeros, and the \emph{matrix-free} evaluation may become preferable (see Appendix~\ref{sec:Appendix_Add_Res_Neuman}).

Finally, for all PDEs, \modelname is trained with the same objective: a one-step, teacher-forced loss, \ie each ground-truth state is mapped to the next one, complemented by a rollout loss, \ie the normalized mean squared error between the autoregressively emulated and the reference trajectory, with backpropagation through the entire unrolled window. The rollout loss is activated at epoch $250$ with a window of $5$ steps, which is doubled every $300$ epochs until it spans the training horizon ($200$ steps), and the gradient norm is clipped at $1$. A small penalty on the magnitude of the learned correction is added (weight $10^{-8}$).

\begin{table}[t]
\centering
\caption{Number of trainable parameters per model and PDE (in thousands).}
\label{table:parameters}
\scriptsize
\resizebox{\columnwidth}{!}{
\setlength{\tabcolsep}{10pt}
\begin{tabular}{lccccccccc}
\toprule
PDE & TensorPILS & \modelname & MGN & FEN & RIGNO & DeepONet & Transolver & Geo-FNO & Lumped \\
\midrule
AC   & 94.6  & 270.9 & 2332.2 & 29.2 & 1906.2 & 54.9  & 458.7 & 1546.5 & 0 \\
Wave & 98.4  & 135.4 & 2332.2 & 29.8 & 1906.6 & 111.0 & 458.7 & 1546.5 & 0 \\
AD   & 146.2 & 136.4 & 2332.2 & 58.5 & 1906.2 & 60.0  & 458.7 & 1546.5 & 0 \\
\bottomrule
\end{tabular}}
\end{table}

\paragraph{TensorPILS.} Implemented as a physics-informed Galerkin GNN, following \citet{wen2026tensorgalerkin}'s backbone: the mesh is converted into an element graph by connecting every pair of nodes within each element as a fully connected subgraph (for the triangular elements used throughout this work, this coincides exactly with the element's own three edges). This element graph is then processed with a GraphSAGE message-passing backbone (3 layers, hidden layer width 64, ReLU activations) with frequency-based encoder/decoder MLPs (2 layers each, $K=4$ Fourier features, matching the frequency-encoding setting used elsewhere in this framework). For all PDEs, each network call maps a window of 4 states to the next 4 (for the wave equation, together with the corresponding velocities), and the model is trained on the 200-step horizon and rolled out autoregressively to 400 steps, with a batch size of 16 (two optimizer steps per epoch over the 20 training trajectories). For AD, this backbone cannot represent the directional transport (relative $L_2$ error $0.43/0.87$); we therefore use the alternative interaction-network processor considered by the authors,  appending the nodal coordinates to the node inputs, which reduces the error to $0.11/0.19$ (Table~\ref{table:operator_results}). 

\paragraph{MeshGraphNets (MGN).}
We employ the encoder--processor--decoder graph neural network architecture introduced by \citep{pfaff2021meshgraphnets}. Adhering to the original architecture, the encoder maps the input node and edge features to a latent representation of dimension $128$, which is subsequently processed by $15$ message-passing blocks before being mapped to the output by the decoder. The encoder and processor are parameterized by two-hidden-layer MLPs with SiLU activations, followed by LayerNorm. The decoder uses the same structure but omits the final LayerNorm, following the original architecture. The graph is the mesh-edge graph induced by the finite element discretization, \ie the same connectivity used for \modelname's baseline operators, with no additional world-space or collision edges, since none of the PDEs considered here involve contact. The model is trained with a one-step loss on normalized targets, a batch size of $8$, and gradient-norm clipping at $1.0$. The form of the predicted quantity follows the two settings of \citep{pfaff2021meshgraphnets}. Specifically, ($i$) for the first-order AC and AD equations, the network receives the last two states and predicts the normalized per-step state increment, with Gaussian noise of standard deviation $0.02$ added to the input states. Then, ($ii$) for the second-order wave equation, we adopt the second-order integration used for the cloth dynamics in the original work, \ie the network receives the current state together with its finite-difference velocity, $[U^{n},\, U^{n}-U^{n-1}]$, predicts the normalized acceleration $U^{n+1}-2U^{n}+U^{n-1}$, and the next state is recovered as $U^{n+1}=2U^{n}-U^{n-1}+\hat{a}^{n}$. For the wave equation, no input noise is used, since the per-step change of the solution ($\approx 8\cdot10^{-4}$) is about $25$ times smaller than the noise level of the original fluid-dynamics recipe ($0.02$).

The node inputs comprise the state, the nodal coordinates, and a boundary flag, and the edge inputs the relative position of the two endpoints and its length. We also trained the variant without nodal coordinates, as in the original architecture: it reached a similar accuracy on the training mesh ($0.008/0.020$ vs.\ $0.006/0.010$) but did not improve the transfer to other meshes and domains, where its error increased to $1.5$--$4.3$ (ID) and $4$--$36$ (OOD) on meshes of $\approx 10^4$ nodes and we therefore retain the nodal coordinates. The effect of reducing the number of message-passing steps is reported in Appendix~\ref{sec:Appendix_Operator_Learning_Additional_Results} (see Table~\ref{table:mgn_depth}).

\paragraph{FEN.} We use the original  implementation of \citep{lienen2022fen}, configured with a
single stationary and autonomous free-form dynamics term. For AD, the free-form term is complemented by FEN's transport term, which models the advective flux between neighbouring cells. For the wave and AC equations, which contain no advection, it is omitted. The free-form term, which models the instantaneous cell-wise dynamics, is parameterized by a four-layer MLP of width $96$ with Tanh activations and zero initialization. In contrast to the discrete one-step prediction employed by the other learned baselines, FEN defines a continuous-time vector field, and predictions are obtained by integrating the resulting finite element dynamics in time using a fourth-order Runge--Kutta scheme. For the second-order-in-time wave equation, we rewrite the dynamics as a first-order system by augmenting the state with the velocity, $(U, V)$. Rather than letting the network learn both time derivatives, we build the kinematic relation $\dot{U}=V$ directly into the dynamics, such that the free-form term only needs to learn $\dot{V}$. The initial velocity is the one prescribed by the initial condition, whilst the velocity at subsequent states, required as a training target, is computed from the displacement trajectories through central differences. Following the training procedure of the original work, the model is trained on short windows starting at arbitrary time instants within the training horizon, rather than exclusively from the initial condition. At every epoch we sample random (trajectory, start time) pairs, and the window length is increased by one step per epoch, starting from $3$ target steps, up to $32$ target steps ($64$ windows per epoch). 
The loss is the $L_1$ distance between the integrated and the reference states, with a batch size of $8$, and the evaluation is always performed through a single rollout from the initial state over the full trajectory.

\paragraph{RIGNO.}
RIGNO is a multiscale graph neural operator designed to propagate information beyond the immediate neighborhood of the discretization mesh \citep{mousavi2026rigno}. Rather than performing message passing exclusively on the original physical graph, the input field is first projected from the physical nodes onto a coarsened regional interaction graph, on which a hierarchy of multiscale message-passing operations captures both local and longer-range spatial interactions. The resulting latent representation is subsequently projected back onto the physical nodes to predict
the evolved solution field. To construct the point-to-region and region-to-point mappings we use a mesh subsampling factor of $4$, with $6$ multiscale levels. Node and edge quantities are embedded into latent representations of dimension $128$, followed by $12$ processor/message-passing steps, with single-hidden-layer MLPs employed throughout
the architecture. During training, stochastic edge masking with probability $0.5$ is applied to regularize the regional interaction graph. The network is conditioned on the lead time $\tau$ and predicts the time derivative of
the state, \ie $\hat{U}(t+\tau)=U(t)+\tau\,\mathcal{F}_\theta(U(t),t,\tau)$. For the wave equation, the state is augmented with the velocity, using the same initial velocity and finite-difference targets as for FEN, and we adopt
the lead-time training of the original work: the model is trained on pairs $(t,\,t+\tau)$ with
$\tau\in\{\Delta t,\dots,K\Delta t\}$ and $K=16$, and at inference it directly predicts the $K$ subsequent states from a single input before advancing autoregressively from the last one. For AC and AD, where lead-time training did not yield any benefit, consecutive states are used ($K=1$).

\paragraph{Transolver.} We use the Physics-Attention architecture of \citep{wu2024transolver} in its irregular-mesh formulation, which requires no explicit graph connectivity: each mesh node is embedded, and the resulting features are softly grouped into a fixed number of ``slices'' by physical-state similarity, attended to in this reduced slice space, and then projected back onto the original nodes. The model uses $5$ attention layers comprising  $128$ neurons each, $8$ attention heads, $32$ slices, an MLP ratio of $1$, and no dropout. The unified positional encoding used in the original work's structured-grid benchmarks is disabled, as our meshes are unstructured triangulations handled natively by the slice mechanism. The nodal coordinates are concatenated with the input states and embedded per node before the slicing. For training, we follow the autoregressive pipeline of the original time-dependent (Navier--Stokes)
benchmark: the network receives the last $10$ states and is trained on windows of consecutive steps with a relative
$L_2$ loss accumulated over the window, using AdamW with a one-cycle learning-rate schedule (peak learning rate $10^{-3}$, weight decay $10^{-5}$), a batch size of $8$, and $500$ epochs; its validation error plateaus well within the 500 epochs (best checkpoint at epoch 260 and 360 for the wave and AD equations, respectively). We depart from the original pipeline in three aspects, each of which improved its accuracy on our datasets: the input states and the predicted quantity are normalized; the network predicts the per-step state increment rather than the next absolute state; and, within each $20$-step training window, the model's own predictions are fed back to the
input, with backpropagation through the whole window, whereas the original pipeline feeds back the ground truth.

\paragraph{Geo-FNO.} Geo-FNO \citep{geo_fno} extends the Fourier neural operator \citep{li2021fno} to irregular geometries by learning a deformation that maps the physical nodes onto a latent regular grid, on which the Fourier layers are applied, before mapping the
result back onto the nodes. We use the original implementation with $12$ Fourier modes per direction, width $32$, and a $32\times32$ latent grid,  which exceeds the nodal resolution of all training meshes, with the nodal coordinates rescaled to the unit square through the bounding box of the domain, as assumed by the original formulation. Since the original work considers steady-state problems, the model is employed autoregressively according to \citep{li2021fno}: it receives the last two states and predicts a normalized per-step quantity, with normalized inputs, and it is trained with Adam (learning rate $10^{-3}$, weight decay $10^{-4}$), a step learning-rate schedule (decay factor $0.8$ every $250$ epochs), a batch size of $8$, and gradient-norm clipping at $1.0$. 
For AC, the network predicts the state increment and is trained with a one-step loss, with Gaussian input noise of standard deviation $10^{-4}$ and $0$ for the \emph{Explicit} and \emph{Implicit} datasets, respectively, selected on the validation split. For AD, it likewise predicts the state increment, with input noise $10^{-4}$ and a one-step loss. For the wave equation, we employ the same second-order integration as for MGN, with input noise $10^{-5}$, and complement the one-step loss with a rollout loss whose window is doubled every $100$ epochs, from $5$ steps at epoch $250$ up to the full training horizon.

\paragraph{DeepONet.} DeepONet is employed as a standard data-driven neural operator that learns a direct mapping from the initial condition to the solution evaluated at arbitrary space and time locations. Unlike the autoregressive baselines, it does not predict the next state and recursively advance the solution in time. Instead, the branch network encodes the initial condition sampled over all mesh nodes, while the trunk network independently encodes continuous query coordinates $(x,y,t)$. Their latent representations are combined to directly predict the solution at each requested space--time point. Consequently, DeepONet is trained through one-shot supervised regression against points sampled from the training horizon ($200$ steps) of the reference trajectories and is likewise evaluated directly at the desired query times, without autoregressive rollout or temporal error accumulation. Both the branch and trunk networks employ three-layer MLPs with Tanh activations and hidden width $64$, producing representations in a shared $100$-dimensional latent space. During training, $32{,}768$ space--time query points are sampled per example, using a batch size of $8$.

\paragraph{Training Curriculum.} Unless otherwise noted, all methods are trained with Adam, an initial learning rate of $10^{-3}$ ($2\times10^{-4}$ for RIGNO, as in the original implementation), and a step learning-rate schedule (step size 250, decay factor 0.8), on the same train/validation/test split of Section~\ref{sec:Results_Op_learning} (20 train / 4 validation / 8 held-out test trajectories), for up to 4{,}000 epochs, with early checkpoint selection on the validation split. All autoregressive models are initialized with the first $k$ reference states of each test trajectory, where $k$ is the length of their input history ($k=2$ for \modelname\ on the wave equation and $k=1$ on AC and AD, $k=2$ for MGN and Geo-FNO, $k=4$ for TensorPILS, and $k=10$ for Transolver), whereas FEN, RIGNO, and DeepONet start from the initial condition only. These initialization states are included in the ID window with zero error.

\subsection{Evaluation Metrics}\label{sec:Metrics}

Let $U_{s,t,i}$ and $\hat U_{s,t,i}$ denote the reference and predicted state for test trajectory $s=1,\dots,S$, timestep  $t=0,\dots,T-1$, and node $i=1,\dots,N$.  For a time window $[t_0,t_1)$ (for ID, this is  $[0,\text{train\_horizon})$, whilst for  OOD: $[\text{train\_horizon},T)$),  we compute for each trajectory the   \emph{trajectory-level relative $L_2$ error} over all timesteps and nodes in the window, defined as: 
\begin{equation}
    e_s = \frac{\big\lVert \mathrm{vec}(\hat U_{s,t_0:t_1}) - \mathrm{vec}(U_{s,t_0:t_1}) \big\rVert_2}{\big\lVert \mathrm{vec}(U_{s,t_0:t_1}) \big\rVert_2}, \qquad s = 1,\dots,S,
    \label{eq:traj-rel-l2}
\end{equation}

where $\mathrm{vec}(\cdot)$ flattens the (time, node) axes of the window into a single vector. Then, the error metrics reported in Section~\ref{sec:Results_Op_learning}, correspond to the mean and standard deviation of $\{e_s\}_{s=1}^S$  for the ID and OOD windows separately. For visualizations presented later on in this section, we also compute the  \emph{per-timestep relative} $L_2$ error, by taking, for each sample $s$, the relative $L_2$ norm over nodes only and then averaging  that ratio across samples.

\begin{equation}
      \text{Relative $L_2$ error=} \quad  \frac{1}{S}\sum_{s=1}^{S} \frac{\big\lVert \hat U_{s,t,:} - U_{s,t,:} \big\rVert_2}{\big\lVert U_{s,t,:} \big\rVert_2}, \label{eq:rel-l2-t}
\end{equation}

 as well as the \emph{running root mean square error (RMSE)}, computing the  joint RMSE  over every sample, node, and elapsed timestep from $0$ to $t$:
  
\begin{equation}
  \text{Running RMSE} =  \quad   \sqrt{\frac{1}{(t+1)SN}\sum_{\tau=0}^{t}\sum_{s,i}\big(\hat U_{s,\tau,i}-U_{s,\tau,i}\big)^2}, 
  \label{eq:running-rmse}
\end{equation}

 \begin{figure}[t]
    \centering
    \includegraphics[width=1\columnwidth,trim={0 0 0 0pt},clip]{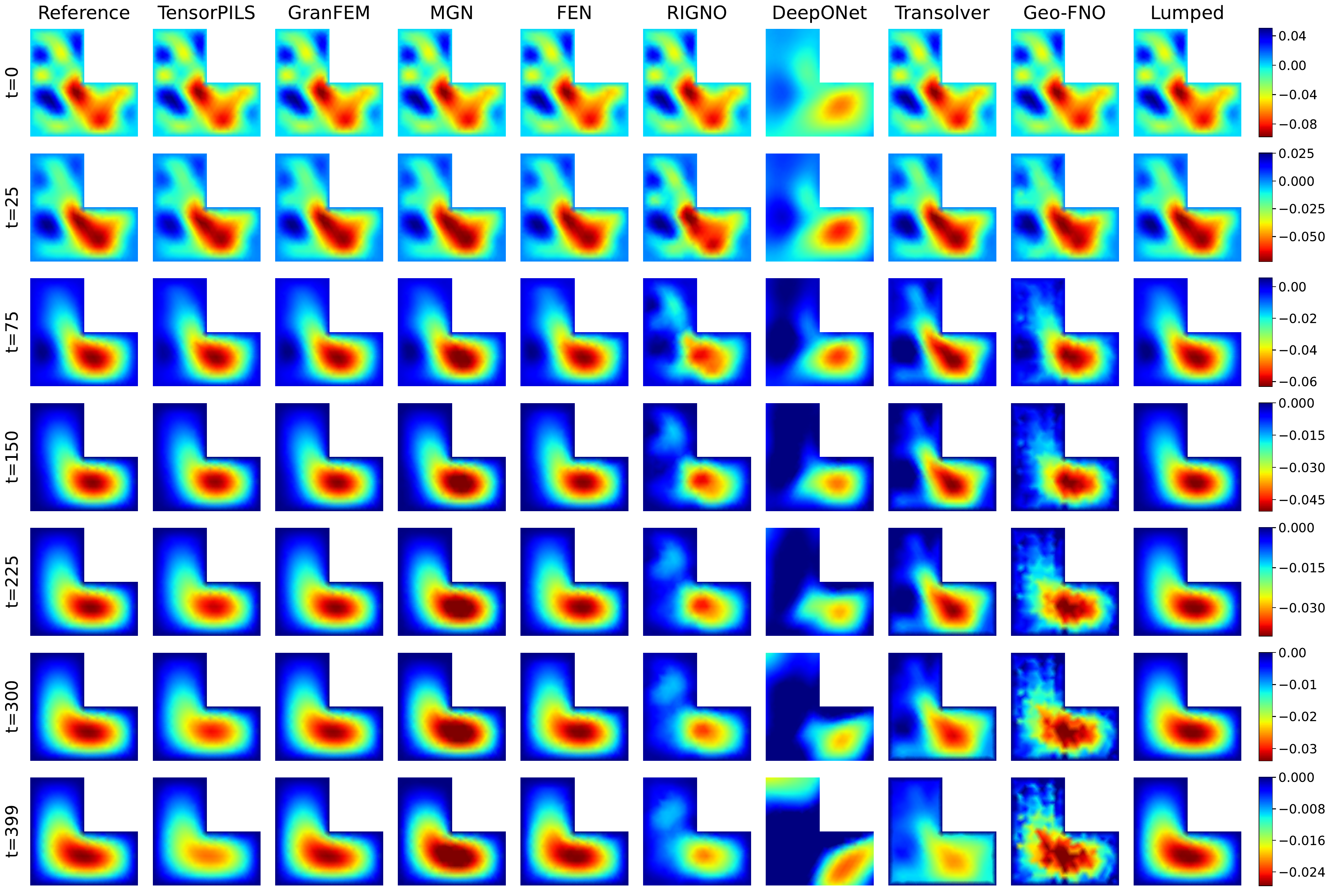}
   
    \caption{Allen--Cahn (\emph{Implicit}) rollout comparison. Representative ID and OOD solution snapshots across operator-learning methods.}
    \label{fig:Rollout_AC}
   
\end{figure}

\subsection{Additional Experiments and Visualizations} \label{sec:Appendix_Operator_Learning_Additional_Results}

\subsubsection{Graph-Driven Operator Learning}\label{sec:Appendix_Graph_Driven_Op_learning}

\begin{figure}
    \centering
    \includegraphics[width=0.95\columnwidth,trim={0 0 0 5pt},clip]{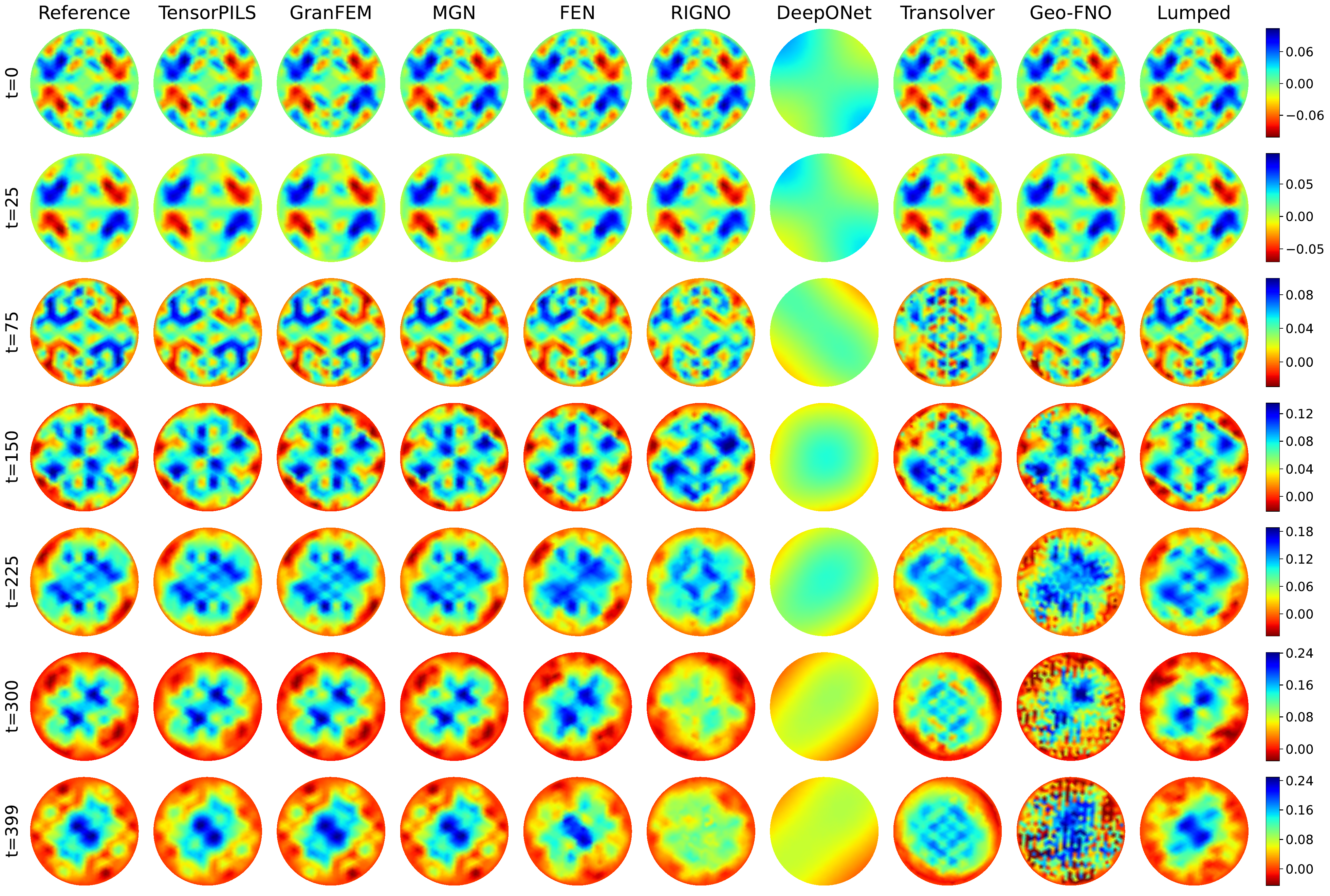}
    \vspace*{-6pt}
    \caption{Wave rollout comparison. Representative ID and OOD solution snapshots across operator-learning methods}
    \label{fig:Rollout_Wave}
    \vspace*{-12pt}
\end{figure}

\begin{figure}
    \centering
    \includegraphics[width=0.95\columnwidth,trim={0 0 0 5pt},clip]{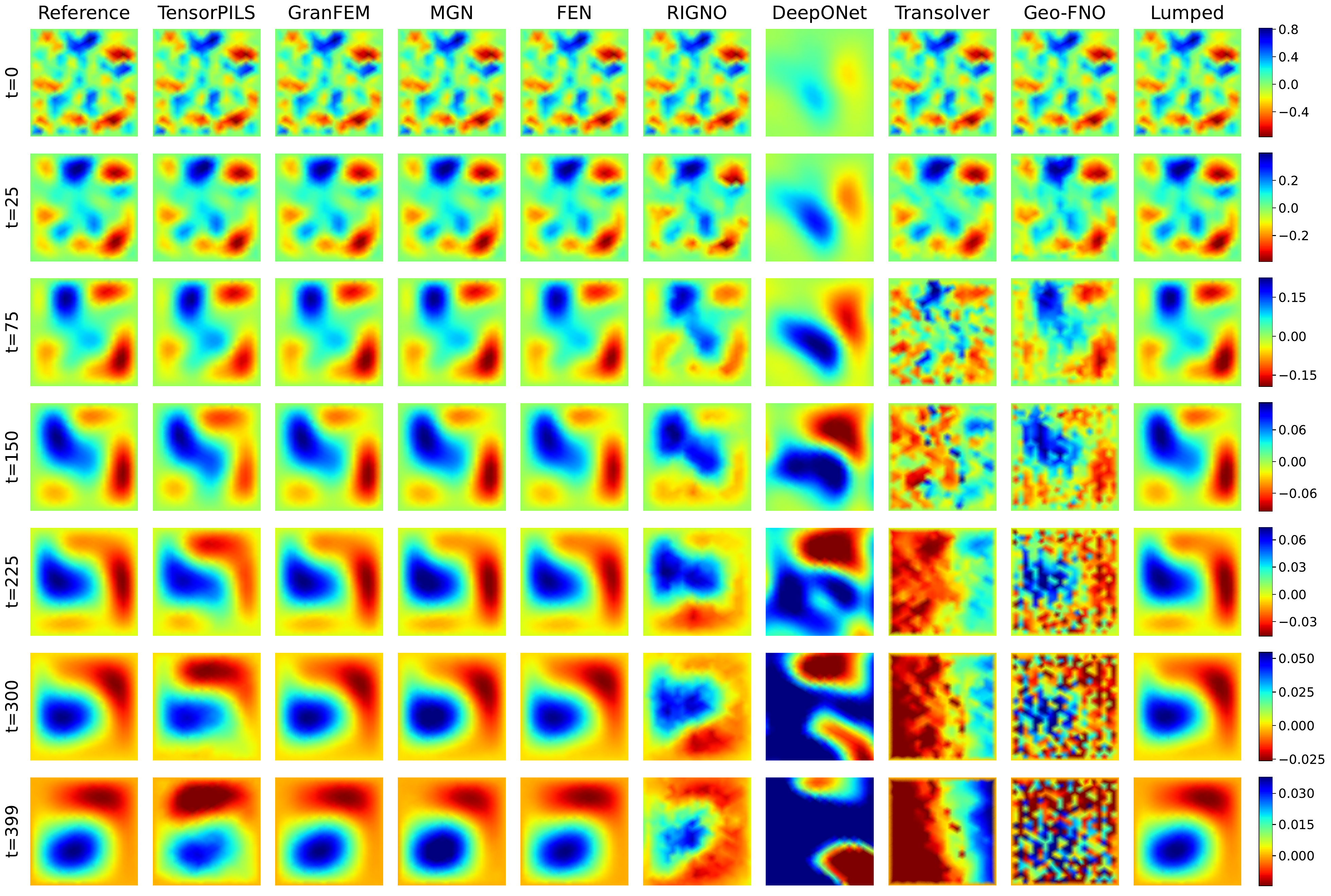}
    \vspace*{-6pt}
    \caption{AD rollout comparison. Representative ID and OOD solution snapshots across operator-learning methods}
    \label{fig:Rollout_AD}
    \vspace*{-12pt}
\end{figure}

\paragraph{Visualizations.}  Figures~\ref{fig:Rollout_AC}-~\ref{fig:Rollout_AD}  provide a qualitative visual comparison between the solution snapshots emulated by each learnable model and the ground truth/reference FEM-generated data for the AC, wave, and AD equations, respectively. Specifically, the columns comprise (i) the reference, (ii) TensorPILS, (iii) \modelname, (iv) MGN, (v) FEN, (vi) RIGNO, (vii) DeepONet, (viii) Transolver, (ix) Geo-FNO, and (x) Lumped, whilst the rows correspond to the snapshots taken at $t = 0, 25, 75, 150, 225, 300, 399$, from a held-out test trajectory, depicting the solution evolution within the ID and OOD regimes (the color scale of each row is set by the reference).  For the AC equation,  we can observe that TensorPILS, \modelname , MGN,   FEN, and the lumped approximation remain visually close to the reference heatmap across the full rollout. For RIGNO, the heatmap visual similarity starts to deteriorate in the OOD regime, Geo-FNO develops node-level noise,  Transolver progressively loses track of the trajectory,  whilst DeepONet's one-shot predictions deviate significantly from the ground truth in both regimes. Similar observations hold for the wave and AD trajectories, where the noise of Geo-FNO is more pronounced.

\begin{figure}
    \centering
    \includegraphics[width=1\columnwidth,trim={0 0 0 0pt},clip]{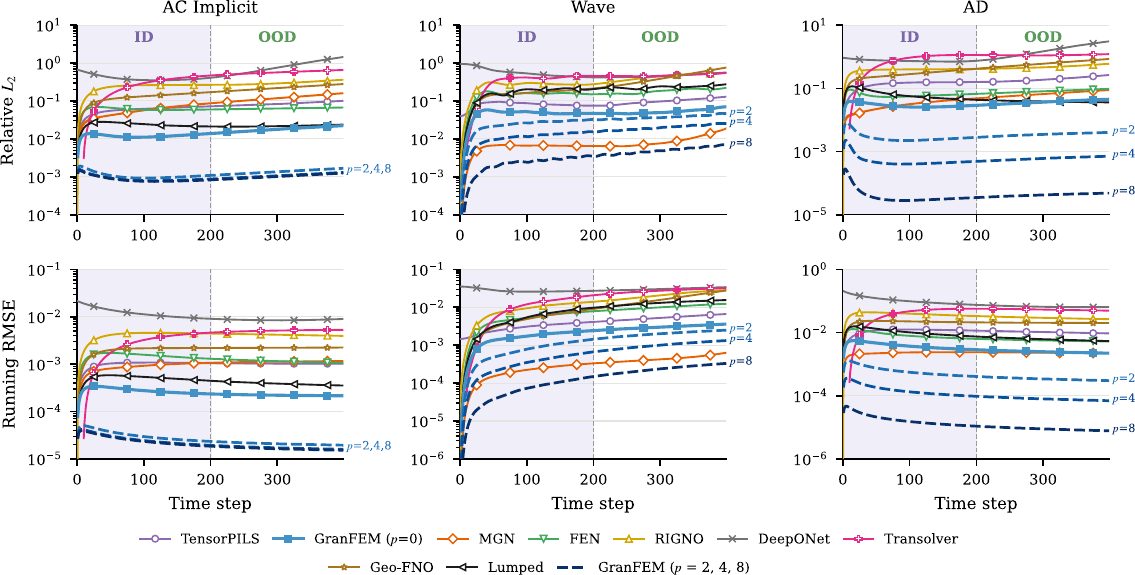}
    \caption{Per-step relative $L_2$ error (top) and running RMSE (bottom) over the rollout for AC Implicit, wave, and AD, averaged over the test trajectories; shaded: training horizon; dashed: \modelname\ with Neumann order $p$.}
    \label{fig:Error_Curves_Operator_Learning}

\end{figure}

Complementary to heatmap visualizations, in Figure~\ref{fig:Error_Curves_Operator_Learning} we present the relative $L_2$ error at each time step (top) and the running RMSE, \ie the absolute error accumulated up to the current step (bottom).  The error metrics are consistent with those observed earlier: \modelname yields the lowest errors for AC and, beyond the initial transient, for AD, whereas for the wave equation, MGN is more accurate than \modelname with $p=0$. 
Note that for AC and AD, the solution decays over the rollout (for AD, its norm decreases by a factor of about six). Since the trajectory-level error in Table~\ref{table:operator_results} normalizes the error accumulated over the whole window by the norm of the whole reference trajectory, as per (\ref{eq:traj-rel-l2}), each step contributes in proportion to its squared amplitude. Hence, like the running RMSE, this error is dominated by the early, large-amplitude steps, whereas the per-step relative error of (\ref{eq:rel-l2-t}) weighs all steps equally and also exposes the late-time behavior. For AD, MGN is the most accurate model during the initial transient, which explains its lower ID error in Table~\ref{table:operator_results}; beyond that, its per-step error increases steadily, whereas that of \modelname\ remains nearly constant. A similar growth of the MGN error is observed for AC. For both dissipative PDEs, the lumped base operator approaches \modelname\ at late times, as the smoothing of the solution removes the fine scales that mass lumping misrepresents, and the learned correction is thus most effective in the transient regime. For the non-dissipative wave equation, the errors of all methods grow over the rollout, most markedly for the lumped base operator, which accumulates phase errors.

Overall, \modelname\ shows the most stable error growth among the learned models, and increasing the Neumann order, $p$, shifts its entire error curve downwards. For AC Implicit, all $p \ge 2$ lie about an order of magnitude below $p=0$, for the wave equation, $p=8$ remains below MGN over the whole rollout, and for AD, $p=2$, $4$, and $8$ lie about one, two, and three orders of magnitude below all other methods. As shown in Table~\ref{table:mgn_depth}, this fidelity improvement comes at no additional computational cost at this resolution (the scaling with the number of DoFs is discussed in Appendix~\ref{sec:Appendix_Add_Res_Neuman}). Indeed,  the rollout time of \modelname\ remains practically unchanged and two orders of magnitude below that of MGN, which, although the most accurate model for the wave equation at $p=0$, is also one of the slowest, exceeding the rollout time of the reference solver.  To examine whether a lighter MGN attains a more favorable accuracy--rollout time trade-off, we retrain MGN  with a reduced number of message passing steps: $L = 12$, $8$, and $4$, keeping all other settings unchanged. As shown in Table~\ref{table:mgn_depth}, reducing the depth to $L=8$ retains the accuracy and lowers the rollout time to $1.7$\,s, whereas $L=4$ degrades the accuracy; however, all variants remain slower than the reference solver, since on these meshes the rollout time is dominated by the per-operation latency of the network layers rather than by the number of nodes. In contrast, \modelname\ with $p=8$ attains a lower error than every MGN variant at a rollout time two orders of magnitude lower.

\begin{table}[h]
\centering
\caption{Wave equation: MGN with $L$ message-passing steps, compared with \modelname\ and the reference solver. Test relative $L_2$ error (ID/OOD) and rollout time per trajectory.}
\vspace{-2mm}
\label{table:mgn_depth}
\scriptsize
\setlength{\tabcolsep}{4pt}
\resizebox{0.88 \columnwidth}{!}{
\begin{tabular}{lccccccc}
\toprule
& \multicolumn{4}{c}{MGN} & \multicolumn{2}{c}{\modelname} & \multirow{2}{*}{Solver (CG)} \\
\cmidrule(lr){2-5}\cmidrule(lr){6-7}
& $L=15$ & $L=12$ & $L=8$ & $L=4$ & $p=0$ & $p=8$ & \\
\midrule
Relative $L_2$ (ID/OOD) & 0.006/0.010 & 0.007/0.010 & 0.005/0.010 & 0.010/0.017 & 0.047/0.053 & 0.003/0.005 & -- \\
Rollout time (ms)       & 2809 & 2315 & 1676 & 1103 & 9.5 & 9.5 & 916 \\
\bottomrule
\end{tabular}}
\end{table}

  Finally, to separate the contribution of the improved approximation of $\mathbf{M}^{-1}$ from that of the learned correction, Table~\ref{table:neumann_base} compares \modelname with the deterministic Neumann update, \ie without the learned correction, for the same $p$.   At every $p$, \modelname\ attains an ID error lower than or equal to that of the deterministic update. The learned correction is most decisive at $p=0$, where it reduces the error of the lumped update  across all PDEs. For the wave and AD equations, its relative contribution decreases as $p$ grows, since the truncated series itself converges towards the consistent scheme, and for $p=8$ both attain errors well below $1\%$. Notably, for AD, the deterministic update already yields lower OOD errors for $p \geq 2$.  For AC, in contrast, the deterministic update saturates for $p \geq 2$ ($0.012/0.015$ and $0.006/0.013$ for the \emph{Explicit} and \emph{Implicit} datasets, respectively), 
  the learned correction can, however, further lower the error. 
  The Neumann expansion and the learned correction are thus complementary: the former recovers the long-range part of $\mathbf{M}^{-1}$, whereas the latter compensates for the remaining modeling errors of the baseline update, without increasing the per-step rollout cost.   We finally note that the reported rollout times in Section~\ref{sec:Results_Op_learning} exclude the one-off setup, \ie the evaluation of the learned edge weights and, for $p>0$, the precomputation of the Neumann operator, which at this resolution amounts to $1.5$--$18$ ms.  The same convention is applied to all methods: the rollout times of the baselines exclude the construction of their graphs and static input features, and those of the reference solver exclude the assembly of the finite element matrices and the setup of the preconditioner, so that only the time-stepping loop is measured.


\begin{table*}[t]
\centering
\caption{Contribution of the learned correction: test relative $L_2$ error (ID/OOD) of \modelname\ and of the deterministic Neumann update (Base) without learned correction for the same Neumann order $p$; $p=0$ corresponds to the lumped approximation.}
\label{table:neumann_base}
\scriptsize
\setlength{\tabcolsep}{3pt}
\renewcommand{\arraystretch}{0.95}
\resizebox{0.9\textwidth}{!}{%
\begin{tabular}{c cc cc cc cc}
\toprule
& \multicolumn{2}{c}{AC \emph{Explicit}} & \multicolumn{2}{c}{AC \emph{Implicit}} & \multicolumn{2}{c}{Wave} & \multicolumn{2}{c}{AD} \\
\cmidrule(lr){2-3}\cmidrule(lr){4-5}\cmidrule(lr){6-7}\cmidrule(lr){8-9}
$p$ & Base & \modelname & Base & \modelname & Base & \modelname & Base & \modelname \\
\midrule
0 & 0.030/0.041 & \textbf{0.019/0.026} & 0.024/0.022 & \textbf{0.012/0.018} & 0.191/0.234 & \textbf{0.047/0.053} & 0.081/0.040 & \textbf{0.031/0.036} \\
2 & 0.012/0.015 & \textbf{0.008/0.009} & 0.007/0.013 & \textbf{0.001/0.001} & 0.043/0.054 & \textbf{0.028/0.038} & 0.006/0.0004 & \textbf{0.004}/0.003 \\
4 & 0.012/0.015 & \textbf{0.008/0.009} & 0.006/0.013 & \textbf{0.001/0.001} & 0.017/0.025 & \textbf{0.013/0.020} & 0.0012/0.0001 & \textbf{0.0010}/0.0006 \\
8 & 0.012/0.015 & \textbf{0.008/0.009} & 0.006/0.013 & \textbf{0.001/0.001} & 0.003/0.006 & \textbf{0.003/0.005} & 0.0001/0.0000 & 0.0001/0.0000 \\
\bottomrule
\end{tabular}%
}
\end{table*}

\subsubsection{DoF and Domain Extrapolation} \label{sec:Appendix_Domain_Extr}


\paragraph{Dataset Generation.} For these experiments, we generate the samples for the different element sizes with the explicit scheme of (\ref{eq:wave_fem_step}) and the CG numerical solver. Note that forming $\mathbf{M}^{-1}$ explicitly becomes impractical beyond $\approx 3\cdot10^4$ DoFs, at $\mathcal{O}(\mathrm{DoF}^2)$ memory and $\mathcal{O}(\mathrm{DoF}^3)$ time. The explicit scheme is conditionally stable, requiring $\Delta t$ to shrink in proportion to the element size $h$. Since  $\Delta t$ is held fixed across the DoF and domain sweeps, so that the per-step operators of all models remain directly comparable (a model trained across time steps is left for future work), 
the finest resolution ($h = 0.0061$, about $10^5$ DoFs) is selected such that it lies close to the stability limit $(c\Delta t)^2\lambda_{\max}(\mathbf{M}^{-1}\mathbf{K}) \le 4$ of the central-difference scheme. 



For the domain extrapolation, as mentioned in Section~\ref{sec:Results_Extrapolation}, we consider the cases where we (i) rescale the disk to radius $2$, (ii) shift its centre to $(1,1)$, and (iii) rescale it to radius $0.5$ and shift its centre to $(0.5,0.5)$. The datasets are generated as above. Note that since the domain size is not the same for each case, the resulting DoFs also vary, as shown in Table~\ref{tab:Domain_extrapolation_h_to_dof}. To isolate the impact of mesh resolution, for the DoF extrapolation case, we consider the same 8  unknown held-out initial conditions and trajectories considered in Section~\ref{sec:Results_Op_learning}. For the domain extrapolation, we use the same 8 initial-condition draws across all three conditions (Rescale/Shift/Rescale+Shift), because though the underlying domain differs from training in every case, the resulting trajectories are genuinely novel regardless of which IC subset is used.

\begin{table}[h]
\centering
\caption{Mapping from mesh resolution $h$ to node count (DoF) for each domain-change condition in Table~\ref{table:Extrapolation_domain}.}
\label{tab:Domain_extrapolation_h_to_dof}
\scriptsize
\setlength{\tabcolsep}{4pt}

\resizebox{0.58 \columnwidth}{!}{\begin{tabular}{lcccccccc}
\toprule
$h$ & 0.16 & 0.12 & 0.10 & 0.08 & 0.04 & 0.02 & 0.01 & 0.008 \\
\midrule
Rescale       & 633 & 1095 & 1550 & 2407 & 9341 & 36798 & 146077 & 227899 \\
Shift         & 181 & 295  & 411  & 634  & 2407 & 9340  & 36805  & 57333 \\
Rescale+Shift & 54  & 89   & 123  & 181  & 634  & 2407  & 9340   & 14478 \\
\bottomrule
\end{tabular}}
\end{table}

\begin{figure}[t]
    \centering

    \begin{subfigure}{0.86\columnwidth}
        \centering
        \includegraphics[width=\linewidth   ]{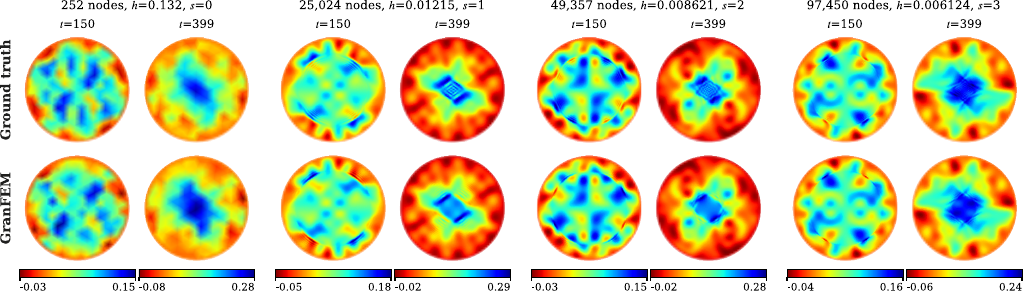}
        \caption{Representative ground-truth and \modelname{} wave solution snapshots across unseen discretizations.}
        \label{fig:Dof_Heatmaps_a}
    \end{subfigure}

    \vspace{4pt}

    \begin{subfigure}{0.86\columnwidth}
        \centering
        \includegraphics[width=\linewidth ]{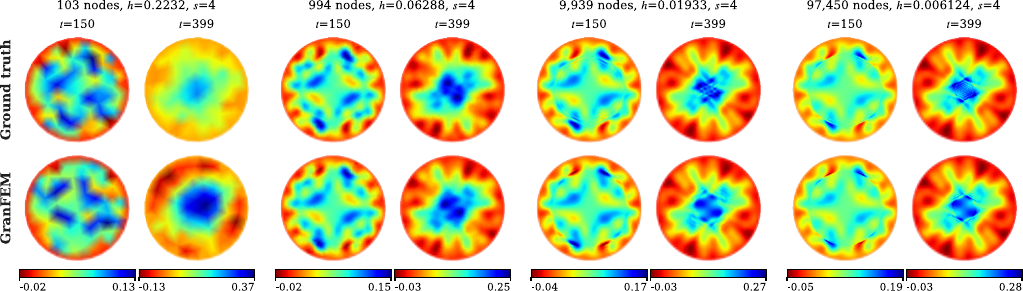}
        \caption{Ground truth and \modelname for the same sample across different mesh resolutions $h$.}
        \label{fig:Dof_Heatmaps_b}
    \end{subfigure}

    \vspace*{-6pt}
    \caption{Zero-shot mesh-resolution extrapolation: ground-truth and \modelname\ wave solutions at an ID ($t=150$) and an OOD ($t=399$) step.}
    \label{fig:Dof_Heatmaps}
    \vspace*{-12pt}
\end{figure}

\begin{figure}[t!]
\centering 
\includegraphics[width=0.9\columnwidth,  ]{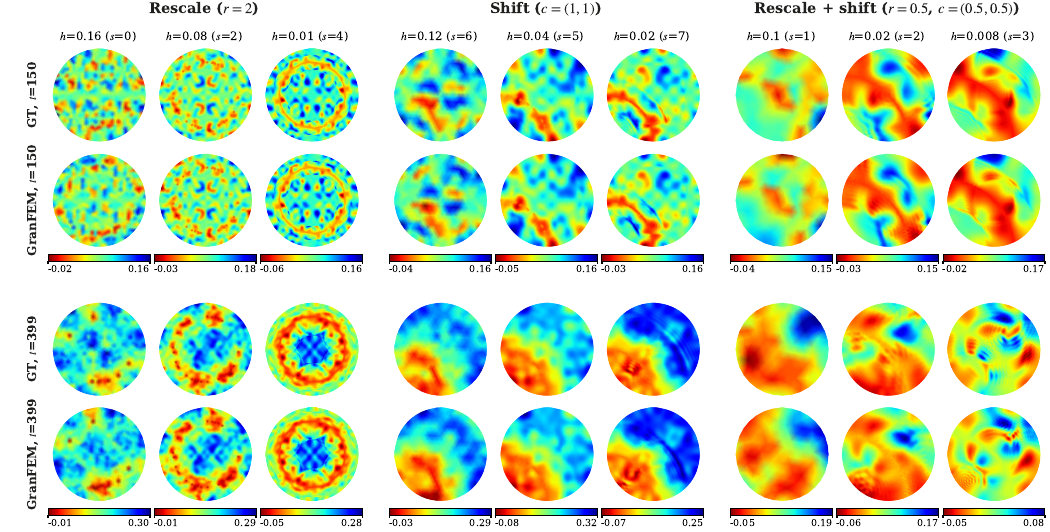} \vspace*{-6pt} 
\caption{Zero-shot domain extrapolation: representative ground-truth and \modelname wave solutions across unseen domains and mesh resolutions.} \label{fig:Domain_Heatmaps} \vspace*{-12pt} 
\end{figure}

\paragraph{Visualizations.} Figures~\ref{fig:Dof_Heatmaps} and~\ref{fig:Domain_Heatmaps} provide a qualitative assessment of the zero-shot DoF and domain-generalization experiments reported in Table~\ref{table:Extrapolation_dof} and~\ref{table:Extrapolation_domain}, respectively, comparing \modelname predictions with the corresponding ground truth solutions. In particular, Figure~\ref{fig:Dof_Heatmaps_a} depicts snapshots of the reference and \modelname-generated wave solution trajectories for distinct samples at different resolutions, at both an ID and an OOD timestep, $t=150$ and $t=399$, respectively. Consistent with the quantitative results in Table~\ref{table:Extrapolation_dof}, the predicted and ground truth solutions exhibit strong visual resemblance at finer resolutions ($\mathrm{DoF}=25{,}000,$ $50{,}000$, and $100{,}000$), whereas, for the coarsest resolution, it can be observed that there is an overt dissimilarity between the two solutions for the OOD snapshot. Figure~\ref{fig:Dof_Heatmaps_b} shows how the ground truth and inferred solutions vary across progressively finer meshes for a fixed sample. Again, \modelname closely reproduces the reference solution at finer resolutions, while  discrepancies arise on coarse meshes. An important remark though, is that the ground-truth solution itself has not yet converged on these coarse discretizations: the field obtained at $h=0.132$ differs substantially from the fine-mesh solution at $h\approx0.006$, whereas the solution steadily becomes consistent as the mesh granularity increases. Consequently, the mismatch at coarse resolutions is less informative, since the corresponding numerical target is not yet representative of the converged solution. Figure~\ref{fig:Domain_Heatmaps} presents the corresponding comparisons for the domain extrapolation experiments (rescale, shift, shift and rescale), depicting snapshots of the solution at ID and OOD timesteps for different samples, $s$, and element sizes, $h$. The coarser meshes, \ie larger element sizes, are presented on the left side of the figure, whilst the mesh granularity increases towards the right. Aligned with the results of Table~\ref{table:Extrapolation_domain}, discrepancies are more apparent on the coarsest target discretizations, \eg $h = 0.16 $ and $\ 0.12$ for the domain rescale and shift, respectively, whereas agreement with the FEM reference becomes increasingly close as the target meshes are refined ($h < 0.08$).

\subsubsection{Neumann terms across resolutions and domains} \label{sec:Appendix_Add_Res_Neuman}

\begin{figure}[h]
\centering 
\includegraphics[width=0.9\columnwidth,  ]{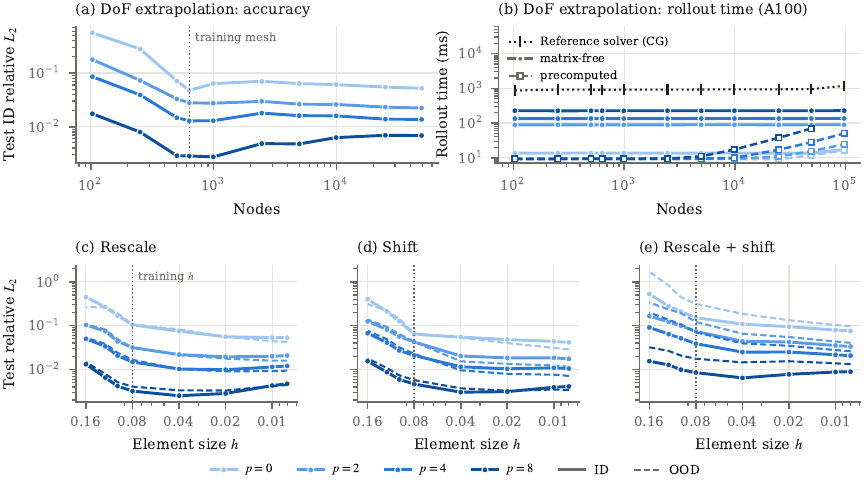} \vspace*{-6pt} 
\caption{Zero-shot extrapolation of \modelname\ (wave) for increasing Neumann order $p$. (a) Test ID relative $L_2$ error against the number of nodes (dotted: training mesh); (b) rollout time per trajectory on an A100 with the matrix-free (solid) and the precomputed (dashed) evaluation of the Neumann terms, and the reference CG solver (dotted). (c)--(e) ID (solid) and OOD (dashed) error against the element size $h$ on the rescaled, shifted, and rescaled-and-shifted domains (dotted: training $h$).}
 \label{fig:neumann_scaling} \vspace*{-12pt} 
\end{figure}

Figure~\ref{fig:neumann_scaling} shows the zero-shot accuracy and trajectory rollout time of \modelname\ while increasing Neumann order $p$. On meshes with more than $10^3$ nodes, the ID error decreases from $5$--$7\%$ ($p=0$) to $2$--$3\%$ ($p=2$), $1.3$--$1.8\%$ ($p=4$), and $0.3$--$0.7\%$ ($p=8$), whilst the coarsest meshes benefit most (at $250$ nodes, from $0.28$ to $0.008$). 
With the precomputed operator, the rollout time stays at about $9$\,ms up to $2.5\times10^3$ nodes, whilst beyond that the rollout time starts increasing for all $p$ values ($17$, $38$, and $69$\,ms for $p=8$ at $10^4$, $2.5\times10^4$, and $5\times10^4$ nodes), remaining though at least an order of magnitude below the reference solver. At $10^5$ nodes, the \emph{precomputed} operator with $p=8$ exceeds the memory of our A100 40 GB GPU, where the \emph{matrix-free} evaluation (which keeps the memory of the one-hop graph at the cost of $p$ additional sparse products per step) can be used instead.  With the matrix-free evaluation, the rollout time is independent of the mesh size on the meshes considered ($14$, $89$, $134$, and $226$\,ms for $p = 0, 2, 4, 8$ from $10^2$ to $10^5$ nodes), as it is dominated by the latency of the $p+1$ sparse products per step, and  still remains below the reference solver.

\end{document}